\documentclass[preprint,12pt]{elsarticle}

\usepackage{lineno,hyperref}
\modulolinenumbers[5]
\usepackage[table]{xcolor}
\usepackage{booktabs} 
\usepackage{hyperref}       
\usepackage{url}            
\usepackage{booktabs}       
\usepackage{amsfonts}       
\usepackage{nicefrac}       
\usepackage{microtype}      

\usepackage{amsmath,amssymb,amsthm}             
\usepackage{dsfont}
\usepackage{mathtools}
\usepackage{amsthm}
\usepackage{amsmath}

\usepackage{algorithm}
\usepackage{algpseudocode}

\usepackage{tikz}

\usepackage{array}
\usepackage{multirow}
\usepackage{cleveref}

\usetikzlibrary{shapes,backgrounds, calc, shadings, arrows,decorations.pathmorphing,backgrounds,fit,positioning,shapes.symbols,chains}
\pgfdeclareradialshading{ballshading}{
 \pgfpoint{-10bp}{10bp}}
 {color(0bp)=(gray!50!black);
  color(9bp)=(gray!50!black);
  color(18bp)=(gray!50!black);
  color(25bp)=(gray!50!black);
  color(50bp)=(black)}

\tikzstyle{peers}=[draw,circle, minimum width=10pt]
\tikzstyle{superpeers}=[draw,circle,minimum width=20pt]

\tikzset{
>=stealth',
help lines/.style={dashed, thick},
axis/.style={<->},
important line/.style={thick},
connection/.style={thick, dotted},
}

\newcommand\cmbox[1]{
  \fbox{\lower0.75cm
    \vbox to 1.0cm{\vfil
      \hbox to 1.7cm{\hfil\parbox{1.4cm}{\centering #1}\hfil}
      \vfil}%
  }%
}

\newcommand\cmlegend[1]{
  {\lower0.75cm
    \vbox to 1.0cm{\vfil
      \hbox to 0.7cm{\hfil #1}
      \vfil}%
  }%
}

\DeclareMathOperator*{\argmin}{arg\,min}

\def\HCBR{{\sc HCBR}}
\def\bfHCBR{{\sc \bf HCBR}}

\newenvironment{nalign}{
    \begin{equation}
    \begin{aligned}
}{
    \end{aligned}
    \end{equation}
    \ignorespacesafterend
}

\theoremstyle{definition}
\newtheorem{definition}{Definition}[section]

\journal{Information Systems}

\begin{document}

\begin{frontmatter}

\title{Binary Classification in Unstructured Space With Hypergraph Case-Based Reasoning}

\author{Alexandre Quemy}
\address{IBM Krakow Software Lab, Cracow, Poland}
\ead{aquemy@pl.ibm.com}
\address{Faculty of Computing, Pozna\'n University of Technology, Pozna\'n, Poland}

\begin{abstract}
 Binary classification is one of the most common problem in machine learning. It consists in predicting whether a given element belongs to a particular class. In this paper, a new algorithm for binary classification is proposed using a hypergraph representation. The method is agnostic to data representation, can work with multiple data sources or in non-metric spaces, and accommodates with missing values. As a result, it drastically reduces the need for data preprocessing or feature engineering. Each element to be classified is partitioned according to its interactions with the training set. For each class, a seminorm over the training set partition is learnt to represent the distribution of evidence supporting this class. 

Empirical validation demonstrates its high potential on a wide range of well-known datasets and the results are compared to the state-of-the-art. The time complexity is given and empirically validated. Its robustness with regard to hyperparameter sensitivity is studied and compared to standard classification methods. Finally, the limitation of the model space is discussed, and some potential solutions proposed.
\end{abstract}


\end{frontmatter}


\section{Introduction}\label{sec:intro}

In many real-life situations, one tries to take a decision based on previous {\it similar} situations. Each situation is described by a certain amount of relevant information, either collected by an expert, or automatically e.g. by some sensors. Those situations share similarities on which to make analogies or counter-examples in order to take new decisions. Conversely, in general, if two situations do not share any common characteristic, then they are totally independent, i.e. it is impossible to infer one's outcome from the other one. The purpose of supervised machine learning algorithms is to exploit the available information and interactions between past cases or examples in order to build a model or infer the key rules to take correct decisions.

Due to the large variety of concrete situations that can be reduced to binary classification, it is one of the most studied problem in machine learning. In this paper, we investigate the problem of binary prediction under a supervised setting.

This paper {\bf contributes} to binary classification with a new algorithm called Hypergraph Case-Based Reasoning (\HCBR). The idea is to create a hypergraph where each element of the training set is a hyperedge and vertices are represented by the features describing the elements. The intersections between edges create a partition, unique to a hypergraph. For each case, we model the support as a convex combination of the elements of this partition. Each of those elements is valued according to its importance w.r.t. the set of all the hyperedges it belongs to and their respective labels.

The paper is an extension of \cite{DBLP:conf/dolap/Quemy18} with a focus on separating the abstract framework from the specific implementation used for the experiments. On top of the previous comparison to the best literature results, we extended the empirical validation with a comparison between \HCBR~and nine alternative methods. {We also added a validation on unstructured datasets.} A study of learning curves and model spaces, highlighted some properties of \HCBR~and the limitations of the current model space, thus shaping priority axes for future work. Among other refinements, matricial formulation for the model space and model selection is provided to ease efficient implementation. More evidence concerning model calibration is added.

The plan of the paper is as follows: in Section \ref{sec:problem}, we present the problem of binary classification {and related work.
In particular, in Section \ref{sec:linear_class}, we present the well-understood and unified framework to solve binary classification. Section \ref{sec:beyond_lin_class} briefly introduces other approaches to classification. Section \ref{sec:bc_unstructured} highlights the necessity of working in unstructured spaces and presents related work w.r.t. metric learning and data wrangling. A formulation of binary classification in an abstract space of information is proposed in Section \ref{sec:unstructured_spaces_problem}.
The main contribution of this paper is Section \ref{sec:hcbr} which defines \HCBR~ framework. The rest of the paper focuses on empirical validation: Section \ref{sec:experiments} presents empirical results on seven structured datasets from the UC Irvine Machine Learning Repository (UCI)\footnote{\href{http://archive.ics.uci.edu/ml/index.php}{http://archive.ics.uci.edu/ml/index.php}} and the LIBSVM\footnote{\url{https://www.csie.ntu.edu.tw/~cjlin/libsvmtools/datasets/binary.html}}, while Section \ref{sec:unstructured_datasets} is dedicated to unstructured datasets for text classification. Section \ref{sec:intrinsic_perf} studies important properties, namely the computational time, learning curves and hyperparameters usage. In Section \ref{sec:discussion}, we discuss the current limitations and possible model space extensions. The paper ends in Section \ref{sec:conclusion} with a discussion about the results, future work, and possible improvements.}

{This paper is accompanied by Additional Material\footnote{\url{http://www.cs.put.poznan.pl/events/2018-IS-special-issue.html}}.}

\section{Binary classification and related work}
\label{sec:problem}

Before introducing the problem of binary classification, we present the notations used throughout this paper. Vectors are denoted in bold and small case (e.g. $\mathbf x$) and their components in small case (e.g. $x_i$). A collection of vectors is denoted in bold and large case (e.g. $\mathbf X$). $|\mathbf x|$ (resp. $|\mathbf X|$) denotes the cardinal of the vector $\mathbf x$ (resp. the collection $\mathbf X$) while $||\mathbf x||$ is its norm. The scalar product is denoted by $<.,.>$. For a matrix $A = (a_{ij})$, $a_{i:}$ is the $i$th row vector and $a_{:j}$ the $j$th column vector. 

In machine learning, the problem of classification consists in finding a mapping from an input vector space $\mathcal X$ to a discrete decision space $\mathcal Y$ using a set of examples. The binary classification problem is a special case such that $\mathcal Y$ has only two elements. It is often viewed as an approximation problem s.t. we want to find an estimator $\bar J$ of an unknown mapping $J$ available only through a sample called {\it training set}. A training set $(\mathbf{X}, \mathbf{y})$ consists of $N$ input vectors $\mathbf{X} = \{ \mathbf{x}_1, ..., \mathbf{x}_N \}$ and their associated correct class $\mathbf{y} = \{ y_i =   J(\mathbf{x}_i) \}^{N}_{i=1}$.

Let $\mathcal{J}(\mathcal X, \mathcal Y)$ be the class of mappings from $\mathcal X$ to $\{-1,1\}$, or simply $\mathcal{J}$ if there is no ambiguity. A machine learning algorithm to solve binary classification is an application $\mathcal A: \mathcal{X}^N \times \mathcal{Y}^N \to \mathcal{J}$ capable of providing a good approximation for any $J \in \mathcal{J}$ under some assumptions on the {\it quality} of the training set. In practice, it is not reasonable to search directly in $\mathcal{J}$ and some assumptions on the ``shape'' of $J$ are made s.t. $\bar J = \mathcal{A}(\mathbf{X}, \mathbf{y})$ belongs to a {\it hypothesis} space or model space $\mathcal H \subset \mathcal{J}$. This restriction implies not only that the exact mapping $J$ is not always reachable but might also not be approximated correctly by any element of $\mathcal H$. The choice of the model space is thus crucial as it should be large enough to represent fairly complex functions and small enough to easily find the best available approximation.

In general, a robust classification algorithm must able to approximate correctly any possible mapping. The problem of finding such algorithm consists in minimizing the generalization error for all possible mappings. Formally, it consists in solving:
\begin{equation}
\label{eq:pb}
\underset{\mathcal A}{\min} \sum_{J \in \mathcal{J}} \int_{\mathcal{X}} ||J(\mathbf{x}) - \bar J(\mathbf{x}) ||\mu(\mathbf{x}) \mathop{d\mathbf{x}}
\end{equation} where $\mu$ is a probability measure over $\mathcal{X}$.

In practice, the generalization error cannot be computed, and the set of possible mappings is unreasonably large. For those reasons, we aim at minimizing the empirical classification error on a reasonably large set of datasets $\mathcal D =  \{ (\mathbf{X}_1, \mathbf{y}_1), ..., (\mathbf{X}_K, \mathbf{y}_K)  \}$, i.e.
\begin{equation}
\underset{\mathcal A}{\min} \sum_{(\mathbf{X}, \mathbf{y}) \in \mathcal D}  \sum_{ (\mathbf{x}, y) \in (\mathbf{X}, \mathbf{y})} \mathbb{I}_{\{y \neq \bar J(\mathbf{x})\}}.
\end{equation}

To highlight the differences between standard classification problem and our approach, we briefly present linear models for classification followed by the challenges of classification in unstructured spaces. For an overview of theoretical results on classification, we refer the reader to \cite{boucheron2005theory}.

\subsection{Linear binary classification}
\label{sec:linear_class}

The problem of binary classification is commonly studied with $\mathcal{X} = \mathbb{R}^{M}$. Many popular classification approaches such as SVM \cite{vapnik2013nature}, perceptron \cite{rosenblatt1958perceptron} or logistic regression \cite{cox1958regression} define the model space as the set of $M$-hyperplanes. A $M$-hyperplane is uniquely defined by a vector $\mathbf{w} \in \mathbb{R}^{M}$ and a bias $w_0 \in \mathbb{R}$, and is formulated by
\begin{align}
 \label{eqn:hyperplan}
 h_{\mathbf{w}}(\mathbf{x}) = <{\mathbf{w}}, {\mathbf{x}}> + w_0 = 0
\end{align} The {\it homogeneous} notation consists in adding $w_0$ to $\mathbf{w}$ by rewriting $\mathbf{x}$ such that $\mathbf{x} = (1, x_1, ..., x_M)$. The hyperplane equation \eqref{eqn:hyperplan} is then expressed by $h_{\mathbf{w}}(\mathbf{x}) = <{\mathbf{w}}, {\mathbf{x}}>$.
A hyperplane separates $\mathbb{R}^{M}$ into two regions, and thus, can be used as a discriminative rule s.t.
\begin{align}
 \bar J_{\mathbf{w}}(x) = \left\{\begin{matrix}
1 & ~h_{\mathbf{w}}(\mathbf{x}) > 0\\ 
-1 & ~h_{\mathbf{w}}(\mathbf{x}) \leq 0
\end{matrix}\right.
\end{align}
Then,  given a training set $(\mathbf{X}, \mathbf{y})$, the classification problem is equivalent to finding the best hyperplane s.t. it minimizes a certain {\it loss} function over the training set \cite{lin2004note}:
\begin{align}
 \label{eqn:param_model}
\mathbf{w}^* = \underset{\mathbf{w}}{\argmin} ~ \underset{i=1}{\overset{N}{\sum}} \ell(\mathbf{w}, \mathbf{x}_i) + \lambda R(\mathbf{w})
\end{align} where $R(\mathbf{w})$ is {\it regularization} term to prevent overfitting (usually $||\mathbf{w}||^2_2$ or $||\mathbf{w}||_1$) and $\lambda > 0$ a hyperparameter controlling the effect of regularization. Note that \eqref{eqn:param_model} is a parametric problem due to the choice of model space. Several losses functions exists and are generally based on the {\it margin} of $\mathbf{x}_i$ which is defined by $m(\mathbf{w}, \mathbf{x}_i) = J(\mathbf{x}_i) h_{\mathbf{w}}(\mathbf{x}_i)$. The margin represents the distance of a vector $\mathbf{x}_i$ to the hyperplane defined by $h_{\mathbf{w}}$. It is positive if $\mathbf{x}_i$ is correctly classified, negative otherwise. For the most known, the 0-1 loss, hinge loss and log loss are defined by
\begin{align}
  \begin{matrix}
    \ell_{01}(\mathbf{w}, \mathbf{x}) & = & \mathbb{I}_{\{ m(\mathbf{w}, \mathbf{x}) \leq 0 \}} \\
    \ell_{\text{hinge}}(\mathbf{w}, \mathbf{x}) &= & \max(0, 1 - m(\mathbf{w}, \mathbf{x})) \\
    \ell_{\log}(\mathbf{w}, \mathbf{x})& = & \ln(1 + e^{- m(\mathbf{w}, \mathbf{x})}) 
  \end{matrix}
\end{align}
The Perceptron algorithm uses the 0-1 loss, while SVM minimizes the hinge loss and the Logistic Regression the log loss.

\subsection{Other approaches to binary classification}
\label{sec:beyond_lin_class}

{The method developed in this paper is discriminative and fits the mathematical framework introduced above. Therefore, it models $p(y|\mathbf x)$. There exists other framework to handle binary classification. We briefly introduce them as they will be compared to \HCBR~in Section \ref{sec:experiments}.}

Generative techniques such as Naive Bayes or Bayesian Networks \cite{friedman1997bayesian} estimate the joint probability $p(\mathbf x, y)$. They make assumptions on the data distribution (e.g. a mixture of gaussians) and estimate the unknown parameters to generate predictions. 

Another popular family of classification techniques are based on trees \cite{venables2002tree, breiman2017classification}. For instance, Random Forest \cite{ho1995random,breiman2001random} is one of the most widely used method due to its good performances \cite{couronne2018random} and several extensions have been developed \cite{geurts2006extremely,friedman2001greedy}

Last but not least, Deep Learning is a family of techniques based on stacking layers of neurons whose weights are adjusted using gradient back-propagation \cite{lecun2015deep}. It represents the state-of-the-art in classification in multiple domains \cite{schmidhuber2015deep}: vision, audio and natural language processing to name few.

\subsection{On the necessity and problems of unstructured spaces}
\label{sec:bc_unstructured}

{In machine learning, and broadly speaking in mathematics, the term structured refers to objects that are not living in $\mathbb R^{n}$. In particular, there is no obvious metric between objects. Graphs are an example of structured object and the right notion of distance between two graphs depends on the considered problem.}

{In the database and data science community, the usage is different. A dataset is structured if it can be easily mapped into pre-defined fields or represented by a relational database or tabular-like schema. By opposition, any other dataset is unstructured or semi-structured (raw text, BoW, JSON, XML, graph, etc.). A structured dataset does not guarantee a norm on the rows because some columns might not be expressed in a space with an obvious metric (e.g. categorical variables).}

{A key element to apply classification methods to datasets is a (meaningful) metric, independently of whether or not the data are stored in a structured format. Thus, we qualify of unstructured a space without obvious or informative metric\footnote{A discrete set can always be indexed, however, in this case, the usual distance in $\mathbb N$ is uninformative, therefore not defining a structure.}.} 

{A tremendous amount of data is collected but never used. The data lakes turn into {\it data swamps}, notably because the natural form of data is {\it messy}: not sanitized, in different formats, without informative metric, etc. Most of the time, solving optimally a concrete problem requires using data from multiple sources, thus multiplying the above-mentioned problems. This highlights the necessity to develop algorithms capable to work in any unstructured space.}
{In this paper, we present a method working for any unstructured space, allowing to relax usual constraints on the input space:
\begin{enumerate}
\item {\bf non-metric space:} there is no metric embedded with the data space,
\item {\bf non-homogeneous cardinality:} the number of features per element is not fixed a priori,
\item {\bf representation agnostic:} the concrete {\it representation} of features does not influence the model. 
\end{enumerate}}

{To handle the problem of {\bf non-metric spaces}, one can use {\it metric learning} methods. We present those methods in Section \ref{sec:metric_learning} and discuss how they do not handle the two other points and differ from the method presented in this paper.} 

{A special yet very common case of {\bf non-homogeneous cardinality} is some missing values. As the data may be missing for different reasons, there exists different ways of handling them that we briefly present in Section \ref{sec:data_wrangling}. However, none of them is as straightforward as having a classification method that works for non-homogeneous cardinality.}

{Regarding {\bf representation agnosticity}, the representation is very often imposed by the classification method itself and when the dataset does not respect those requirements, it has to be transformed.} Of course, exploiting the specificities of features often leads to better results but also requires manual expertise that blocks a wider adoption by end-users.

\subsubsection{Metric learning}
\label{sec:metric_learning}

{Choosing an appropriate pairwise metric to measure distance between two points is crucial in the success of classification algorithms \cite{davis2007information}. Learning a metric consists in finding a projection $f$ from an initial space to a Euclidian space s.t. for any elements $\mathbf x$ and $\mathbf x'$, $d(\mathbf{x}, \mathbf{x}') = ||f(\mathbf{x}) - f(\mathbf{x}')||$. The metric should reflect the semantic difference between objects. If the elements are non-numerical, a metric can be viewed as a proxy to represent the elements in a vector space s.t. it becomes possible to apply standard classification methods. However, as reported by \cite{Bellet2013ASO}, the literature mostly focused on numerical data, i.e. when the elements already belong to a vector space.}

The most common setting for metric learning is to find a parametrization $M$ of the Mahalanobis pseudometric $d_M(\mathbf x, \mathbf x') = \sqrt{(x - x')^T M (x - x')}$ using the training set $\mathbf X$ and under the constraint that $M$ is Positive Semi-Definite (PSD). {Using $M$ as the identity recovers the usual Euclidian metric, whereas setting $M$ as the covariance matrix of $\mathbf X$ gives the original Mahalanobis distance \cite{malahanobis}}. Most methods aim at finding the parameters that best agree with the following three sets of relations on the examples:
\begin{itemize}
\item $\mathcal{S} = \{ (\mathbf x_i, \mathbf x_j) ~:~ \mathbf x_i \text{ and } \mathbf x_j \text{ are similar} \}$, \hfill (must-link)
\item $\mathcal{D} = \{ (\mathbf x_i, \mathbf x_j) ~:~ \mathbf x_i \text{ and } \mathbf x_j \text{ are dissimilar} \}$, \hfill (cannot-link)
\item $\mathcal{R} = \{ (\mathbf x_i, \mathbf x_j, \mathbf x_k) ~:~ \mathbf x_i \text{ more similar to } \mathbf x_j \text{ than to } \mathbf x_k \}$. \hfill(relative)
\end{itemize}
However, supervised methods usually derived those sets (implicitly or explicitly) from the examples and a proper notion of neighborhood.

{We now present the approaches with similarities to the one introduced in this paper. For more detailed surveys on metric learning, we refer the reader to \cite{Bellet2013ASO,Wang:2015:SDM:2738484.2738549}.}

{\sc LMNN} \cite{weinberger2006distance,weinberger2009distance} is one the most popular Mahalanobis metric learning techniques and many other linear methods are based on it. The constraint sets $\mathcal S$ and $\mathcal R$ are defined by a notion of neighborhood:
\begin{itemize}
\item $\mathcal{S} = \{ (\mathbf x_i, \mathbf x_j) ~:~ y_i = y_j \text{ and } \mathbf x_j \text{ belongs to the $k$-nearest neighbors of } \mathbf x_i \}$,
\item $\mathcal{R} = \{ (\mathbf x_i, \mathbf x_j, \mathbf x_k) ~:~ (\mathbf x_i, \mathbf x_j) \in \mathcal S, ~ y_i \neq y_k \}$.
\end{itemize}
In the original work, the neighborhood is defined using the Euclidian distance, and thus, assumes the elements lives in a vector space. {To avoid this, the method presented in this paper defines the neighborhood based on set intersections to derive $\mathcal{D}$ and $\mathcal{R}$.}
The PSD matrix $W$ is found by solving the following convex optimization problem:
\begin{nalign}
M^* = & ~ \underset{M \in \mathbb{S}^n_+}{\argmin} ~ (1 - \mu) \underset{(\mathbf x_i, \mathbf x_j) \in \mathcal S}{\sum} d^2_{M}(\mathbf x_i, \mathbf x_j) + \mu \underset{i,j,k}{\sum} \varepsilon_{i,j,k}\\  
       & \text{s.t.} ~ d^2_M(\mathbf x_i, \mathbf x_k) - d^2_M(\mathbf x_k, \mathbf x_j) \geq 1 - \varepsilon_{i,j,k}, \hfill \forall (\mathbf x_i, \mathbf x_j, \mathbf x_k) \in \mathcal R\\
       & \varepsilon_{i,j,k} > 0
\end{nalign} with $\mu \in [0,1]$ a tradeoff parameter, and $\varepsilon_{i,j,k}$ some slack variables.

{\sc OASIS} \cite{Chechik:2010:LSO:1756006.1756042} learns a bilinear similarity metric of the form $d_M(\mathbf x, \mathbf x') = \mathbf x^T M \mathbf x'$ without the PSD constraint on $M$. It can define similarity between instances of different dimensions such as Bag-of-Words. Relaxing the PSD constraint allows the author to use an efficient online Passive-Aggressive algorithm \cite{crammer2006online} to solve for $M$:
\begin{nalign}
M^t = & ~ \underset{M,~ \varepsilon}{\argmin} ~  \frac 1 2 ||M - M^{t-1}||^2_{\mathcal F} + C\varepsilon\\  
       & \text{s.t.} ~ 1 - d^2_M(\mathbf x_i, \mathbf x_j) + d^2_M(\mathbf x_i, \mathbf x_k) \leq \varepsilon\\
       & \varepsilon > 0
\end{nalign} with $C$ a regularization factor and $||.||_{\mathcal F}$ the Frobenius norm. {Similarily, \HCBR~works with input vectors of different dimensions. The iterative form of the optimization program is close to the training phase introduced in Section \ref{sec:decision_training}}

{\sc SLLC} \cite{bellet:hal-00708401} focuses on learning a bilinear similarity matrix as a quadratic constraint program defined by
\begin{nalign} \label{sllc}
\underset{M \in \mathbb{R}^{n \times n}}{\min} ~ \frac 1 n \underset{i=1}{\overset{n}{\sum}} \ell_{\text{hinge}}(1 - y_i \frac 1 {\gamma |S|} \underset{\mathbf x_j \in \mathcal D}{\sum} y_j d_M(\mathbf x_i, \mathbf x_j) ) + \beta||M||^2_{\mathcal F} 
\end{nalign} where $\mathcal D$ is a sample from the training set, $\gamma$ a margin parameter and $\beta$ a regularization parameter. The objective of SLLS is to find a metric s.t. the elements of one class are in average more similar than the elements of the other class by a margin $\gamma$. {The idea behind \HCBR~is similar except that it learns one seminorm per class such that we can impose to calibrate the distance (i.e. the distance is a confidence measure) over the training set rather than having an average distance parameter. Also, the seminorms are parametrized by a vector and not a matrix.}

{{\sc MMDA} \cite{kocsor2004margin} learns the Euclidian distance between $W^T \mathbf{x}_i$ and $W^T \mathbf{x}_j$ and thus is a linear method.} To do so, it learns $k$ projection hyperplanes $\{ \mathbf w_r \}_{r=1}^{k}$ by solving the following optimization program:
\begin{nalign}
 & \underset{W, \mathbf b, \mathbf{\varepsilon}_r}{\min} ~  \frac 1 2 \underset{r = 1}{\overset{k}{\sum}} ||\mathbf{w}_r||^2_2 + \frac C n \underset{r = 1}{\overset{k}{\sum}} \underset{i = 1}{\overset{n}{\sum}} {\varepsilon}_{ri}\\  
 & \text{s.t.} ~ \ell_{\text{hinge}}(\mathbf{w}_r^T\mathbf{x}_i + b_i) \leq 1 - {\varepsilon}_{ri}\\
 & \varepsilon_{ri} > 0\\
 & W^T W = Id
\end{nalign} where $\mathbf{\varepsilon}_r$ are slack variables to penalize margins and $C$ a regularization parameter. {\HCBR~adopts the opposite direction: it gives explicitly a projection per $\mathbf x_i$ and tries to find a vector $\mu$ such that the sum of hinge loss $\ell_{\text{hinge}} (\mathbf{w}_i^T \mu)$ is minimized over the training set.}

{{\sc LSMD} \cite{chopra2005learning} learns a non-linear projection $G_W$ parametrized by $W$ s.t. $d(\mathbf x, \mathbf x') = || G_W(\mathbf x) - G_W(\mathbf x')||_1$ is small when the $\mathbf x$ and $\mathbf x'$ are in the same class and large otherwise. The parameter $W$ represents the weights of a neural network used to learn the projections.} {\HCBR~also learns a non-linear projection but such that the sign indicates the class and the distance to 0 the confidence in this class. Therefore, a small distance between two elements does not indicate the same class but the same level of confidence.}

{The method proposed in this paper is not a pairwise metric learning method. However, many similarities exist as it needs a proxy to express non-numerical data in a numerical decision space s.t. a simple decision rule well performs. What is learnt is a seminorm per class, representing the distribution of discriminative information to support this class w.r.t. the interactions between the training set examples. 
The pairwise metric can be artificially defined using the pseudo-norms but is never directly used in the current state of the method. }

{According to the usual taxonomy of metric learning methods \cite{Bellet2013ASO,Wang:2015:SDM:2738484.2738549}, \HCBR ~ can be viewed as a fully-supervised, non-linear and global\footnote{In this paper, we use the notion of locality in a different meaning than in the metric learning community. The locality of our model expresses the fact the model cannot be used outside of a certain neighborhood, but the metric has been defined using the entire training set.} metric learner algorithm dedicated to classification on non-vector space. The dimensionality reduction is implicitly defined by the hypergraph representation. The literature work on non-numerical data mostly focused on distance between strings \cite{oncina:hal-00114106} and trees \cite{Dalvi:2009:RWE:1559845.1559882,bernard2006learning}, with a problem of combinatorial explosion. The scalability study in Section \ref{sec:complexity} shows this is not the case with \HCBR.}

\subsubsection{Data wrangling and missing data}
\label{sec:data_wrangling}

Data wrangling consists in selecting, transforming and curating data { that are unstructured or not suitable for the selected algorithm}. It is commonly accepted that up to 80\% of data scientists time is spent on data wrangling \cite{chessell2014governing}. On top of the time consumption, the whole preprocessing step has a huge impact on the final model quality. For instance, in \cite{CRONE2006781}, the authors showed that the accuracy obtained by Neural Network, SVM and Decision Trees are significantly impacted by data scaling, sampling and encoding. For a more comprehensive view on data processing impact, we refer the reader to \cite{dasu2003exploratory}.

Data wrangling offers several challenges notably dealing with missing data and outliers or combining data from multiple sources in an automated way \cite{https://doi.org/10.5441/002/edbt.2016.44,Kandel:2011:RDD:2336221.2336223}. {In this section, we present the work related to those challenges. However, }{we believe that another way to tackle data wrangling challenges is to design models that are less sensitive to data preprocessing, missing data or data representation. Not only it decreases the time allocation needed to create efficient data pipelines, but it also decreases the CPU time by shrinking the data pipeline itself. For instance, the method proposed in this paper does not need to remove outliers because they will be marginalized by the model selection itself. As \HCBR~works with any unstructured space, it becomes easy to combine data from multiple sources. Despite intelligent data transformation might increase the overall performances, we show in Section \ref{sec:robustness} that it is not a requirement to perform well.}

Modeling datasets in a latent space using meta-features allows to predict the impact of preprocessing operators on the model accuracy \cite{Bilalli:2017:PPM:3214035.3214049}. To merge multiple data sources semi-automated tools exists such a Clio \cite{haas2005clio} or Schema Mapper \cite{robertson2005visualization}. For knowledge discovery, SeeDB \cite{vartak2015s} automatically generates useful visualization of data relations {that can help to perform feature selection.} 
{Recently, \cite{DBLP:conf/dolap/Quemy19} proposed a fully automated way of building data pipelines using standard hyperparameter tuning techniques. The approach still involves a large CPU time overhead.} 

{The most common technique to deal with missing value is replacing the value (imputation) using the mode, average or a value obtained by a model, over the available data. The imputation might be done using a $k$-nearest neighbors algorithm \cite{batista2002study}, local regression \cite{preda2010nipals} or Random Forest \cite{stekhoven2011missforest}. In general, machine learning approaches tend to outperform traditional statistical methods \cite{kuhn2013applied}. To handle values under missing at random (MAR) settings, one can use multiple imputations \cite{royston2004multiple,van2018flexible}. However, this technique requires to generate several variations of the training set and build several models that are then averaged s.t. once again, it requires a large computational effort.}

{Even when values are missing completely at random (MCAR), there are scalability issues with practical machine learning. Assume some values are MCAR (e.g. due to a defective sensor for some time), or that new pieces of information are available after an initial model was built (e.g. adding new sensors). In the case of linear classification, e.g. in $\mathbb{R}^3$, if an element has only 2 components it does not describe a point but a plane and as a result, there is no way to separate it with a (hyper)plane.}

{Conversely, consider a model built in $\mathbb{R}^2$ with some additional information available afterward, s.t. the classification instance now lives in $\mathbb{R}^3$. Mathematically, there is no guarantee that the model in two dimensions is even close to the model built from scratch in three dimensions. For instance, if the points are not linearly separable in two dimensions but are in three dimensions, there is no chance for the projection of the plane into the subspace of two dimensions to be the model found while working only in this subspace. Despite multiple algorithms have an {\it online} counterpart, as far as we know, there is no significant work on having this {\it horizontal} scalability (as opposed to vertical scalability). One step toward horizontal scalability is to work with unstructured spaces as proposed in this paper.}

\subsection{Classification and hypergraph}
\label{sec:state_of_art}

Hypergraphs generalize graphs and can be used to represent higher order relations while graphs are limited to binary relations. A hypergraph is defined by a set of vertices and a collection of hyperedges where each hyperedge is a subset of this set of vertices. Therefore, a graph is a special case of hypergraph for which each hyperedge contains only two vertices. We will formally introduce hypergraphs in Section \ref{sec:rep}.
Recently hypergraphs have been used as data representation, and some classification algorithms on hypergraph have been proposed. A vast majority of approaches models the objects to classify as the set of vertices and constructs the hyperedges as the representation of a metric. This conventional approach is known as {\it neighborhood-based} hypergraph. The metric relies on some assumptions on the data or a specific representation (e.g. Zernike moment and histogram of oriented gradient to measure the distance between images in \cite{ijcai2017-387}) and for each vertex, a hyperedge is created to represent its $k$-nearest neighbors \cite{5540012}. 

The problem of classification on hypergraph consists in labeling some unlabeled vertices given a training set such that all vertices in a same hyperedge have the same label. As all the vertices are known a priori, the problem is part of transductive learning. To learn the labels, the standard approach is to minimize a cost function based on a hypergraph equivalent of a graph Laplacian \cite{ijcai2017-387,NIPS2006_3128} with a structural risk:
\begin{nalign}
C(\mathbf{x}) = \mathbf{x}^t \Delta \mathbf{x} + \mu ||\mathbf{x} - \mathbf{y}||^2
\end{nalign} where $\Delta$ is the hypergraph Laplacian, $\mu >0$ a regularization factor and $||.||$ a norm. The vector $\mathbf{y}$ represents the initial labels for all vertices with $y_i = 0$ for unlabeled vertices, a negative (resp. positive) value for label -1 or 1.

On the contrary, the {\bf method proposed in this paper} models the elements to classify as the hyperedges and the vertices as the different components of those elements. As far as we know, there is no previous work that uses this modeling choice. In addition, it does not require knowing all the elements before building the model: our approach is inductive. 
More important, as most previous work consists in building metrics based on the feature representation, it obviously conflicts with our goal of agnosticity described in the previous section.

\subsection{Binary classification in unstructured spaces}
\label{sec:unstructured_spaces_problem}

With all the considerations of Section \ref{sec:bc_unstructured} in mind, we reformulate the problem of binary classification for unstructured space. The only {\it ad-hoc} operation we allow is checking if a particular feature belongs to the element to classify, and by extension we allow elementary set operations. We consider a countable space $\mathbb{F}$ and define $\mathcal X = 2^{\mathbb{F}}$. By abuse of notation, we call  $\mathbf{x} \in \mathcal X$ an ``vector'' despite $\mathcal X$ is not a vector space. We also refer to $\mathbf{x}$ as a {\it case}. In practice, it is very likely that only a subset of $2^{\mathbb{F}}$ may appear (for instance if two features encode two contradictory propositions or if every case has the same number of features).
The real class for any input vector $\mathbf{x}$ of $2^{\mathbb{F}}$ is given by the mapping:
\begin{nalign}
    \label{assumption_mapping}
   J \colon 2^{\mathbb{F}} & \to \{-1,1 \} \\
   \mathbf{x} &\mapsto J(\mathbf{x})
\end{nalign}
Assuming the unknown mapping takes value from the powerset of $\mathbb{F}$ allows us not to have to know $\mathbb{F}$ at all. {Concretely, the elements of $\mathbb{F}$ can take different forms depending on the data: a pair ``variable=value", a word, a proposition that is true or false, etc. The only minimal requirement is that there exists a way to index the data elements. In the experimental validation performed in Section \ref{sec:experiments}, the datasets are structured (with some missing values) and thus, the features are all represented by a pair ``variable=value". In Section \ref{sec:unstructured_datasets}, the features are represented by ``word=occurrences" if a word appears at least once. However, as $\mathbb{F}$ can be seen as a subset of $\mathbb{N}$, the method presented below would work on datasets mixing representations (descriptive features, Bag-of-Words, etc.).}

Notice that in this paper we do not consider {\it uncertainty}: if two situations are described the same in $\mathbb{F}$, then they have the same label.

\section{Hypergraph Case-Based Reasoning} 
\label{sec:hcbr}

In this section, we introduce our main contribution with {\bf a new framework for binary classification in unstructured spaces} called Hypergraph Case-Based Reasoning (\HCBR). The presentation is broken down into five steps. First, we present in Section \ref{sec:rep} how to represent our training set as a hypergraph and how to project any element of $2^{\mathbb F}$ onto it. Those elements allow us to formally define the model space in Section \ref{sec:model}. Section \ref{sec:model_selection} is dedicated to parameter estimations while Section \ref{sec:decision_training} focuses on the training. Finally, Section \ref{sec:dec_rule} is dedicated to the decision rule refinement and hyperparameters. Section \ref{sec:complexity} provides the time complexity of the  main phases of the algorithm. 

\subsection{Representation and projection}
\label{sec:rep}

Before defining the projection operator used by \HCBR~ to make predictions, we recall the definition of a hypergraph. For additional results on hypergraphs, we refer the reader to \cite{berge1984hypergraphs}.

\begin{definition}[Hypergraph]
A hypergraph is defined by $H = (V, \mathbf{X})$ with $V$ a set of vertices, $\mathbf{X}$ the hyperedges s.t. $\forall \mathbf{x} \in \mathbf{X}, ~ \mathbf{x} \subseteq V$.
\end{definition}
\noindent
A hypergraph can be viewed as a collection of subsets $\mathbf{X}$ of a given set of vertices $V$. It is sometimes convenient to define a hypergraph solely by a collection of sets. In this case, the set of vertices, denoted $V_\mathbf{X}$, is implicitly defined as the union of edges.

\begin{figure}[!h]
  \centering
  \def\firstcircle{(0,0) ellipse (1.5cm and 1cm)}
  \def\secondcircle{(80:-1cm) ellipse (1.5cm and 1cm)}
  \def\thirdcircle{(0:2cm) ellipse (1.5cm and 1cm)}
  \begin{tikzpicture}[scale=0.8, every node/.style={scale=1}]
      \begin{scope}[fill opacity=0.3]
          \filldraw[fill=gray, xscale=1.4, yscale=1.4] (-0.0,-0.0) plot [smooth cycle,tension=0.7, shift={(-4,-3.2)}] coordinates {(3,1) (5,1.2) (7,1) (8,3) (7,4.5) (5,4.5) (2,4) (1.7,2.5)};
          \fill[red, rotate=30, xscale=1.4, yscale=1.4] \firstcircle;
          \fill[green, rotate=35, xscale=1.4, yscale=1.4] \secondcircle;
          \fill[blue, xscale=1.4, yscale=1.4] \thirdcircle;
          \draw[rotate=30, xscale=1.4, yscale=1.4] \firstcircle;
          \draw[rotate=35, xscale=1.4, yscale=1.4] \secondcircle;
          \draw[xscale=1.4, yscale=1.4] \thirdcircle;

      \end{scope}
      \node at (4,-2.7) {$\mathbb{F}$};
      \node at (0.5,1) {$x_1$};
      \node at (0,-2.3) {$x_2$};
      \node at (3,1) {$x_3$};

      \node at (-0.8  ,0.5) {$e_1$};
      \node at (1.4  ,0.7) {$e_2$};
      \node at (1.2  ,-0.3) {$e_3$};
      \node at (3.6  , -0.1) {$e_4$};
      \node at (2  , -0.7) {$e_5$};
      \node at (-0.4  ,-1) {$e_6$};
      \node at (1  ,-1.7) {$e_7$};
  \end{tikzpicture}
  \caption{\label{case_base} The family $\mathcal E = \{ \mathbf e_i \}_{i}$ forms the partition obtained by the union of the projection of cases and represents how the three cases share information. }
\end{figure}
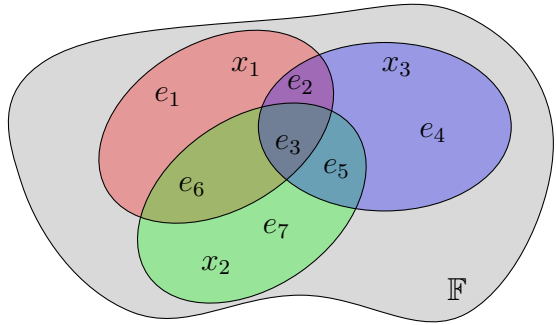

A training set $\mathbf{X}$ can be seen as a hypergraph $H = (\mathbb F, \mathbf{X})$, i.e. such that each example is a hyperedge (Figure \ref{case_base}). In practice, we do not need to know $\mathbb{F}$ as we can always use the implicit hypergraph $H = (\mathbb{F}_{\mathbf X}, \mathbf{X})$ where $\mathbb{F}_\mathbf{X}$ is the restriction of $\mathbb{F}$ to the features present in the sample $\mathbf{X}$. For structured datasets, the elements of $\mathbb{F}_\mathbf{X}$ are all the distinct pairs ``variable=value" in $\mathbf{X}$.
For any hypergraph $H = (\mathbb{F}_{\mathbf X}, \mathbf{X})$, there exists a unique partition $\mathcal{E}_H = \{\mathbf{e_i}\}^M_{i=1}, ~\forall 1 \leq i \leq M, ~ \mathbf{e_i}\subseteq \mathbb{F}_{\mathbf X}$ defined by the intersections of its edges as illustrated by Figure \ref{case_base}.

The projection of a case over a hypergraph returns the elements of $\mathcal{E}_H$ it intersects with.
\begin{definition}[Projection over a hypergraph]
The projection operator $\pi_H$ over a hypergraph $H = (V, \mathbf{X})$ for any $A \subseteq V$ is defined by $\pi_H(A) = \{ \mathbf{e} \in \mathcal E_H ~ | ~ \mathbf{e} \cap \mathbf{x} \neq \emptyset \}$.
\end{definition}

\begin{figure}[!h]
  \def\firstcircle{(0,0) ellipse (1.5cm and 1cm)}
  \def\secondcircle{(80:-1cm) ellipse (1.5cm and 1cm)}
  \def\thirdcircle{(0:2cm) ellipse (1.5cm and 1cm)}
  \begin{tikzpicture}[scale=0.7, every node/.style={scale=0.7}]
      \begin{scope}[fill opacity=0.3]
          \filldraw[fill=gray, xscale=1.4, yscale=1.4] (-3.5,-3.5) plot [smooth cycle,tension=0.7, shift={(-4,-3.2)}] coordinates {(3,1) (5,1.2) (7,1) (8,3) (7,4.5) (5,4.5) (2,4) (1.7,2.5)};
          \fill[red, rotate=30, xscale=1.4, yscale=1.4] \firstcircle;
          \fill[green, rotate=35, xscale=1.4, yscale=1.4] \secondcircle;
          \fill[blue, xscale=1.4, yscale=1.4] \thirdcircle;
          \draw[rotate=30, xscale=1.4, yscale=1.4] \firstcircle;
          \draw[rotate=35, xscale=1.4, yscale=1.4] \secondcircle;
          \draw[xscale=1.4, yscale=1.4] \thirdcircle;

          \fill[yellow, rotate=110, xscale=1, yscale=1, xshift=-10, yshift=15] \secondcircle;
          \draw[rotate=110, xscale=1, yscale=1, xshift=-10, yshift=15] \secondcircle;

      \end{scope}
      \node at (4,-2.7) {$\mathbb{F}$};
      \node at (0.5,1.3) {$x_1$};
      \node at (0,-2.3) {$x_2$};
      \node at (3,1) {$x_3$};

      \node at (-0.8  ,0.5) {$e_1$};
      \node at (0.2  ,0.6) {$e_1'$};
      \node at (1.5  ,0.7) {$e_2$};
      \node at (1.05  ,0.5) {$e_2'$};
      \node at (1.2  ,-0.2) {$e_3'$};
      \node at (1.675  , 0.15) {$e_3$};
      \node at (3.6  , -0.1) {$e_4$};
      \node at (2  , -0.7) {$e_5$};
      \node at (1.5 , -0.8) {$e_5'$};
      \node at (-0.6  ,-1.2) {$e_6$};
      \node at (0.3  ,-0.7) {$e_6'$};
      \node at (1  ,-1.4) {$e_7'$};
      \node at (0.5  ,-2) {$e_7$};
  \end{tikzpicture}
  \hfill
  \begin{tikzpicture}[auto, thick, scale=0.8, every node/.style={scale=0.8}, baseline={(2,-4.5)}]
    \node[superpeers, fill=red!30] (c1) at  (0,-4) {$x_1$};
    \node[superpeers, fill=blue!30] (c2) at  (2,-4) {$x_2$};
    \node[superpeers, fill=green!30] (c3) at  (-2,-4) {$x_3$};
    \node[superpeers, fill=yellow!30] (c4) at (0,0) {$x$};
     \node[peers, fill=black!10, label=above:{$e_2$}] (e2) at (-2.5,-2) {};
     \node[peers, fill=black!10, label={[xshift=0.05cm, yshift=-0.05cm]{$e_3$}}] (e3) at (-1.5,-2) {};
     \node[peers, fill=black!10, label={[xshift=-0.1cm, yshift=-0.05cm]{$e_5$}}] (e5) at (-0.5,-2) {};
     \node[peers, fill=black!10, label={[xshift=0.1cm, yshift=-0.1cm]{$e_1$}}] (e1) at (0.5,-2) {};
     \node[peers, fill=black!10, label={[xshift=-0.1cm, yshift=-0.1cm]{$e_6$}}] (e6) at (1.5,-2) {};
     \node[peers, fill=black!10, label=above:{$e_7$}] (e7) at (2.5, -2) {};

     \path[<-] (c4) edge (e1);
     \path[<-] (c4) edge (e2);
     \path[<-] (c4) edge (e3);
     \path[<-] (c4) edge (e5);
     \path[<-] (c4) edge (e6);
     \path[<-] (c4) edge (e7);

     \path[->] (c1) edge (e1);
     \path[->] (c1) edge (e2);
     \path[->] (c1) edge (e3);
     \path[->] (c1) edge (e6);

     \path[->] (c2) edge (e3);
     \path[->] (c2) edge (e5);
     \path[->] (c2) edge (e6);
     \path[->] (c2) edge (e7);

     \path[->] (c3) edge (e2);
     \path[->] (c3) edge (e3);
     \path[->] (c3) edge (e5);

  \end{tikzpicture}
  \caption{\label{new_case_schema} The projection of $\mathbf{x}$ (in yellow) on $H$ is the set of $\mathbf e_i$ intersecting with it. On the right, the graph represents the projection elements $\{ \mathbf e_i\}_i$ and their respective connections to the cases $\{\mathbf x_i\}_i$, in particular, $D_{\mathbf x} = \emptyset$.}
\end{figure}
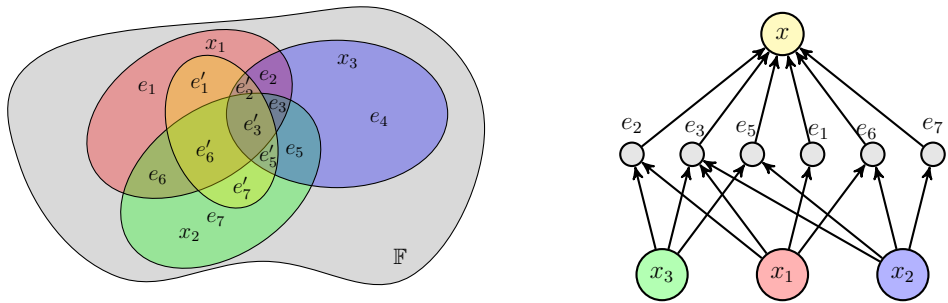

\noindent
{\bf Example 1:} Each element of $\pi_H(\mathbf{x})$ is a (sub)set of features. For instance, in Figure \ref{case_base}, $\pi_H(\mathbf{x_1}) = \{ \mathbf{e_1}, \mathbf{e_2}, \mathbf{e_3}, \mathbf{e_6} \}$ and in Figure \ref{new_case_schema}, the projection of $\mathbf{x}$ (in yellow) on $H$ is the whole partition $\mathcal E_H$ as $\mathbf{x}$ intersects with every $\mathbf e \in \mathcal E_H$. \\

We call {\it discretionary features} of $\mathbf{x}$ (w.r.t. $H$) the (possibly empty) set of features that are not in $V_{\mathbf{X}}$, denoted $D_\mathbf{x}$. It can be interpreted as the features of $\mathbf{x}$ that do not belong to any hyperedge. When adding hyperedges, the discretionary features of $\mathbf x$ are new pieces of information that were never encountered before. \\

\noindent
{\bf Example:} Considering the hypergraph composed of $\mathbf{x_1}$ and $\mathbf{x_2}$ as illustrated by Figure \ref{case_base}, the set of discretionary features of $\mathbf{x_3}$ is $\mathbf{e_4}$. In Figure \ref{new_case_schema}, the yellow case $\mathbf{x}$ has no discretionary feature: all its features are present at least in one example.\\

For any set of features $\mathbf{x} \subseteq \mathbb F$, we define $d_H(\mathbf{x}) = \{ \mathbf{x}' \in \mathbf X ~ | ~ \mathbf{x} \cap \mathbf{x}' \neq \emptyset \}$ the set of edges sharing some features with $\mathbf{x}$. In particular, the set $d_H$ can be split into $d_H^{(1)}$ and $d_H^{(-1)}$ depending on the label of the case, and defined by $d_H^{(l)}(\mathbf{x}) = \{\mathbf{x}' \in d_H(\mathbf{x}) ~ | ~ J(\mathbf{x}') = l \}$. 
Note that if $\mathbf{x} \not \in \mathbf{X}$ and $|d_H(\mathbf{x})| = 0$, then $\mathbf{x} = D_\mathbf{x}$, i.e. the case $\mathbf{x}$ is in relation with no case in $\mathbf{X}$. In the hypergraph literature, $|d(\mathbf{x})|$ is called the {\it degree} of a hyperedge and its domain of definition is restricted to $\mathbf{X}$ while here, it is extended to $2^{\mathbb F}$. Finally, to use matricial notations, we define the vectors $\mathbf d^{l} = (|d^{(l)}(\mathbf x_i)|)^N_{i=1}$.

From now, we consider only the specific hypergraph generated by the set of examples $\mathbf{X}$. For the sake of readability, we remove the subscript $H$.

\subsection{Model space}
\label{sec:model}
As discussed in Section \ref{sec:bc_unstructured}, we relaxed some implicit constraints on the input vector space.
As a counterpart, \HCBR~ relies on the assumption that if two input vectors $\mathbf{x}$ and $\mathbf{x}'$ do not share any feature, they are {\it independent} i.e. $\mathbf{x}$ cannot help to understand the correct class of $\mathbf{x}'$ and vice versa. This limitation comes from the fact there is no metric on $\mathbb F$ to determine a distance between two elements s.t. we can rely only on intersections. As a result, \HCBR~  produces only {\it local} models because if a new input vector is independent of all examples, it is impossible to generate a prediction. On the contrary, a hyperplane model is {\it global} in a sense that it can produce a prediction for any element of $\mathbb{R}^M$. {We discuss and propose a solution to this issue in Section \ref{sec:model_locality}.}

An example of concrete situation for which such assumption is natural is a justice trial. Cases are composed of some elements, and the correct label is the result of a reasoning that can possibly use analogies or counter-examples with a set of past cases on top of the legal texts. However, if a judge or lawyer wants to use $\mathbf{x}$ to justify the outcome of $\mathbf{x}'$, $\mathbf{x}'$ must have similarities with $\mathbf{x}$. 

First, we give the {\bf intuition} behind our model space. For each element $\mathbf e_i$ in the projection, we value 1) its importance in the case $\mathbf x$ and 2) its support toward a specific class given the whole hypergraph. Intuitively, 1) consists in answering the question: {\it how important is $\mathbf e_i$ w.r.t. $\mathbf x$?} or {\it what is the potential for analogies or counter-examples between $\mathbf x$ and the other examples also containing $\mathbf e_i$?} The purpose of 2) is to value how important is $\mathbf e_i$ with regards to the other elements of $\mathcal E$. For instance, to explain the case $\mathbf x_1$ in Figure \ref{case_base}, we will use only the elements of the projection $\pi_H(\mathbf{x_1}) = \{ \mathbf{e_1}, \mathbf{e_2}, \mathbf{e_3}, \mathbf{e_6} \}$. Without prior information, we will use the size of $\mathbf e_i$ in $\mathbf x$ to measure their importance $\mathbf x$.
The way of measuring the support with regard to the whole hypergraph as well as an intuitive interpretation will be given in Section \ref{sec:model_selection}.

Let us now {\bf formally} define the model space. Given the hypergraph $H = (\mathbb F, \mathbf{X})$ defined by a training set $\mathbf{X}$ and its associated partition $\mathcal{E} = \{\mathbf{e_i}\}_{i=1}^m$, the relation between an example $\mathbf{x}$ and its class is modeled by
\begin{align}
\label{eqn:model}
\left\{\begin{matrix}
s_{w, \mu}(\mathbf{x}_j) & = & &\underset{i = 1}{\overset{M}\sum} w(\mathbf{e}_i, \mathbf{x}_j) \mu(\mathbf{e}_i, \mathbf{X}, \mathbf{y})
 \\
\underset{i = 1}{\overset{M}\sum} w(\mathbf{e}_i, \mathbf{x}_j) & = & ~ 1 & ~ \hfill \forall 1 \leq j \leq N \hfill\\
\underset{i = 1}{\overset{M}\sum} \mu(\mathbf{e}_i, \mathbf{X}, \mathbf{y}) & = & ~ 1 \hfill
\end{matrix}\right.
\end{align} where $w(\mathbf{e}_i, \mathbf{x}_j) \geq 0$ models the importance of $\mathbf{e}_i$ in $\mathbf{x}_j$ and $\mu(\mathbf{e}_i, \mathbf{X}, \mathbf{y})$ the support of $\mathbf{e}_i$ for class 1 w.r.t. whole training set. For this reason, we call $\mu$ the {\it intrinsic strength} of $\mathbf e_i$. The above-mentioned assumption implies that if $\mathbf{e}_i \cap \mathbf{x} = \emptyset$ then $w(\mathbf{e}_i, \mathbf{x}) = 0$. For readability, we write $s$ in place of $s_{w, \mu}$.

The classification rule consists in selecting the class with the total highest support:
\begin{align} \tag{R1} \label{eqn:decision_rule}
 \bar J_{w, \mu}(x) =  \left\{\begin{matrix}
1 & ~s_{w, \mu}(\mathbf x) > 0\\ 
-1 & ~s_{w, \mu}(\mathbf x) \leq  0
\end{matrix}\right.
\end{align} 
The classification problem consists then in finding the couple $(w, \mu)$ that minimizes the classification error:
\begin{align}
\label{eqn:model_opt}
(w^*, \mu^*) = \underset{w, \mu}{\argmin} ~ \sum_{ (\mathbf{x}, y) \in (\mathbf{X}, \mathbf{y})} \mathbb{I}_{\{y \neq \bar J(\mathbf{x})\}}
\end{align}
The problem described by (\ref{eqn:model_opt}) is a functional problem, in general much harder than parametric ones such as (\ref{eqn:param_model}), and without restrictions on the search space it may not be reasonable to look for a solution. For this reason, and for the rest of this paper, we will fix $w$ a priori. As we do not formulate any particular assumption on the feature space and allow only basic set operations, a natural yet trivial way to model the importance of $\mathbf{e}_i$ in $\mathbf{x}_j$ is to set $w(\mathbf{e}_i, \mathbf{x}_j) = \frac{|\mathbf{x}_j \cap \mathbf{e}_i|}{|\mathbf{x}_j|}$. However, if one has some information about the impact of some features in some particular cases, $w$ might be redefined to integrate this prior knowledge.

Once the method to calculate $w$ and $\mu$ is specified,
the model can be expressed matricially by
\begin{align}
\label{eqn:model_mat}
\left\{\begin{matrix}
\mathbf s & = & W \mu &~ W \in \mathcal{M}(\mathbb{R})_{N \times M}, ~ \mu \in \mathbb{R}^M \\
||\mathbf{w}_{j:}||_1 & = & 1 & ~ \forall 1 \leq j \leq N \hfill\\
||\mu||_1 & = & 1  \hfill
\end{matrix}\right.
\end{align}
{By setting $d_\mu(\mathbf x, \mathbf x') = |s(x) - s(x')|$, we can see the problem as learning a metric parametrized by $\mu$ s.t. the training set elements are properly classified.
Rather than learning directly $\mu$, we split it into $\mu^{(1)}$ and $\mu^{(-1)}$, representing the distribution of the support toward class -1 and 1 over $\mathcal E$. }

To fit a model capable of generalization, it is not enough to simply select $\mu$ s.t. it minimizes the classification error. Assuming $W$ is fixed and $\mathbf s$ set to some arbitrary values s.t. each example is correctly classified, $\mu$ can be obtained by using the Moore-Penrose pseudo-inverse of $W$. {The Moore-Penrose pseudo-inverse generalizes the inverse for non-square matrix. It always exists and is unique. Depending on $\mathbf s$ and $W$, it might be impossible to classify correctly all the elements using $\mu = W^+ \mathbf s$. However, by construction of the Moore-Penrose pseudo-inverse, $W^+\mathbf s$ is a least-square minimizer of $\mathbf s = W\mu$. If this system has many solutions, $W^+\mathbf s$ is the solution with the lowest norm $||.||_2$.}

 By artificially setting $\mathbf s$ to obtain $\mu$ with the Moore-Penrose inverse, the constructed model holds no discriminative information and there is no chance it can provide good results on new cases. For this reason, $\mu$ has to be designed such that $\mathbf s$ is {\it calibrated}, i.e. the higher $\mathbf s$ is, the more confident the model is about the prediction. Ideally, among all the elements with e.g. a (normalized) strength close to 0.8, 80\% of them should be correctly classified. 

To summarize, after defining a general model space to solve the binary classification problem in unstructured space, we made additional assumptions to reduce the problem of model selection to find a vector $\mu$ s.t. the classification rule \eqref{eqn:decision_rule} provides a good solution to the original problem \eqref{eq:pb}. For this purpose, we will proceed in three steps:

  \begin{itemize}
  \item {\bf Step 1:} Define $\mu$ to capture as much information as possible from $\mathcal E$ for any $\mathbf X$ (Section \ref{sec:model_selection}).
  \item {\bf Step 2:} Train the model to adjust $\mu$ on a specific $\mathbf X$ using the classification rule \eqref{eqn:decision_rule} (Section \ref{sec:decision_training}).
  \item {\bf Step 3:} Refine the classification rule \eqref{eqn:decision_rule} to take into account the local nature of the model (Section \ref{sec:dec_rule}). {The refined classification rule \eqref{eqn:updated_cr} is presented in Section \ref{sec:dec_rule}.}
  \end{itemize} A summary and high level view of \HCBR~ is given by Algorithm~\ref{algo:HCBR}.

\algrenewcommand\algorithmicindent{1.0em}%
\begin{algorithm}
  \caption{HCBR (High level view)}\label{algo:HCBR}
  \begin{algorithmic}[1]
    \State Build $H$ and $\mathcal E$ from $\mathbf X$.
    \State Calculate $w$ and $\mu$ on $\mathcal E$.
    \State Adjust $\mu$ with training algorithm \ref{training} on $\mathbf X$ using rule \eqref{eqn:decision_rule}
    \For{each $\mathbf x$ in the test set}
        \State Calculate the projection $\pi(\mathbf x)$.
        \State Calculate the support $s(\mathbf x)$ using the projection.
        \State Predict using the refined rule \eqref{eqn:updated_cr}.
    \EndFor
  \end{algorithmic}
\end{algorithm}

While SVM aims at separating the data using a single hyperplane in the original input vector space, \HCBR~ tries to explain the decision for each case by a convex combination expressed in a lower dimensional space $\mathcal E$. 

\subsection{Step 1 - Model selection}
\label{sec:model_selection}

In this section, we define how to concretely select $\mu$ for a given hypergraph.  
Ultimately, $\mu$ is a measure over $\mathbb{F}_{\mathbf X}$ directly, and cannot be interpreted as a probability: $\mathbb{F}_{\mathbf X} \in 2^{\mathbb{F}_{\mathbf X}}$ but by definition of equation \eqref{eqn:model}, $\mu(\mathbb{F}_{\mathbf X}) = 1$. However, there is absolutely no reason to think that $J(\mathbb{F}_{\mathbf X}) = 1$. Our definition of $\mu$ tries to answer the following question: {\it knowing a certain amount of information materialized by the intersection family $\mathcal E$, where are located the features holding more discriminative information than the others?} Once again, with further information about the features, a prior might be introduced to select $\mu$.
 
For $\mathbf x$ and $\mathbf x'$ in $2^{\mathbb F}$, a basic quantitative measure on the importance of $\mathbf x'$ w.r.t. $\mathbf x$ can be expressed by $\frac{|\mathbf x \cap \mathbf x'|}{|\mathbf x|}$, i.e. how much $\mathbf x'$ overlaps with $\mathbf x$. This measure is a sort of {\it potential} for an analogy with $\mathbf  x$. Potential because it does not account for the individual importance of the features and simply holds the idea that the larger is a subset of features in a case, the higher is the chance it holds important features to decide the outcome.

Let us consider $\mathcal E$ and an example $\mathbf x \in \mathbf X$.
\begin{definition}[Intrinsic strength w.r.t. $\mathbf x$]
  The intrinsic strength of $\mathbf e \in \mathcal E$ w.r.t. $\mathbf x \in \mathbf X$ is defined by \begin{align}
    \begin{split}
    \forall l \in \{ -1, 1\}, ~ \bar S^{(l)}(\mathbf e, \mathbf x) & = \frac{|d^{(l)}(\mathbf e)| \frac{|\mathbf x \cap \mathbf e|}{|\mathbf x|}}{\underset{\mathbf e_j \in \mathcal E}{\sum}|d^{(l)}(\mathbf e_j)|  \frac{|\mathbf x \cap \mathbf e_j|}{|\mathbf x|}} \\
    & = \frac{|d^{(l)}(\mathbf e)|  |\mathbf x \cap \mathbf e|}{\underset{\mathbf e_j \in \mathcal E}{\sum}|d^{(l)}(\mathbf e_j)| |\mathbf x \cap \mathbf e_j|}
    \end{split}
    \end{align}

  Matricially, for any $\mathbf e_i$ and $\mathbf x_j$,
  \begin{equation}
   \bar S^{(l)}_{i,j} = \frac{d^{(l)}_i w_{ji}}{<\mathbf{d}^{(l)}, \mathbf{w}_{:i}>}
  \end{equation}
\end{definition}
\noindent
In particular, for a given $\mathbf x \in \mathbf X$, $\bar S^{(l)}(\mathbf e_i, \mathbf x) =0$ if $\mathbf e_i$ is not part of the projection of $\mathbf x$ on $H$.

The more $\mathbf e_i$ belongs to many cases with the same class $l$ and the higher $\bar S^{(l)}(\mathbf e_i, \mathbf x)$ is. Conversely, for a fixed number of cases, the more $\mathbf e_i$ describes $\mathbf x$, the higher $\bar S^{(l)}(\mathbf e_i, \mathbf x)$ is. As $\forall \mathbf e_i \in \mathcal E, ~ |d(\mathbf e_i)| > 0$, either we have $\bar S^{(1)}(\mathbf e_i, \mathbf x) \neq 0$ or $\bar S^{(-1)}(\mathbf e_i, \mathbf x) \neq 0$. We have $\bar S^{(l)}(\mathbf e_i, \mathbf x) = 0$ only when all the cases in which $\mathbf e_i$ results of are labeled with the opposite class $\bar l$. For $\bar S^{(l)}(\mathbf e_i, \mathbf x) = 1$, it needs both the unanimity of labels for the cases in which $\mathbf e_i$ belongs to and that $\mathbf e_i = \mathbf x$. The relation $\mathbf e_i = \mathbf x$ implies that $\mathbf x$ does not share any feature with any other example or that $\mathbf x$ is included into another example.
\begin{definition}[Intrinsic strength w.r.t. a hypergraph $H$]
  \label{intrinsic_strength}
  The intrinsic strength of $\mathbf e \in \mathcal{E}$ w.r.t. $H = (\mathbb{F}_{\mathbf X}, \mathbf X)$ is defined by
  \begin{align}
  \begin{split}
  \forall l \in \{ -1,  1 \}, ~ S^{(l)}(\mathbf e) & = \frac{|\mathbf e|}{ \underset{\mathbf e' \in \mathcal{E}}{\sum} |\mathbf e'|} \underset{\mathbf x \in d^{(l)}(\mathbf e)}{\sum} \bar S^{(l)}(\mathbf e, \mathbf x) \\
  & = \frac{|\mathbf e|}{|\mathbb{F}_{\mathbf X}|} \underset{\mathbf x \in d^{(l)}(\mathbf e)}{\sum} \bar S^{(l)}(\mathbf e, \mathbf x)
   \end{split}
  \end{align} 
  Matricially, for any $\mathbf e_i$,
  \begin{equation}
   S^{(l)}_{i} = \frac{|\mathbf e|}{|\mathbb{F}_{\mathbf X}|} ||\mathbf S^{(l)}_{i:}||_1
  \end{equation}

  The more $\mathbf e$ belongs to several cases, the more information it represents to support a class or another. As $\mathcal E$ represents the sets of features that appear all the time together, we favor the larger $\mathbf e \in \mathcal E$ as they hold more information to explain a decision.
  The normalized version is defined by:
    \begin{align}
  \forall l \in \{ -1, 1 \},~ \mu^{(l)}(\mathbf e) & = \frac{S^{(l)}(\mathbf e)}{ \underset{\mathbf e' \in \mathcal{E}}{\sum} S^{(l)}(\mathbf e')}
  \end{align}
\end{definition} 
Finally, the measure $\mu$ is simply defined by the difference between the strength of both classes:
\begin{align}
  \mu(\mathbf e) = \mu^{(1)}(\mathbf e) - \mu^{(-1)}(\mathbf e)
\end{align} which can be expressed matricially for a given $\mathbf{e}_i$ by
\begin{align}
  \mu_i = \frac{|\mathbf{e}_i|}{|\mathbb{F}_{\mathbf X}|}\big[ \frac{S^{(1)}_{i}}{||S^{(1)}||_1} - \frac{S^{(-1)}_{i}}{||S^{(-1)}||_1} \big]
\end{align}

~\\\noindent
{\bf Example (Numerical example):} Consider the hypergraph in Figure \ref{case_base} made of $\mathbf x_1$, $\mathbf x_2$ and $\mathbf x_3$ arbitrarily labeled with resp. 1, -1 and 1. Arbitrarily, we assume the cardinal of the elements of $\mathcal E$ to be $\mathbf{\#e} = (2,1,2,3,1,2,3)$ s.t. the cardinal of cases are $\#\mathbf x = (7, 8, 7)$ and $|\mathbb{F}_{\mathbf X}| = 14$. The values of $|d^{(l)}(\mathbf e)|$ can be summarized by the vectors $\mathbf d^{(-1)} = (0, 0, 1, 0, 1, 1, 1)$ and $\mathbf d^{(1)} = (1 , 2, 2, 1, 1, 1, 0)$. Let us calculate $S^{(-1)}(\mathbf e_3)$:
\begin{align*}
\bar S^{(1)}(\mathbf e_3, \mathbf x_1) & = \frac{2 \times 2}{2 \times 1 + 1 \times 2 + 2 \times 2 + 1 \times 2} = \frac 4 {10} \\
\bar S^{(1)}(\mathbf e_3, \mathbf x_2) & = \frac{2 \times 2}{2 \times 2 + 1 \times 1 + 1 \times 2 + 0 \times 3} = \frac 4 {7} \\
\bar S^{(1)}(\mathbf e_3, \mathbf x_3) & = \frac{2 \times 2}{2 \times 1 + 2 \times 2 + 3 \times 1 + 1 \times 1} = \frac 4 {10}
\end{align*}
 which we interpret as $\mathbf e_3$ being responsible for $\frac{4}{10}$ of the support toward class 1 in $\mathbf x_1$ and $\mathbf x_3$, while $\frac{4}{7}$ for $\mathbf x_2$.
This leads to
$$S^{(1)}(\mathbf e_3) = \frac 2 {14} \big[ \frac 4 {10} + \frac 4 {7} + \frac 4 {10}\big] \simeq 0.1959$$

Similarly, we compute the support for each $\mathbf e$ and both labels. We summarize this into the following vectors:
\begin{align*}
\mathbf{S}^{(1)} & \simeq (
0.0286,
0.0286,
0.1959 ,
0.0643,
0.0173,
0.0694, 
0.0000) \\ \mathbf{S}^{(-1)} &\simeq (
0.0000,
0.0000,
0.2024,
0.0000,
0.0327,
0.1071, 
0.0000)\end{align*} After normalization, we obtain the intrinsic strength:
$$\mu \simeq (0.0707, 0.0707, 0.0060, 0.1591, -0.0345, -0.0818, -0.1901)^T$$
Let us evaluate the model on the three examples:
\begin{align*}
  s(\mathbf x_1) & = \frac{2}{7}\mu(\mathbf e_1) + \frac{1}{7}\mu(\mathbf e_2) + \frac{2}{7}\mu(\mathbf e_3) + \frac{2}{7}\mu(\mathbf e_6) \simeq 0.0086\\
  s(\mathbf x_2) & = \frac{2}{8}\mu(\mathbf e_3) + \frac{1}{8}\mu(\mathbf e_5) + \frac{2}{8}\mu(\mathbf e_6) + \frac{3}{8}\mu(\mathbf e_7) \simeq -0.0946\\
  s(\mathbf x_3) & = \frac{1}{7}\mu(\mathbf e_2) + \frac{2}{7}\mu(\mathbf e_3) + \frac{3}{7}\mu(\mathbf e_4) + \frac{1}{7}\mu(\mathbf e_5) \simeq 0.0751\\
\end{align*}
As a result, $\mathbf x_1$ and $\mathbf x_3$ are labeled $1$ and $\mathbf x_2$ is labeled $-1$. All three cases are correctly labeled. The highest support is given for case $\mathbf x_2$ and $\mathbf x_3$ while the support for $\mathbf x_1$ is one order of magnitude lower than for $\mathbf x_3$. This is because the discretionary features of $\mathbf x_3$ are larger while the intersection with $\mathbf x_2$ is lower than for $\mathbf x_1$ ($\frac 3 8$ of $\mathbf x_3$ against $\frac 4 7$ of $\mathbf x_1$).

Consider a new case $\mathbf x$ as described in Figure \ref{new_case_schema}. Its support is given by $s(\mathbf x) = \underset{\mathbf e \in \pi(\mathbf x)}{\sum} w(\mathbf e, \mathbf x)\mu(\mathbf e)$ with $\pi(\mathbf x) = \{ \mathbf e_1, \mathbf e_2, \mathbf e_3, \mathbf e_5, \mathbf e_6, \mathbf e_7\}$. It is impossible for $\mathbf x$ to be classified as $1$ because the highest support would be for a maximal intersection with $\mathbf e_1$, $\mathbf e_2$, $\mathbf e_3$ and minimal for $\mathbf e_5$, $\mathbf e_6$ and $\mathbf e_7$ s.t. $s(\mathbf x) = \frac{1}{8}(2 \mu(\mathbf e_1) + \mu(\mathbf e_2) + 2 \mu(\mathbf e_3) + \mu(\mathbf e_5) + \mu(\mathbf e_6) + \mu(\mathbf e_7)) \simeq -0.0103 < 0$. It can be explained by the fact that the support for 1 is provided by a larger set of features (11 features versus 8). On top of that, the intersections between positive cases ($\mathbf e_2$ and $\mathbf e_3$) are too small (1 for $\mathbf e_2$ compared to e.g. 3 for $\mathbf e_7$) or include also negative cases ($\mathbf e_3$).

~\\\noindent {
{\bf Example (Car accident):} We are interested in knowing if a driver is responsible for an accident involving a pedestrian. From the past cases and the considered case, we collected seven facts: 1) the driver was not speeding, 2) the driver was speeding, 3) the pedestrian was outside the crosswalk, 4) the driver was drunk, 5) the accident happened at night, 6) the driver was a young driver, 7) the accident happened on the highway, 8) the crosswalk light was red.}

{To simplify, we assume two past cases: $\mathbf{x}_1 = \{2,4,5,7\}$, $\mathbf{x}_2 = \{1,3,5,6\}$ with $y_1 = 1$, $y_2 = -1$. We are interested in $y_3$ knowing $\mathbf{x}_3 = \{2,5,6,8\}$.
The partition is given by $\mathcal E = \{ \mathbf e_1 = \{2,4,7\}, \mathbf e_2 = \{ 5,6\}, \mathbf e_3 = \{ 1,3 \} \}$. In particular, $\pi(\mathbf x_3) = \{ \mathbf e_1, \mathbf e_2 \}$.}

{We assume $\mu^{(1)} = (0.75, 0.25, 0)$ and $\mu^{(-1)} = (0, 0.3, 0.7)$ s.t. after normalization $\mu = (0.5, -0.03, -0.46)$.
\begin{align*}
s(\mathbf x_3) & = w(\mathbf e_1, \mathbf x_3)\mu_1 + w(\mathbf e_2, \mathbf x_3)\mu_2 \\
               & = \frac 1 4 0.5 - \frac 1 2 0.03 = 0.11 > 0
\end{align*}
The model concludes that the driver is guilty. The fact the crosswalk light was red is not taken into account. The decision can be explained mostly by the features of $\mathbf e_1$. The elements of $\mathbf e_2$ are not discriminative enough to reverse the judgement. If the driver's lawyer would highlight the fact his client is not responsible because of the outcome of $\mathbf x_2$  and the fact it shares many similarities, the defense could argue that the main reason $y_2 = -1$ holds in $\mathbf e_3$. By building a chain of analogies and counter-examples, each decision can be explained w.r.t. past cases.}

\subsection{Step 2 - Training}
\label{sec:decision_training}

{At this stage, it is already possible to generate predictions for new cases.} However, the intrinsic strength vector calculated on the hypergraph might not be perfectly accurate on the training set because of the lack of information contained in the training set (or some limitations on the model space itself that we will discuss in Section \ref{sec:intrinsic_perf}). In this section, we give an algorithm to adjust the intrinsic strength in order to correct the initial estimation.

Once the model is built, it can be evaluated on the training set. Analogously to SVM, we define the margin as the distance to the correct class, i.e. $m(w, \mu, \mathbf x) = J(\mathbf x) s_{w,\mu}(\mathbf \mathbf x)$. To improve the pertinence of the strength of the elements of $\mathcal{E}$, we use the iterative algorithm described by Algorithm \ref{training} to minimize the total margin over the training set $\mathbf X$.

\algrenewcommand\algorithmicindent{1.0em}%
\begin{algorithm}
  \caption{Model training}\label{training}
  \begin{flushleft}
  \textbf{Input:} \\
    ~~- $\mathbf X$: training set \hfill\\
    ~~- $\mathbf y$: correct labels for $\mathbf X$\\
    ~~- $k$: number of training iterations\\
    ~~- $\mu^{(1)}, \mu^{(-1)}$: weights calculated with \eqref{intrinsic_strength}\\
    \textbf{Output:} \\
    ~~- Modified vectors $\mu^{(1)}, \mu^{(-1)}$
  \end{flushleft}
  \begin{algorithmic}[1]
      
      \For{$k$ \texttt{iterations}}
        \For{$\mathbf x_i \in \mathbf X$}
          \State $\bar y_i \gets \bar J(\mathbf x_i)$
          \If{$\bar y_i \neq y_i$}
            \For{$\mathbf e \in \pi(\mathbf x_i)$}
              \State $\mu^{(y_i)}(\mathbf e) \gets \mu^{(y_i)}(\mathbf e) + w(\mathbf e,\mathbf x_i) |\mu(\mathbf e)|$
              \State $\mu^{(\bar y_i)}(\mathbf e) \gets \mu^{(\bar y_i)}(\mathbf e) - w(\mathbf e,\mathbf x_i) |\mu(\mathbf e)|$
            \EndFor
          \EndIf
        \EndFor
      \EndFor
  \end{algorithmic}
\end{algorithm}

{The order in which points are considered is fixed in the current implementation}. When a decision is incorrect for $\mathbf x$, the algorithm modifies each element of the projection by lowering its strength for the wrong class and increasing it for the proper class. The margin is split between the element of the projection w.r.t. their respective weight in $\mathbf x$ i.e. $w(\mathbf e, \mathbf x)$. If a case $\mathbf x$ is wrongly classified, it is due to the cases intersecting with it. Indeed, if $\mathbf x$ was not intersecting with any other example, its projection would be itself, and its support toward the wrong class would be 0 and positive for the real class. In other words, $\mathbf x$ would be correctly classified. Thus, the idea is not to directly bring the support of $x$ to the correct class but to gradually adjust the weights s.t. the neighbors are modified enough for $\mathbf x$ to be correctly classified. 
In particular, it is sensitive to the order in which the cases are considered: a modification in the strength of any $\mathbf e \in \mathcal{E}$ impacts all cases in which it appears and potentially changes the predicted class for those cases.

The update rule being a contracting mapping because $|\mu(\mathbf e)| < 1$ at the initial step and $w(\mathbf e,\mathbf x_i) < 1$ unconditionally, Algorithm \ref{training} is guaranteed to converge. {{A more formal justification can be found in the Additional Material, Section 1}}. However, too many iterations may lead to overfitting $\mu$ to the training set. Empirical experiments suggest that the result after one to five training iterations is (near-)optimal (see Section \ref{sec:intrinsic_perf}).

\subsection{Step 3 - Decision rule refinement}
\label{sec:dec_rule}

{This step is not mandatory in a sense that the model can be built and already generates predictions. 
The hyperparameters introduced in this section can help either to increase prediction accuracy (see Section \ref{sec:result_literature_comp}) or control the risk associated to a prediction (see Section \ref{sec:hyperparam}).}

The measure $\mu$ is defined as the difference of support for both classes. Thus, by linearity we can rewrite
\begin{align}
 \begin{split}
s(\mathbf x) & = \underset{i = 1}{\overset{M}\sum} w(\mathbf {e}_i, \mathbf {x}) \mu^{(1)}(\mathbf {e}_i) - \underset{i = 1}{\overset{M}\sum} w(\mathbf {e}_i, \mathbf {x}) \mu^{(-1)}(\mathbf {e}_i) \\
           & = s^{(1)}(\mathbf x) - s^{(-1)}(\mathbf x)\hfill
 \end{split}
\end{align}
This form is convenient because we can control how much evidence we need to support a specific class using the following constraints and a family $\eta$ of four hyperparameters:
\begin{subequations}
\begin{align}
  \tag{$C_0$}
  s^{(-1)}(\mathbf x) > \max(\frac{\bar \eta_{-1}}{1 - \bar \eta_{-1}}s^{(-1)}(\mathbf x), \eta_{-1}) \geq 0\label{eqn:n0} \\
  \tag{$C_1$}
  s^{(1)}(\mathbf x) > \max(\frac{\bar \eta_1}{1 - \bar \eta_1}s^{(-1)}(\mathbf x), \eta_1) \geq 0\label{eqn:n1}
\end{align}
\end{subequations} with $\eta_{-1}, \eta_1 \in \mathbb{R}^+$ and $\bar \eta_{-1}, \bar \eta_1 \in [0,1]$.
The constraints on $\eta_{-1}$ and $\eta_1$ define a minimal amount of support respectively toward class -1 and 1 while $\bar \eta_{-1}$ and $\bar \eta_1$ requires the support toward a class to be significantly higher than the support for the other class.
As $\mu$ is normalized over $\mathcal E$, the value of $\eta_{-1}$ and $\eta_1$ must be set w.r.t. the hypergraph. On the contrary, $\bar \eta_1$ and $\bar \eta_0$ can be set independently of the hypergraph.

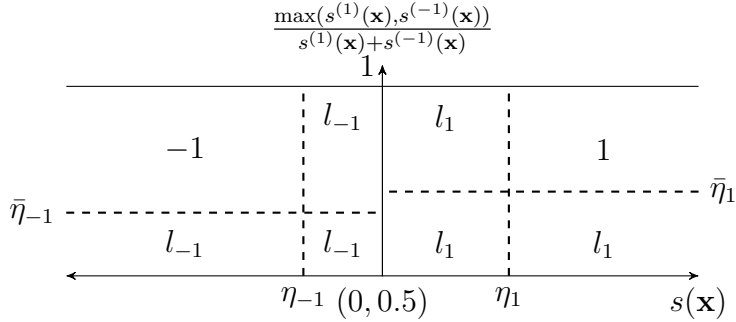
\begin{figure}
\begin{center}
 \begin{tikzpicture}[scale=1.4]
    \coordinate (y) at (0,2);
    \coordinate (x) at (3,0);
    \coordinate (mx) at (-3,0);
    \draw[axis] (y) -- (0,0) --  (x);
    \draw[axis] (y) -- (0,0) --  (mx);
    \coordinate (alphaas) at ($0.4*(y)$);
    \coordinate (alphabs) at ($0.3*(y)$);
    \coordinate (cfas) at ($1*(x)$);
    \coordinate (cfbs) at ($-1*(x)$);
    \coordinate (rl) at ($.4*(x)$);
    \coordinate (rr) at ($-.25*(x)$);

    \draw (0,1.8) node[label={[shift={(-0.2,-0.15)}]$1$}] {} -- (3,1.8);
    \draw (0,1.8) node[left] {} -- (-3,1.8);

    \draw[help lines] let \p1=(alphaas), \p2=(cfas) in 
    (\x2, \y1) node[right] {$\bar{\eta}_1$} -| (\p1);
    \draw[help lines] let \p1=(alphabs), \p2=(cfbs) in 
    (\x2, \y1) node[left] {$\bar{\eta}_{-1}$} -| (\p1);
    \draw[help lines] let \p1=(rl), \p2=(0,1.8) in
    (\p1) node[below] {$\eta_1$} -- (\x1, \y2);

    \draw[help lines] let \p1=(rr), \p2=(0,1.8) in
    (\p1) node[below] {$\eta_{-1}$} -- (\x1, \y2);
    \draw let \p1=($(alphaas)-(alphabs)$), \p2=(rl), \p3=(alphabs) in
    ($(.5*\x2, 2.5*\y3)$) node {$l_1$};
    \draw let \p1=($(alphaas)-(alphabs)$), \p2=($(cfas)-(rl)$),
    \p3=(alphabs), \p4=(rl) in
    ($(.5*\x2+\x4, 2.*\y3)$) node {$1$};
    \draw let \p1=(alphabs), \p2=(rl) in
    ($(.5*\x2, .5*\y1)$) node {$l_1$};
    \draw let \p1=(alphabs), \p2=($(cfas)-(rl)$), \p3=(rl) in
    ($(.5*\x2+\x3, .5*\y1)$) node {$l_1$};

    \draw let \p1=($(alphaas)-(alphabs)$), \p2=(rr), \p3=(alphabs) in
    ($(.5*\x2, 2.5*\y3)$) node {$l_{-1}$};
    \draw let \p1=($(alphaas)-(alphabs)$), \p2=($(cfbs)-(rr)$),
    \p3=(alphabs), \p4=(rr) in
    ($(.5*\x2+\x4, 2.*\y3)$) node {$-1$};
    \draw let \p1=(alphabs), \p2=(rr) in
    ($(.5*\x2, .5*\y1)$) node {$l_{-1}$};
    \draw let \p1=(alphabs), \p2=($(cfbs)-(rr)$), \p3=(rr) in
    ($(.5*\x2+\x3, .5*\y1)$) node {$l_{-1}$};

    \draw (0,0) node[below] {$(0, 0.5)$};
    \draw (3,0) node[below] {$s(\mathbf x)$};
    \draw (0,2) node[above] {$\frac{\max(s^{(1)}(\mathbf x), s^{(-1)}(\mathbf x))}{s^{(1)}(\mathbf x) + s^{(-1)}(\mathbf x)}$};
  \end{tikzpicture}
  \end{center}
  \caption{\label{decision_space} Representation of the updated decision rule \eqref{eqn:updated_cr} in the extended decision space.}
\end{figure}

Those constraints may be used to design a decision rule for new cases depending on the application or the dataset. The most generic decision rule is as follows:
\begin{align}
  \tag{R2} \label{eqn:updated_cr}
   \tilde J(\mathbf x) =  \left\{\begin{matrix}
  1 & ~ s(\mathbf  x) > 0 & \text{ and } C_1 \hfill\\ 
  -1 & ~  s(\mathbf  x) \leq 0 &\text{ and } C_0 \hfill\\
  l_1 & ~ s(\mathbf  x) > 0 & \text{ and not } C_1 \hfill\\ 
  l_{-1} & ~  s(\mathbf  x) \leq 0 & \text{ and not } C_0 \hfill
  \end{matrix}\right.
  \end{align} where $l_{-1}, l_0$ are two labels. A representation is given by Figure~\ref{decision_space}.
Those hyperparameters are intended to model the ``burden of proof''. For instance, in a trial, one is assumed innocent until proven guilty which implies the support for the class ``guilty'' must be {\it beyond a reasonable doubt} (where the term reasonable is defined by the jurisprudence of the applicable country). In case $\eta_{-1} = \eta_1 = \bar \eta_{-1} = \bar \eta_1$ (and $l_{-1} = 0$ and $l_1 = 1$), then the decision rule is equivalent to the original one defined by \eqref{eqn:decision_rule}.

\subsection{Time complexity}
\label{sec:complexity}

\noindent
{\bf Model Building:} Given $\mathbf X \in ({2^{\mathbb F}})^N$, constructing $\mathcal{E}_H$ can be done in $\mathcal{O}(\underset{\mathbf x \in \mathbf X}{\sum}|\mathbf x|)$ by using a Partition Refinement data structure~\cite{Paige:1987:TPR:37185.37186}. Given $\mathbf x \in \mathbf X$, calculating the family $\{\bar S(\mathbf e, \mathbf x)\}_{\mathbf e \in \mathcal{E}_H}$ can be done in $\mathcal{O}(|\mathbf x|)$ by asking for each feature of $\mathbf x$ the $\mathbf e$ it belongs to and maintaining the size of each $\mathbf e$ during the construction of $\mathcal{E}_H$. Thus, calculating $\{\bar S(\mathbf e, \mathbf x)\}_{\mathbf e \in \mathcal{E}_H}$ for all $\mathbf x \in \mathbf X$ can be done in $\mathcal{O}(\underset{\mathbf x \in \mathbf X}{\sum}|\mathbf x|)$. On $m$-uniform hypergraphs (when all cases are described with $m$ features), it becomes $\mathcal{O}(mN)$. 

Calculating $\{S(\mathbf e)\}_{\mathbf e \in \mathcal{E}_H}$ and $\mu$ can be done in $\mathcal{O}(|\mathcal{E}_H|)$ because it requires to iterate over $\mathcal{E}_H$. An obvious upper bound on $|\mathcal{E}_H|$ is $|\mathbb{F}_{\mathbf X}|$ i.e. the number of vertices in the hypergraph. The worst-case cardinal of $\mathcal{E}_H$ is when each $\mathbf x \in \mathbf X$ intersects with all the others and none of them is strictly a subset of any other. Thus, $|\mathcal{E}_H| \leq \min(2^N -1, |\mathbb{F}_{\mathbf X}|)$.\\

\noindent
{\bf Learning Phase:} For each wrongly classified $\mathbf x \in \mathbf X$, a training iteration requires at most $\mathcal{O}(|\mathbf x|)$ steps (maximal cardinal for $\pi(\mathbf x)$). The worst-case scenario is when the model wrongly classifies every $\mathbf x \in \mathbf X$. Thus, the learning phase worst-case complexity is $\mathcal{O}(k \underset{\mathbf x \in \mathbf X}{\sum}|\mathbf x|)$ and on $m$-uniform hypergraphs it becomes $\mathcal{O}(k m N)$.\\

\noindent
{\bf Model Query:} For a case $\mathbf x \in 2^{\mathbb F}$, the projection can be done in $\mathcal{O}(|\mathbf x|)$. Calculating the classification rule also requires at most $\mathcal{O}(|\mathbf x|)$ (maximal cardinal for $\pi(\mathbf x)$).

\section{Experiments on structured datasets}
\label{sec:experiments}

{In this section, we validate \HCBR~on well-known structured datasets. Two series of experiments are performed. The first one  compares \HCBR~to the best results from the literature and includes hyperparameter tuning. The second one focuses on the robustness, i.e. the capacity to deliver good performances on a wide range of datasets without spending time on feature engineering, data preprocessing or hyperparameter tuning.}

{Some specific elements such as the learning curves or computing times are discussed in Section \ref{sec:intrinsic_perf} dedicated to intrinsic performances and properties.}

\subsection{Data and method}

We used seven structured datasets for binary classification. All of them are available either from the UCI Machine Learning Repository\footnote{\href{https://archive.ics.uci.edu/ml/index.php}{https://archive.ics.uci.edu/ml/index.php}}
 or provided with LIBSVM\footnote{\url{https://www.csie.ntu.edu.tw/\~cjlin/libsvmtools/datasets/binary.html}}
 : \texttt{adult},
  \texttt{breasts}, \texttt{heart}, \texttt{mushrooms}, \texttt{phishing}, \texttt{skin} and \texttt{splice}. For each dataset, the original features \texttt{(name=value)} are converted into a unique identifier and the union of all such identifiers constitutes the information set $\mathbb{F}$.

The datasets are described by Table \ref{table:dataset}. The minimal, maximal and average size give information about the case sizes (notice \texttt{adult}, \texttt{heart} and \texttt{mushrooms} datasets are missing values). The column unique reports the cardinal of $\mathbb F$. Two datasets have at least one real-valued attribute as indicated by the column ``Real''. Three datasets (\texttt{adult}, \texttt{breasts} and \texttt{skin}) are highly imbalanced.

\begin{table*}[tb]
\begin{center}
  \caption{Datasets description. }
  \begin{small}
  \begin{tabular}{|l|r|r|r|r|r|r|l|r|}
    \hline
     & Cases & Total features & Unique & Min. size & Max. size & Avg. size & Real & Prev. \\
    \hline
    \texttt{adult} & 32561 & 418913 & 118 & 10 & 13 & 12.87 & No & 0.7586\\
    \texttt{breasts} & 699 & 5512 & 80 & 8 & 8 & 8 & No & 0.3338\\
    \texttt{heart} & 270 & 3165 & 344 & 12 & 13 & 12.99 & Yes & 0.5107\\
    \texttt{mushrooms} & 8124 & 162374 & 106 & 20 & 20 & 20 & No & 0.4804\\
    \texttt{phishing} & 11055 & 319787 & 808 & 29 & 29 & 29 & No & 0.5562\\
    \texttt{skin} & 245057 & 734403 & 768 & 3 & 3 & 3 & Yes & 0.2075\\
    \texttt{splice} & 3175 & 190263 & 237 & 60 & 60 & 60 & No & 0.5164\\
    \hline
  \end{tabular}
  \end{small}
  \label{table:dataset}
\end{center}
\end{table*}

For both experiments, we saved the confusion matrix obtained over all runs and after each prediction. From this confusion matrix, we calculated standard performance indicators: accuracy, recall, specificity, precision, negative prediction value, $F_1$-score and Matthews correlation coefficient. Denoting by TP the number of true positives, TN the true negatives, FP the false positives and FN the false negative, the accuracy, the $F_1$-score and Matthews correlation coefficient (MCC) are defined by:
\begin{align*}
\text{ACC} & = \frac{\text{TP} + \text{TN}}{\text{TP} + \text{TN} + \text{FP} + \text{FN}} \\
&\\
\text{F}_1 & = \frac{2 \text{TP}}{ 2\text{TP} + \text{FP} + \text{FN}} \\
    & \\
\text{MCC} & = \frac{\text{TP} \times \text{TN} - \text{FP} \times \text{FN}}{\sqrt{(\text{TP} + \text{FP})(\text{TP} + \text{FN})(\text{TN} + \text{FP})(\text{TN} + \text{FN})}}
\end{align*}
The accuracy, $\text{F}_1$-score and MCC respectively belongs to $[0,1]$, $[0,1]$ and $[-1,1]$. The closer to 1, the better it is. $\text{F}_1$-score and MCC take into account false positive and false negatives. Furthermore, MCC has been shown to be more informative than other metrics derived from the confusion matrix \cite{Chicco2017}, in particular with imbalanced datasets.

\subsubsection{Experiment 1 -- Literature comparison}

The objective is to compare \HCBR~performance to the best results from the literature. As most studies do not report $\text{F}_1$-score nor MCC, we will base our comparison on the accuracy. 

We used a 10-fold cross-validation with stratified sample to preserve the original dataset prevalence. The number of training steps $k$ was adjusted with a manual trial-and-error approach. 

{To measure the impact of hyperparameters on \HCBR, the runs have been performed twice (with the same seed). The first campaign was done with the family $\eta$ sets to $\eta_{-1} = \eta_1 = \bar \eta_{-1} = \bar \eta_1 = 0$. The second one was done with a nested 10-fold cross-validation in order to tune hyperparameters. The training set was itself divided into 10 folds. The first nine folds were used to build and train the model, the last one used as validation set to find the optimal values for the family $\eta$. On the test set, we used for $\eta$ the average values found over the 10 validation sets.} 

{It is difficult to know in advance the range of values for the components of $\bf \mu$ such that setting a grid is not convenient. Instead, we define a hyperparameter space that depends on the strength expressed in the decision defined by Figure \ref{decision_space}.}
Consider the upper half-plane of this space, i.e. the points with positive support. The hyperparameters can define a vertical split, horizontal split or another (open) half-plane with a certain margin from the axes. 
We performed an exhaustive search for the family $\eta$. We describe here the procedure for cases with a positive support i.e. to find $\eta_1$ and $\bar \eta_{1}$. The procedure is symmetric for negative support.
\begin{enumerate}
  \item For each point $(x,y)$ from the validation set, we considered three points: $p_1 = (x + \varepsilon,y+ \bar \varepsilon)$, $p_2 = (x  + \varepsilon, 0)$ and $p_3 = (0, y+ \bar \varepsilon)$ where $\varepsilon$ and $\bar \varepsilon$ are positive or negative depending on whether $(x,y)$ is correctly classified or not and set to half the distance to the immediate neighbor of $(x,y)$.
  \item For each point of those three points, we recalculated the accuracy when $(\eta_1, \bar \eta_{1})$ is sets to $p_i$
  \item We choose the values that return the highest accuracy. 
\end{enumerate}

\subsubsection{Experiment 2 -- Robustness comparison}

{The state-of-the-art results obtained per dataset required a lot of efforts in terms of feature engineering, as well as model selection or hyperparameter tuning. As discussed in Section \ref{sec:bc_unstructured}, it represents nowadays most data scientists work and CPU time.}
 If well-calibrated models with high performances on their domain of application is undeniably useful for practical usage, being able to solve a problem on a large variety of instances without effort is also of a vital importance, especially in the industry where the end-user might not be specialized in data science and machine learning. One main-challenge of large scale machine learning systems is thus to achieve the compromise between ready-to-deploy models and good performance metrics.

{There exist tools such as auto-sklearn \cite{NIPS2015_5872} to automatically tune hyperparameters and select the best algorithm in a given portfolio. However, this approach still requires an extensive CPU time which might be prohibitive in case end-users need to build classification models over numerous datasets. For instance, one may think to cloud offers such as IBM Cloud\footnote{\url{https://www.ibm.com/cloud/}} or Amazon AWS\footnote{\url{https://aws.amazon.com/}} that create for their customers thousands of models from scratch every day. Having models that give good results with limited CPU time dedicated to feature engineering and hyperparameter is thus an important competitive advantage.
On top of that, Automated Machine Learning is far from being widely adopted in the industry where hyperparameter tuning might not even be done for several reasons, as noted by \cite{couronne2018random} and confirmed by our experience.} 

Therefore, we would like to quantify how good \HCBR~can perform on different datasets without feature engineering or hyperparameter tuning. To measure this robustness, we performed a 10-fold cross-validation for each of the seven selected datasets using nine standard classification methods: AdaBoost, k-Nearest Neighbors, Linear SVM, Radius-Based Function (RBF) SVM, Decision Tree, Random Forest, Neural Network and Quadratic Discriminant Analysis (QDA). The implementation is provided by Scikit-Learn \cite{scikit-learn}. No hyperparameter tuning was performed and the default values of parameters were used.

\label{sec:classification_performances}

\subsection{Previous work on the datasets}

To compare the results of the proposed method, we explored for each dataset the best results from the literature.
The results are summarized in Table \ref{table:prev_results}. In \cite{doi:10.1504/IJBISE.2016.081590}, 5 rule-based classification techniques dedicated to medical databases are compared and achieve at best 95.85\% and 82.96\% accuracy resp. on \texttt{breast}, and \texttt{heart} datasets. Comparing bayesian approaches, \cite{Jiang:2012:LIW:2124637.2124641} demonstrated 97.35\% (\texttt{breast}) and 83.00\% (\texttt{heart}) accuracy. A 5 layers neural network with fuzzy inference rules achieved 87.78\% on \texttt{heart} \cite{sagir2017hybridised} while a k-NN algorithm reached 99.96\% on \texttt{mushrooms} \cite{Das:2001:FWB:645530.658297}. The best alternative among 6 rules-based classification methods achieved 95.84\% on \texttt{breast} and 100.00\% on \texttt{mushroom} \cite{HADI2017287}. Using 80\% of \texttt{phishing} as training set, an adaptative neural network achieved an average accuracy of 93.76\% (among 6 alternatives) with the best run at 94.90\% \cite{7727750}. Still on \texttt{phishing}, \cite{7881507} proposes to combine several classifiers and reaches 97.75\% accuracy for the best hybrid model (and demonstrates 97.58\% for Random Forest classifier). On \texttt{adult}, the comparison of several classifiers (naive bayes, decision tree, ...) demonstrated at most 86.25\% accuracy \cite{kou2012evaluation} while a Support Vector Machine approach reached 85.35\% \cite{Lee2001}. On \texttt{splice}, a method using Fuzzy Decision Trees \cite{5409447} reaches 94.10\% accuracy and a neural network combined to boosting \cite{catak2017} 97.54\%. On \texttt{breast}, Support Vector Machine approaches reached resp. 96.87\%, 98.53\%, 99.51\% accuracy \cite{CHEN20119014, POLAT2007694, akay2009support}, 99.26\% and 97.36\% for neural network based techniques \cite{MARCANOCEDENO20119573, ubeyli2007implementing}, 98.1\% for a bayesian network method \cite{fallahi2011expert}, or 94.74\% using Decision Trees \cite{quinlan1996improved}. On \texttt{skin}, \cite{catak2017} reports 98.94\% accuracy against 99.68\% for Decision Tree based method~\cite{6627823}. The best result, as far as we know, is 99.92\%, obtained by a Generalized Linear Model~\cite{basterrech2015generalized}.

\begin{table}[!htbp]
\begin{center}
  \caption{Previous literature results measured as the highest accuracy obtained by the authors.}
  \begin{small}
\begin{tabular}{|c|c|l|c|}
\hline
 Dataset & Ref. & Type & Accuracy   \\ \hline
\multirow{2}{*}{\texttt{adult}} & \cite{kou2012evaluation} & Many classifiers &  86.25\% \\
& \cite{Lee2001} & SVM & 85.35\% \\
& & {\bf \bfHCBR (tuned)} & {\bf 82.90\%} \\
& & \bfHCBR & {\bf 82.06\%} \\ \hline
 \multirow{10}{*}{\texttt{breast}} & \cite{akay2009support}  & SVM & 99.51\% \\ 
 & \cite{MARCANOCEDENO20119573} & Neural Net & 99.26\% \\ 
 & \cite{POLAT2007694} & SVM & 98.53\% \\ 
 & \cite{fallahi2011expert} & Bayes & 98.1\% \\ 
 & & {\bf \bfHCBR (tuned)} & {\bf 97.83\%} \\
 & \cite{ubeyli2007implementing} & Neural Net & 97.36\% \\ 
 & \cite{Jiang:2012:LIW:2124637.2124641} & Bayes & 97.35\% \\ 
 & & {\bf \bfHCBR} & {\bf 96.96\%} \\
 & \cite{CHEN20119014} & SVM & 96.87\% \\
 & \cite{doi:10.1504/IJBISE.2016.081590} & Rule-based & 95.85\% \\
 & \cite{HADI2017287} & Rule-based & 95.84\% \\
 & \cite{quinlan1996improved} & Decision Tree & 94.74\% \\ \hline
 \multirow{3}{*}{\texttt{heart}} & & {\bf \bfHCBR (tuned)} & {\bf 90.77\%} \\
 & \cite{sagir2017hybridised} & Neural Network + Rule-based & 87.78\%\\
 & & {\bf \bfHCBR} & {\bf 85.77\%} \\
 & \cite{Jiang:2012:LIW:2124637.2124641} & Bayes & 83.00\% \\ 
 & \cite{doi:10.1504/IJBISE.2016.081590} & Rule-based & 82.96\% \\ \hline
 \multirow{2}{*}{\texttt{mushrooms}} &  \cite{HADI2017287} & Rule-Based & 100.00\% \\ 
  & & {\bf \bfHCBR} & {\bf 100.00\%} \\
  & \cite{Das:2001:FWB:645530.658297} & k-NN & 99.96\% \\ \hline
 \multirow{3}{*}{\texttt{phishing}} & \cite{7881507} & Ensemble & 97.75\% \\ 
 &  \cite{7881507} & Random-Forest & 97.58\% \\ 
 & & {\bf \bfHCBR (tuned)} & {\bf 96.82\%} \\
 & & {\bf \bfHCBR}& {\bf  96.05\%} \\
 & \cite{7727750} & Neural Net &  94.90\% \\\hline
 \multirow{3}{*}{\texttt{skin}} & \cite{basterrech2015generalized} & Generalized Linear Model & 99.92\% \\ 
 & \cite{6627823} & Decision Tree & 99.68\% \\ 
 & \cite{catak2017} & Neural Network + Boosting & 98.94\% \\
 & & {\bf \bfHCBR (tuned)} & {\bf 98.68\%}\\
 & & {\bf \bfHCBR} & {\bf 98.65\%} \\ \hline
 \multirow{2}{*}{\texttt{splice}} & \cite{catak2017} & Neural Network + Boosting & 97.54\% \\
 & & {\bf \bfHCBR (tuned)} & {\bf 95.09\%} \\
 & & {\bf \bfHCBR} & {\bf 94.43\%} \\
 & \cite{5409447} & (fuzzy) Decision Tree & 94.10\% \\ \hline
\end{tabular}
\end{small}
  \label{table:prev_results}
\end{center}
\end{table}

\subsection{Results}

The integrality of the data used for the experiments, as well as the scripts to transform them and analyze the results are available within the \HCBR~Github repository\footnote{\href{https://github.com/aquemy/HCBR}{https://github.com/aquemy/HCBR}} s.t. the whole experimental campaign starting from the raw data can be easily be reproduced. 

\subsubsection{Literature comparison} \label{sec:result_literature_comp}

The average confusion matrix obtained for each dataset is showed in the Additional Material, Section 2. The performance indicators are reported in Table \ref{table:perf_indicators}. The proposed algorithm performs very well on a wide range of datasets as reported by the Additional Material, Section 2 and Table \ref{table:perf_indicators}.\\ 

\noindent
{\bf Without hyperparameter tuning:} The accuracy is contained in a range from 0.8206 (\texttt{adult}) to 1 (\texttt{mushrooms}) while the $\text{F}_1$-score is bounded by 0.8653 (\texttt{heart}) and 1 (\texttt{mushrooms}). On \texttt{adult}, the accuracy is only 6\% higher than the prevalence, i.e. a baseline model consisting in returning 1 for any point would be only 6\% worse. This relatively poor performance in learning the underlying decision mapping is better reflected by the Matthews correlation coefficient of $0.51$.

\begin{table*}[htbp]
\begin{center}
  \caption{Average performances obtained with a 10-fold cross-validation.}
 \resizebox{16cm}{!}{
  \begin{small}
  \begin{tabular}{|l|c|c|c|c|c|c|c|c|c|c|c|}
    \hline
     & Accuracy (std dev.) & Recall & Specificity & Precision & Neg. Pred. Value & $\text{F}_1$ score & MCC\\
    \hline
    \multirow{2}{*}{\texttt{adult}} & 0.8206 (0.0094) & 0.8832 & 0.6233 & 0.8808 & 0.6290 & 0.8820 & 0.5081\\
    & 0.8290 (0.0063) & 0.9008 & 0.6029 & 0.8773 & 0.6029 & 0.8889 & 0.5194 \\ \hline
    \multirow{2}{*}{\texttt{breasts}} & 0.9696 (0.0345) & 0.9691 & 0.9676 & 0.9479 & 0.9844 & 0.9575 & 0.9344\\
    & 0.9783 (0.0204) & 0.9553 & 0.9910 & 0.9833 & 0.9910 & 0.9691 & 0.9526 \\ \hline
    \multirow{2}{*}{\texttt{heart}} & 0.8577 (0.0943) & 0.8695 & 0.8437 & 0.8699 & 0.8531 & 0.8653 & 0.7178\\
    & 0.9077 (0.0659) & 0.9310 & 0.8783 & 0.9060 & 0.8783 & 0.9184 & 0.8126 \\ \hline
    \texttt{mushrooms} & 1.0000 (0.0000) & 1.0000 & 1.0000 & 1.0000 & 1.0000 & 1.0000 & 1.0000\\ \hline
    \multirow{2}{*}{\texttt{phishing}} & 0.9605 (0.0081) & 0.9680 & 0.9514  & 0.9615 & 0.9590 & 0.9647 & 0.9199\\
    & 0.9682 (0.0067) & 0.9689 & 0.9672 & 0.9741 & 0.9672 & 0.9715 & 0.9355 \\ \hline
    \multirow{2}{*}{\texttt{skin}} & 0.9865 (0.0069) & 0.9608 & 0.9932  &0.9736 & 0.9898 & 0.9672 & 0.9587 \\
    & 0.9868 (0.0062) & 0.9740 & 0.9900 & 0.9618 & 0.9900 & 0.9679 & 0.9596 \\ \hline
    \multirow{2}{*}{\texttt{splice}} & 0.9443 (0.0124) & 0.9478 & 0.9398 & 0.9450 & 0.9441 & 0.9463 & 0.8884\\
    & 0.9509 (0.0108) & 0.9478 & 0.9544 & 0.9577 & 0.9544 & 0.9527 & 0.9018 \\
    \hline
  \end{tabular}
  \end{small}
  }
  \label{table:perf_indicators}
\end{center}
\end{table*}

The false positives and false negatives are equilibrated for each dataset, despite a huge variation in the prevalence (between 20\% and 64\%, cf. Table \ref{table:dataset}) which is a desirable property as it is known to be a problem for many machine learning algorithms \cite{He:2009:LID:1591901.1592322}.

\begin{figure}[!h]
\centering
\includegraphics[scale=0.3]{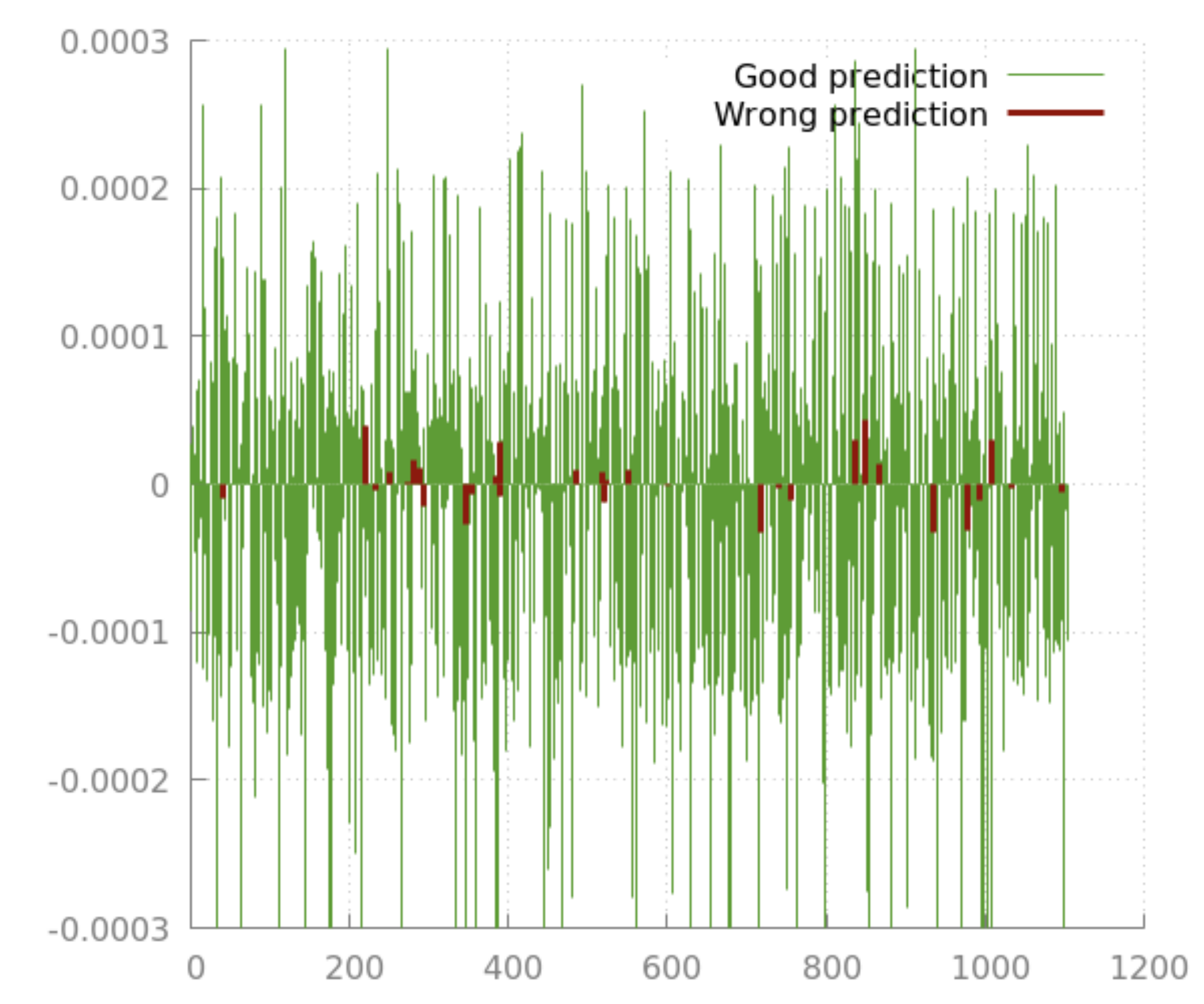}
\includegraphics[scale=0.3]{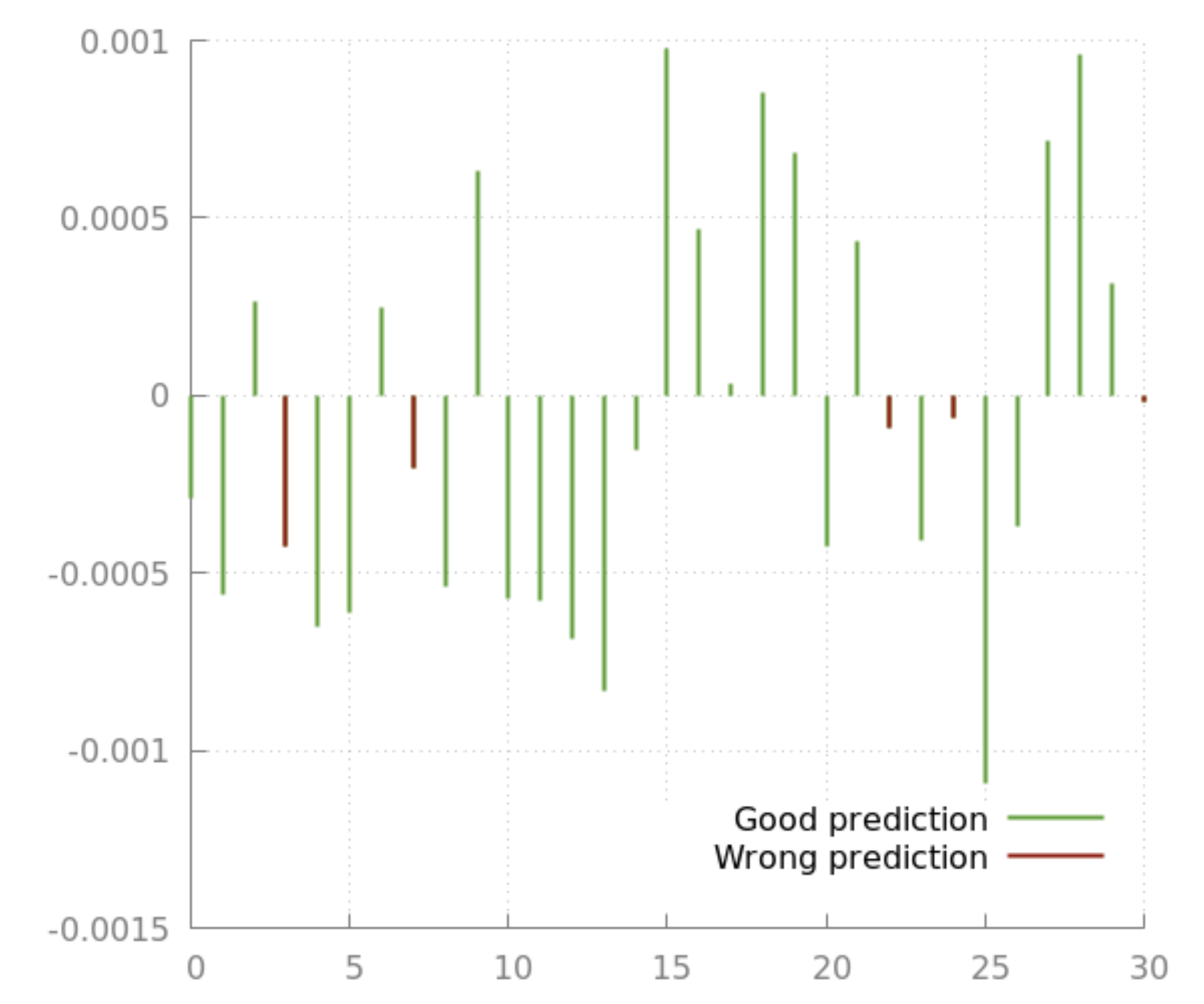}
\caption{Difference between the weight assigned to both classes for each decision on \texttt{phishing} and \texttt{splice} (average). Similar results are observed for all datasets.}
\label{fig:phishing_predictive_measure}
\end{figure}

\begin{figure}[!h]
\centering
\includegraphics[scale=0.4]{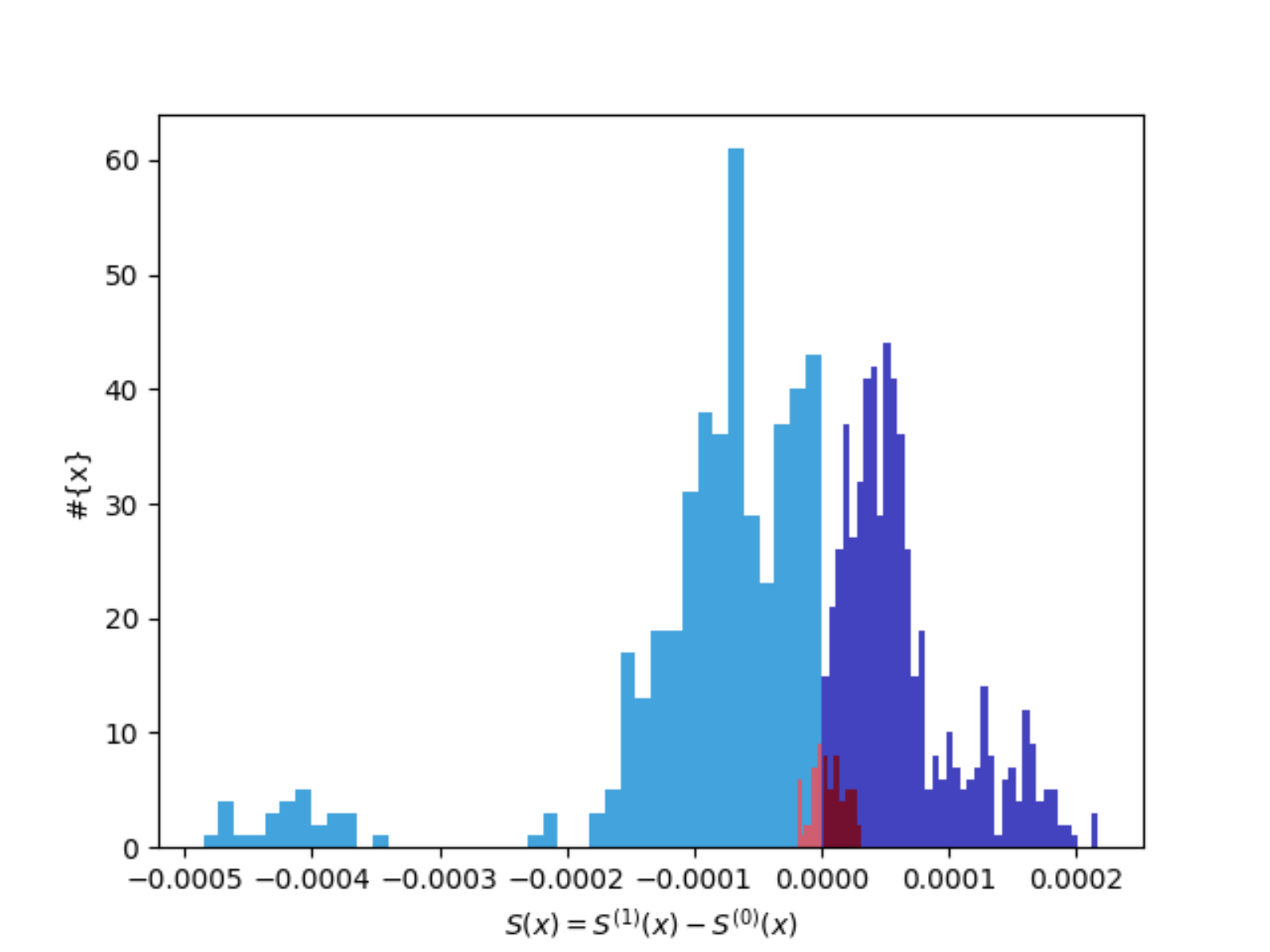}
\includegraphics[scale=0.4]{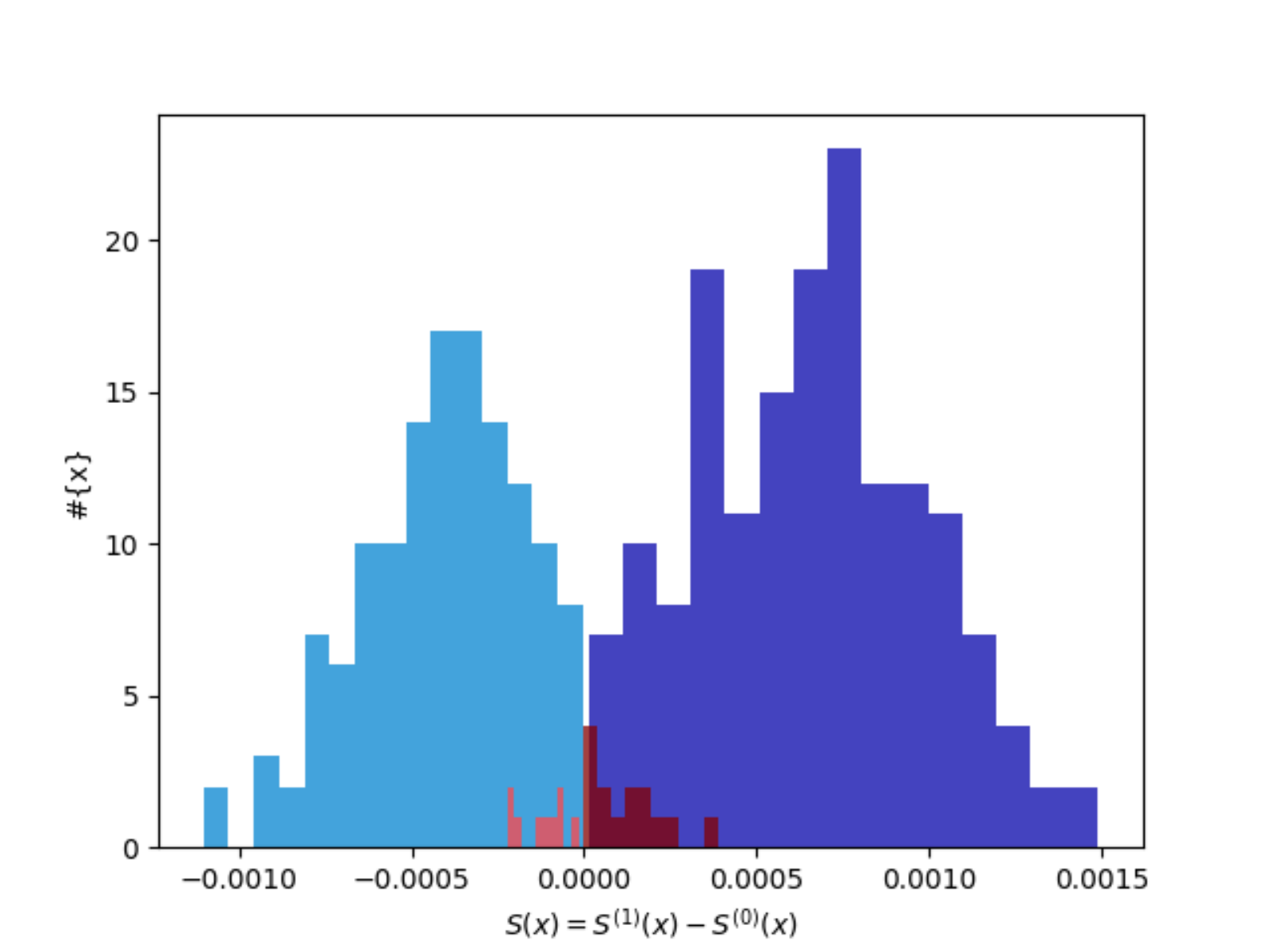}
\caption{Histogram of decisions depending on the strength for \texttt{phishing} and \texttt{splice}. In blue the correctly classified elements, in red the wrongly classified ones. The false positives and false negatives are concentrated around 0. Similar results are observed for all datasets.}
\label{fig:phishing_histogram}
\end{figure}

The support is a metric of confidence for the prediction as illustrated in Figure \ref{fig:phishing_predictive_measure}. In general, the wrongly classified cases have a smaller difference between the evidence for each class.
This can also be observed in Figure \ref{fig:phishing_histogram}.

\HCBR~performs the best on \texttt{mushrooms} and \texttt{heart} datasets. For the rest, the accuracy is slightly lower than the best results from the literature (\texttt{adult} 82.90\% against 86.25\%, \texttt{breast} 97.83\% against 99.51\%, \texttt{phishing} 96.82\% against 97.75\%, \texttt{splice} 95.09\% against 97.54\%, \texttt{skin} 98.68\% against 99.92\%). We explain this by at least two factors. First, the best methods on a given dataset are often dedicated to this dataset with {\it ad-hoc} or engineered parts which is not the case of \HCBR. Secondly, the learning curves study in Section \ref{sec:lc} reveals that the model space is not complex enough.
\HCBR~performed better than Bayes classifier in two thirds of cases. Bayes classifier performs better on \texttt{breast} by approximately 1\% which represents less than one case wrongly classified. Similar results are observed with Decision Trees. However, the 1\% difference on \texttt{skin} represents an average of 7 cases misclassified in comparison in favor of Bayes. It performs better than Rule-based approaches (or gives similar results on \texttt{mushrooms} with an accuracy of 1) in the four considered references on three different datasets. 
Except for \texttt{heart} and \texttt{phishing}, Neural Network returns better results ($0.46$ more cases with correct classification in average for \texttt{breast}, $71$ for skin and almost $10$ for \texttt{splice}). Last, SVM gives better results in all three cases, but appear only as best results in two datasets.\\

\begin{figure}[!h]
\centering
\includegraphics[scale=0.45]{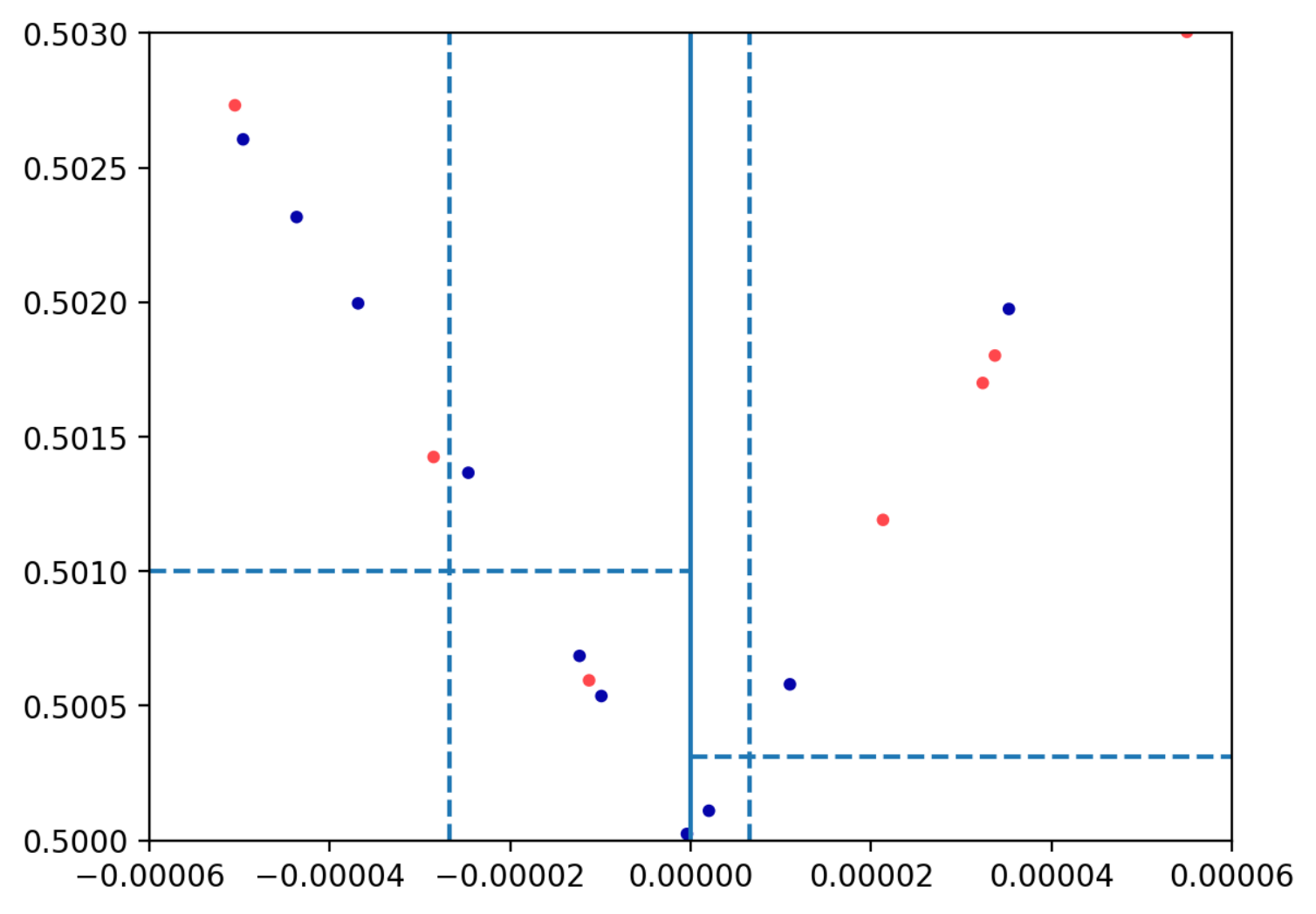}
\caption{Illustration of the hyperparameter optimization on \texttt{splice} (one specific run, zoomed in the origin). The dashed lines are defined by averaging the best parameters obtained on the test set (10 folds). For negative support, it turned 4 FN into FP and 1 TN into FP compared to the version without hyperparameter tuning. For positive support, it turned one FP into TN. The hyperparameter values are not optimal as a better result could have been achieved.}
\label{fig:splice_hyperopt}
\end{figure}

\noindent
{\bf With hyperparameter tuning:}
{An illustration of parameters tuning on a real instance is depicted by Figure \ref{fig:splice_hyperopt}. {With hyperparameter tuning, the accuracy lower bound is obtained by the same dataset (0.8290 with \texttt{adult}). The lower bound on $\text{F}_1$-score is now obtained by \texttt{adult} with 0.8889.}
On \texttt{heart}, the accuracy gains 5 percent point (pp) which represents a 35\% error reduction. This allowed \HCBR~ to rank first. On \texttt{breast}, the gain represents 0.87pp which is about 29\% error improvement and a higher accuracy than \cite{Jiang:2012:LIW:2124637.2124641} and \cite{ubeyli2007implementing}. For the other datasets, the accuracy variation does not result in a better rank and the error improvement lies between 2\% for \texttt{skin} and 19\% for \texttt{phishing}.}

{We explain the limited effect of hyperparameters by the fact that for small support values, there can be several cases with the exact same support but different outcomes. Two different cases have the same support if the elements of their projections are weighted the same despite being composed of different features. In order to be more discriminative, one can increase the model space complexity s.t. the support is obtained by more variables which decreases the probability of collisions. See Section \ref{sec:msl} for a study of the model space limitations and possible remedies.}\\

\noindent
{\bf Discussion:}
Even if \HCBR~does not rank first in most cases, we see at least three reasons to use it in practice. First, it provides consistently good results on all datasets without a need to tune any hyperparameter, without correcting the imbalanced datasets or feature transformation that would require expert knowledge. In other words, it requires absolutely no domain knowledge or data science expertise to deploy and use. Second, \HCBR~works directly in unstructured spaces allowing combining data from multiple sources without tedious transformations. This has a considerable advantage in practice where the information per case is rarely structured by default. Last but not least, \HCBR~provides local explanation: for each case and each group of features $\mathbf e_i$ in this case, \HCBR~provides not only the support for $\mathbf e_i$ but also all the cases that participated in increasing the support. In the medical domain, the practitioner may check, for instance, the top two groups of features, and for each, the five most influencing past cases to find a reasonable justification to the prediction.

\subsubsection{Robustness comparison}
\label{sec:robustness}

\begin{figure}[!h]\centering
\includegraphics[scale=0.3]{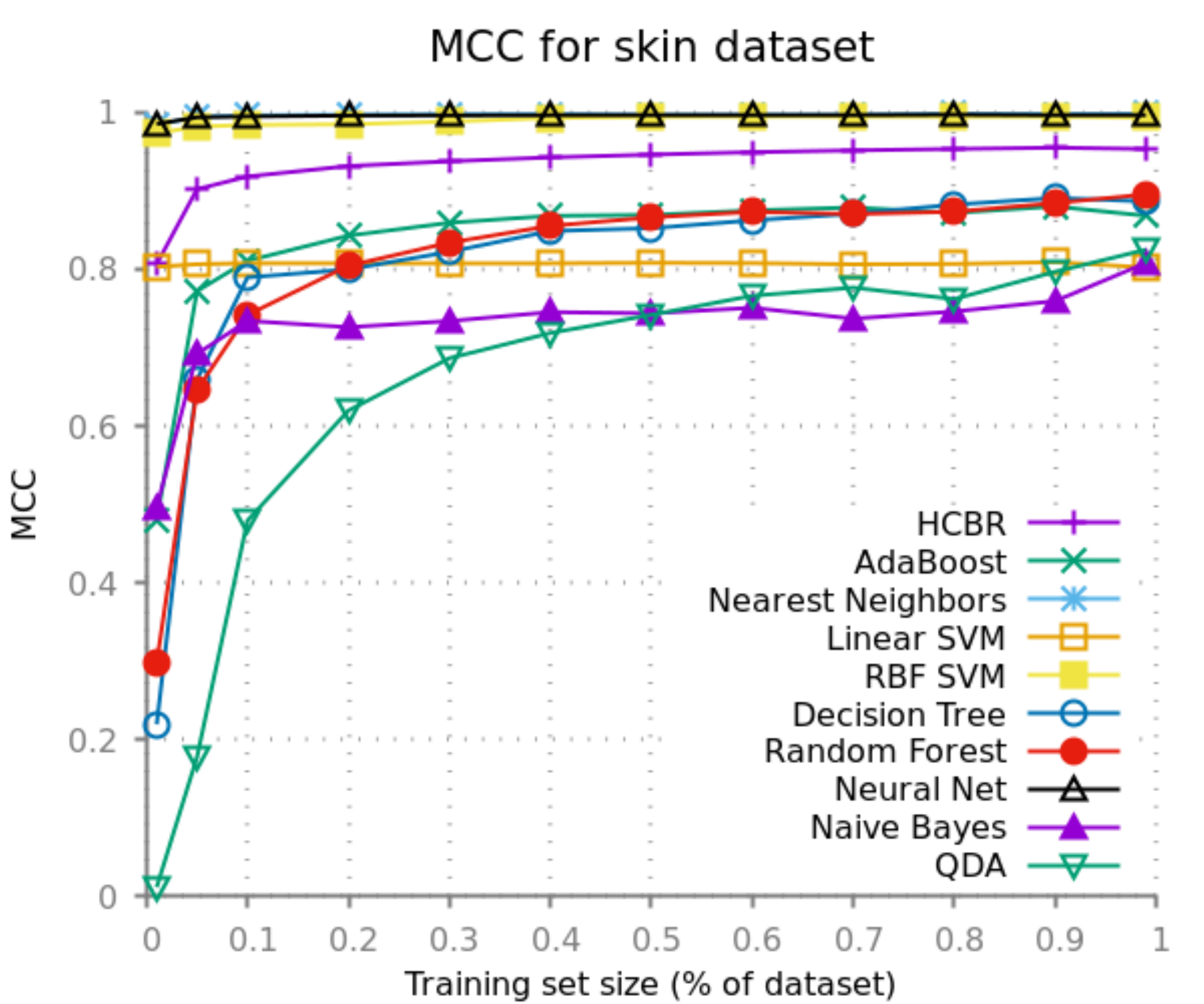}
\caption{Matthew Correlation Coefficient depending on the training set size for \texttt{skin}. See the Additional Material, Section 5 for other datasets.}
\label{fig:mcc_by_examples}
\end{figure}

The average MCC achieved per method over the datasets is reported in Table \ref{table:scikit_results_mcc}. The details per dataset can be found in the Additional Material, Section 3. {The counterpart for the accuracy is provided by Table \ref{table:scikit_all_acc} and the Additional Material, Section 4.} Additionally, we displayed the evolution of MCC with the training set size in Figure \ref{fig:mcc_by_examples}.

\begin{table}[h!]
\begin{center}
  \caption{Average MCC and rank obtained with several methods (Scikit-Learn implementation). The column $\Delta_{\HCBR}$ represents the relative difference in MCC w.r.t. \HCBR.}
  \begin{small}
\begin{tabular}{|l|c|c|r|}
\hline
  Method & MCC & Rank & $\Delta_{\HCBR}$ \\ \hline
 
 Neural Network & 0.8914 & 1 & 5.68\%\\
 \bfHCBR & 0.8435 & 2 & - \\
 RBF SVM & 0.8267 & 3 & 2.00\%\\
 Decision Tree & 0.8066& 4 & 4.37\% \\
 AdaBoost & 0.8063 & 5 & 4.41\% \\
 k-NN & 0.7859 & 6 & 6.82\% \\
 Linear SVM &  0.7858 & 7 & 6.84\% \\
 QDA & 0.7358 & 8 & 12.76\% \\
 Random Forest & 0.7237 & 9 & 14.20\% \\
 Naive Bayes & 0.6953 & 10 & 17.56\% \\
 \hline
\end{tabular}
\end{small}
  \label{table:scikit_results_mcc}
\end{center}
\end{table}

\begin{table}[h!]
\begin{center}
  \caption{Average accuracy and rank obtained with several methods (Scikit-Learn implementation). The column $\Delta_{\HCBR}$ represents the relative difference in accuracy w.r.t. \HCBR.}
  \begin{small}
\begin{tabular}{|l|c|c|r|}
\hline
 Method & Accuracy & Rank & $\Delta_{\HCBR}$ \\ \hline
 \bfHCBR & 0.9360 & 1 & - \\
 Neural Network & 0.9354 & 2 & 0.06\%\\
 RBF SVM & 0.9247 & 3 & 1.16\%\\
 AdaBoost & 0.9207 & 4 & 1.60\% \\
 Linear SVM & 0.9115 & 5 & 2.62\% \\
 Decision Tree & 0.9056 & 6 & 3.25\% \\
 k-NN & 0.9011 & 7 & 3.81\% \\
 Random Forest & 0.8903 & 8 & 4.88\% \\
 QDA & 0.8802& 9 & 6.27\% \\
 Naive Bayes & 0.8235 & 10 & 13.59\% \\
 \hline
\end{tabular}
\end{small}
  \label{table:scikit_all_acc}
\end{center}
\end{table}

The highest MCC value is obtained by Neural Network with 0.8914 followed by \HCBR~ with 0.8435. Neural Network improves \HCBR~result by 5.68\% while \HCBR~improves the result of all other methods from 2\% to 17.56\%. The lowest MCC score is obtained by Naive Bayes with 0.6953, mostly because it performs poorly on some datasets (0.2493 on \texttt{adult}, 0.5292 on \texttt{phishing} or 0.7600 on \texttt{skin}). In general, Neural Network and \HCBR~are the only two methods whose ranks remain consistent across all datasets, close to the first and second rank. On the contrary, methods like $k$-Nearest Neighbors or Decision Tree performed very well on one dataset (\texttt{skin}, resp. \texttt{splice}). Surprisingly, Random Forest not only perform poorer than Decision Tree but also performs worse than most methods.
Naive Bayes and QDA perform very poorly in general which is less surprising knowing the assumptions behind those methods that are likely to be unrealistic on real datasets.

{Regarding accuracy, \HCBR~is the best performing algorithm followed by Neural Network. In general, the ranking is consistent with the one obtained for MCC. }

In other words, \HCBR~and Neural Network have shown to be more versatile than the other approaches on this selected set of classification instances and without parameter tuning. We have no doubt that the results obtained with other methods can be highly improved by a proper hyperparameter selection, in particular for Random Forest or even k-Nearest Neighbors. However, this time-consuming operation can be avoided with \HCBR~that provides the best compromise between ready-to-use and good performances.

\section{Experiments on unstructured datasets for text classification}
\label{sec:unstructured_datasets}

{In this section we study the performance of \HCBR~on unstructured datasets and show it performs better in most cases compared to the reference study that uses SVM.} 

\subsection{Data and method}
{We used the European Court of Human Rights dataset provided by the authors of \cite{10.7717/peerj-cs.93}. This dataset for binary classification is broken down into three articles, namely Article 3, Article 6 and Article 8. For each article, the authors collected some judgments (250, 80 and 254 respectively for Article 3, 6 and 8) s.t. the prevalence is 50\%. For each document, they isolated different sections (Procedure, Circumstances, Relevant Law, Law) and excluded the verdict section. The authors do not provide the raw text but the Bag-of-Words representation obtained by keeping the 2000 most common N-grams for $N \in \{1,2,3,4\}$. The authors used a linear SVM after applying a TF-IDF schema on the dataset to predict whether an article has been violated.}

On top of those Bag-of-Words representations, the authors provide the representation of a judgement in a topic space. This topic space is obtained by calculating the matrix of cosine distance between the documents and keeping the 30 top-components after applying a spectral clustering. {Finally, the authors include two additional parts defined as the mean vector of other sections {\bf after} applying the TF-IDF transformation. Facts is the mean vector of Circumstances and Relevant Law, and Full as mean of all the other parts. Thus, they are continuous contrarily to the other sections.} In this work, we stacked the representations to obtain Facts and Full as \HCBR~does not handle properly continuous variables. 

It has to be noted that for some parts, there are empty cases. For the Law part with Article 3, there are 162 empty cases (64.8\% of the total casebase) with a prevalence of 17\%, for the Article 6, 52 cases are empty with 17\% prevalence and for the Article 8, 146 cases are empty for a 21\% prevalence. This is not compatible with the hypothesis s.t. if two cases are described by the same features, they must have the same outcome used both by \HCBR~and SVM. Therefore, the results are biased for both methods and could be improved by a better preprocessing step\footnote{Ironically, this supports our claim that shortening the data preprocessing steps is necessary in order to avoid such hard-to-notice problems.}.

{For each of the seven sections (Full, Procedure, Circumstances, Relevant law, Facts, Law, Topics), we performed a 10-fold cross-validation and reported the accuracy.}

\subsection{Results}

{Table \ref{table:echr_results} summarizes the results. The accuracy is improved in 14 cases out of 21, deteriorated in 5 and remained unchanged in 2 cases. The improvement ranges from 1 to 20pp while the deterioration ranges from 1pp to 8pp.}

 {All sections have seen their average accuracy unchanged or increased except for Topics. This is not surprising knowing Topics is made out of 30 continuous variables which are not correctly handled by \HCBR. Surprisingly, the conclusions we can draw from this experiment are quite opposed to those of the reference study \cite{10.7717/peerj-cs.93}. The full text was outperformed by using only the section Circumstances while here, taking the full text returns a much higher accuracy. The section with the best predictive power was Circumstances. It is now Relevant Law. The section Law was the worst predictor, it now one of the best, notably thanks to a gain of 20pp on Article 3.}

{In the original study, the best performances are obtained on Sections Topics and Topics and Circumstances with an accuracy of 0.78, 0.84 and 0.78 respectively for Article 3, 6 and 8. Thus, \HCBR~performed 2\% lower on Article 3 and 6 and 1\% better on Article 8. Knowing continuous variables are not properly handled yet, it is encouraging. Also, it turns out the best section for \HCBR~is the full text, which implies that in practice there is no need for feature engineering or time-consuming operations such as splitting the text into subsections. We are confident that \HCBR~could perform better with a larger number of tokens in the bag-of-word representation and let this for future work.}

\begin{table}[h!]
\begin{center}
  \caption{Accuracy obtained by \HCBR~on the European Court of Human Rights dataset, depending on the article and the judgement section. The columns $\Delta$acc represent the difference with the reference study \cite{10.7717/peerj-cs.93}. The color indicates if the result has been improved (green), deteriorated (red) or unchanged (white).}
\small{
\begin{tabular}{|l|c|r|c|r|c|r|c|r|} 
   \hline
     & Article 3 & $\Delta$acc & Article 6& $\Delta$acc &  Article 8 & $\Delta$acc & Average& $\Delta$acc\\
    \hline
    Full &\cellcolor{green!25}.76 & +.06 & \cellcolor{green!25}.83 & +.01 & \cellcolor{green!25}.77 & +.05 & \cellcolor{green!25}.79 & +.04\\
    Procedure & .67 & - & .81 & - & \cellcolor{green!25}.72 & +.01 & .73 & - \\
    Circumstances &\cellcolor{red!25}.67 & -.01 & \cellcolor{red!25}.81 & -.01 & \cellcolor{red!25}.72 & -.05 &  .73& - \\
    Relevant law &\cellcolor{green!25}.71 & +.02 & \cellcolor{green!25}.86 & +.08 & \cellcolor{green!25}.76 & +.04 & \cellcolor{green!25}.78& +.05\\
    Facts &\cellcolor{green!25}.73 & +.03 & \cellcolor{red!25}.76 & -.04 & \cellcolor{green!25}.72 & +.04  & \cellcolor{green!25}.74& +.01 \\
    Law &\cellcolor{green!25}.76 & +.20  & \cellcolor{green!25}.80 & +.12  & \cellcolor{green!25}.73 & +.11  & \cellcolor{green!25}.76& +.14 \\
    Topics  &\cellcolor{red!25}.70 & -.08  & \cellcolor{green!25}.83 & +.02 & \cellcolor{red!25}.74 & -.02  & \cellcolor{red!25}.76& -.02 \\
    \hline
\end{tabular}
}
  \label{table:echr_results}
\end{center}
\end{table}

\section{Intrinsic performances and properties}
\label{sec:intrinsic_perf}

In this section, we validate the time complexity (Section \ref{sec:comp_time}) and discuss the model space limitations of \HCBR~(Section \ref{sec:msl}). We show how the hyperparameters can be used to control the confusion matrix in Section \ref{sec:hyperparam}.

\subsection{Computation time} \label{sec:comp_time} 
\begin{figure}[!h]
\centering
\includegraphics[scale=0.4]{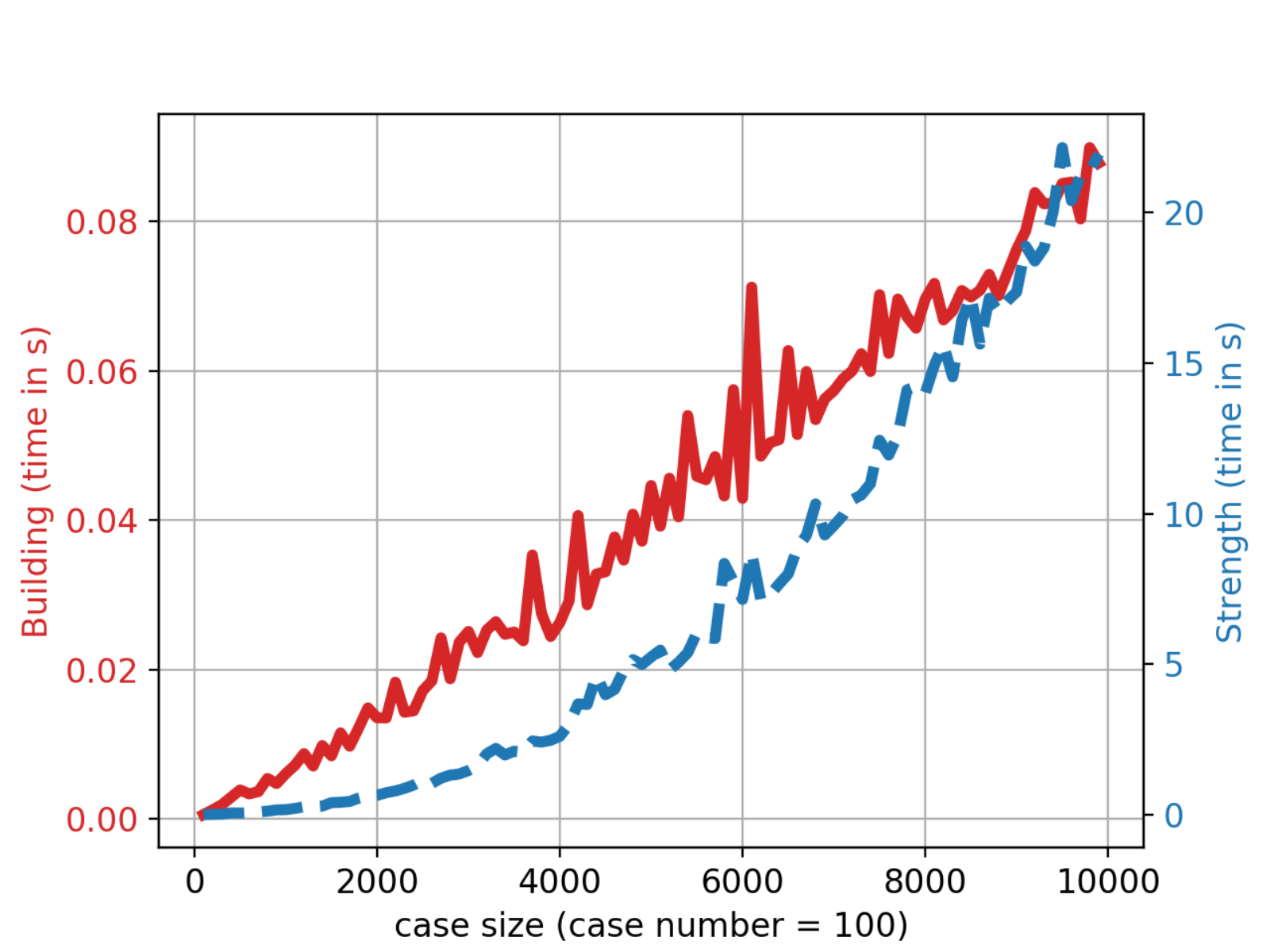}
\hfill
\includegraphics[scale=0.4]{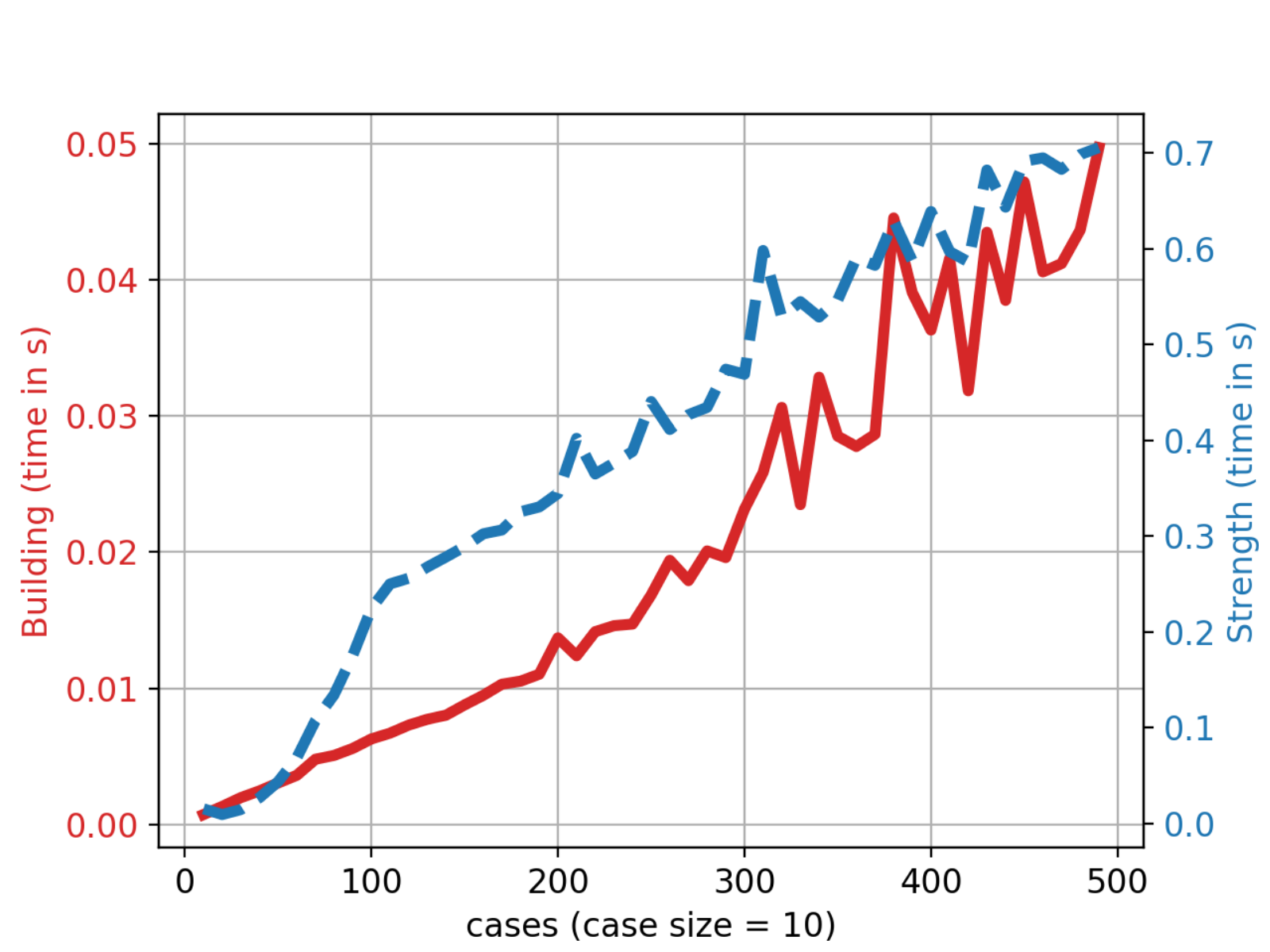}
\caption{On the left, computation time to build the model (hypergraph construction + strength calculation) depending on $N$ ($m = 10$), and on the right, depending on $m$ ($N = 100$). The case $i$ is described by $\{i, ...,  i+m\}$ s.t. each case is partitioned into $m$ elements (one discretionary feature).}
\label{fig:time}
\end{figure}

We generated a casebase of $N$ cases of size $m$ s.t. case $i$ is described by $\{i, ...,  i+m\}$ i.e., each case is partitioned into $m$ elements (one discretionary feature). This is the worst-case scenario in terms of the size of $\mathcal{E}$ if $m < N$ because the family grows exponentially in function of $m$ or $N$. We split the computation time into constructing the hypergraph (and determining the intersection family) and calculating the strength of the partition. The results are illustrated in Figure \ref{fig:time}. By increasing $N$ with a fixed $m$, the partition grows exponentially and thus, it is expected to have an exponential curve for the strength computation. On the contrary, building the hypergraph can be done in linear time when $m$ fixed. When $N$ is fixed and $m$ increases, constructing the hypergraph is still doable in linear time as expected. Interestingly, calculating the strength has two phases: if $m \leq N$, increasing $m$ exponentially increases the time (because $\mathcal{E}$ exponentially increases) but if $m > N$, increasing $m$ cannot results in an exponential growth in the computation time (because $\mathcal{E}$ grows linearly).

\subsection{Model space limitations} 
\label{sec:msl}
We now study the learning curves to show the model space limitations and propose some extensions left for future work. We also discuss the tradeoff between model locality and generalization.

\subsubsection{Learning curves}
\label{sec:lc} The learning curves are useful to study the limit of the model space on different datasets. It consists in plotting the accuracy in function of the training set size for both the training and the test sets. For the training set, it is expected to observe an accuracy starting close to 1 and decreasing fast to reach a plateau. A low stationary accuracy value indicates a high bias in the model or/and an irreducible error contained in the dataset, such as noisy or uninformative features. Oppositely, the accuracy on the test sets starts close to 0 as the training set is small and should increase until a plateau which is very often expected to be lower than the accuarcy of the training set. If the training set curve converges toward a much higher value than the test set curve, then the model has a large variance. In other words, to achieve a good bias-variance tradeoff, the accuracy of both curves should converge toward more or less the same value, expected as high as possible.
\begin{figure}[!h]
\centering
\includegraphics[scale=0.3]{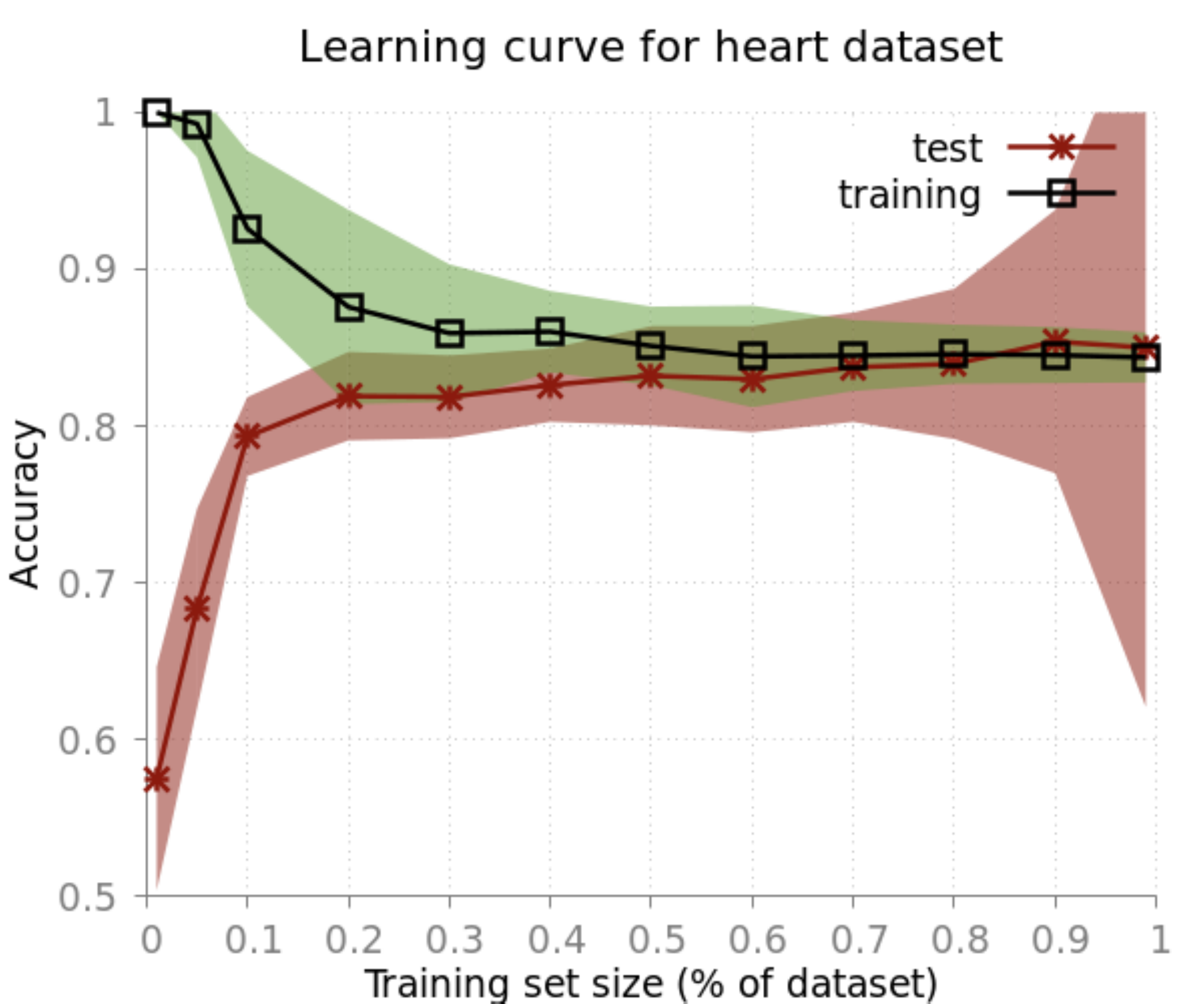}
\hfill
\includegraphics[scale=0.3]{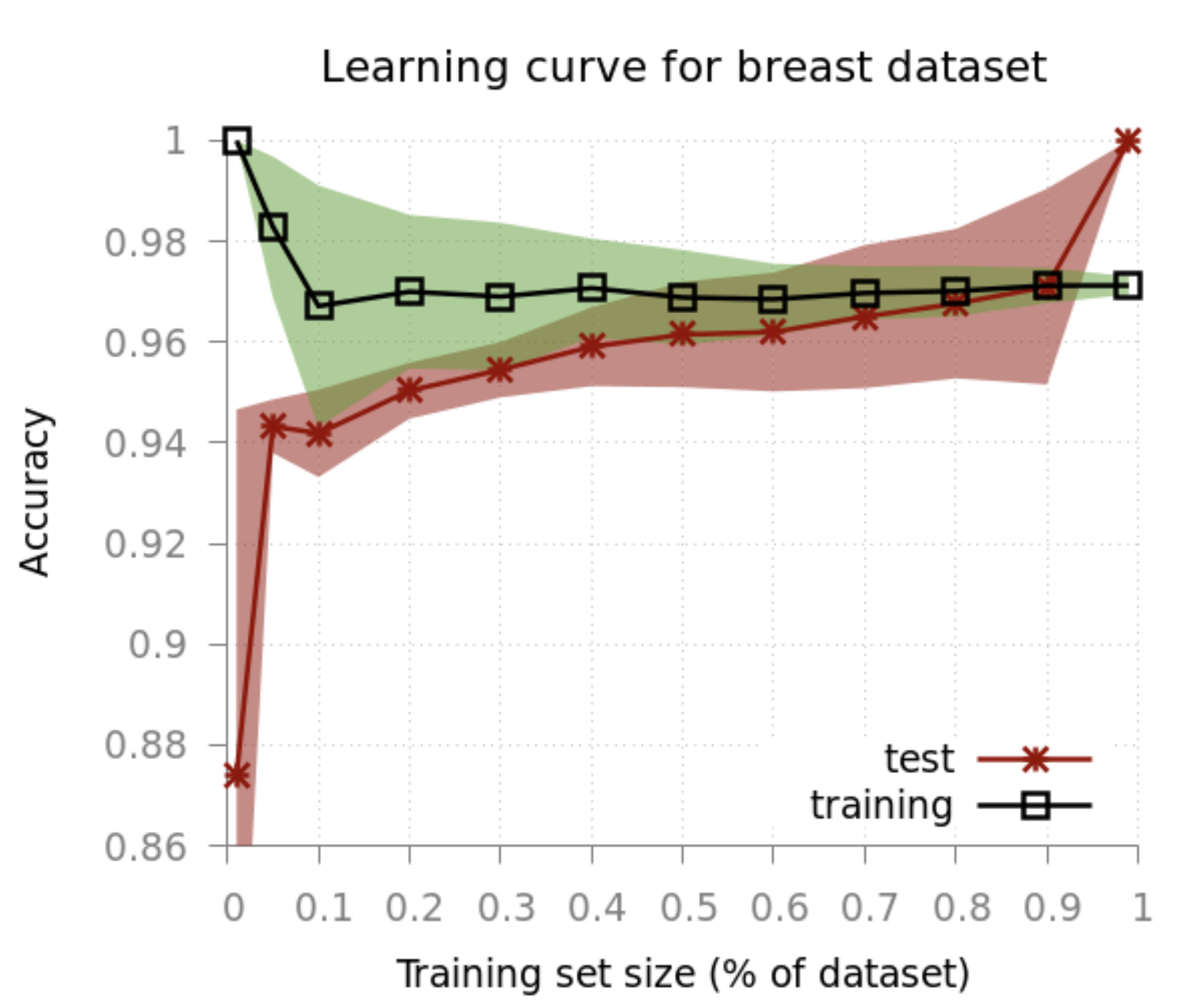}
\includegraphics[scale=0.3]{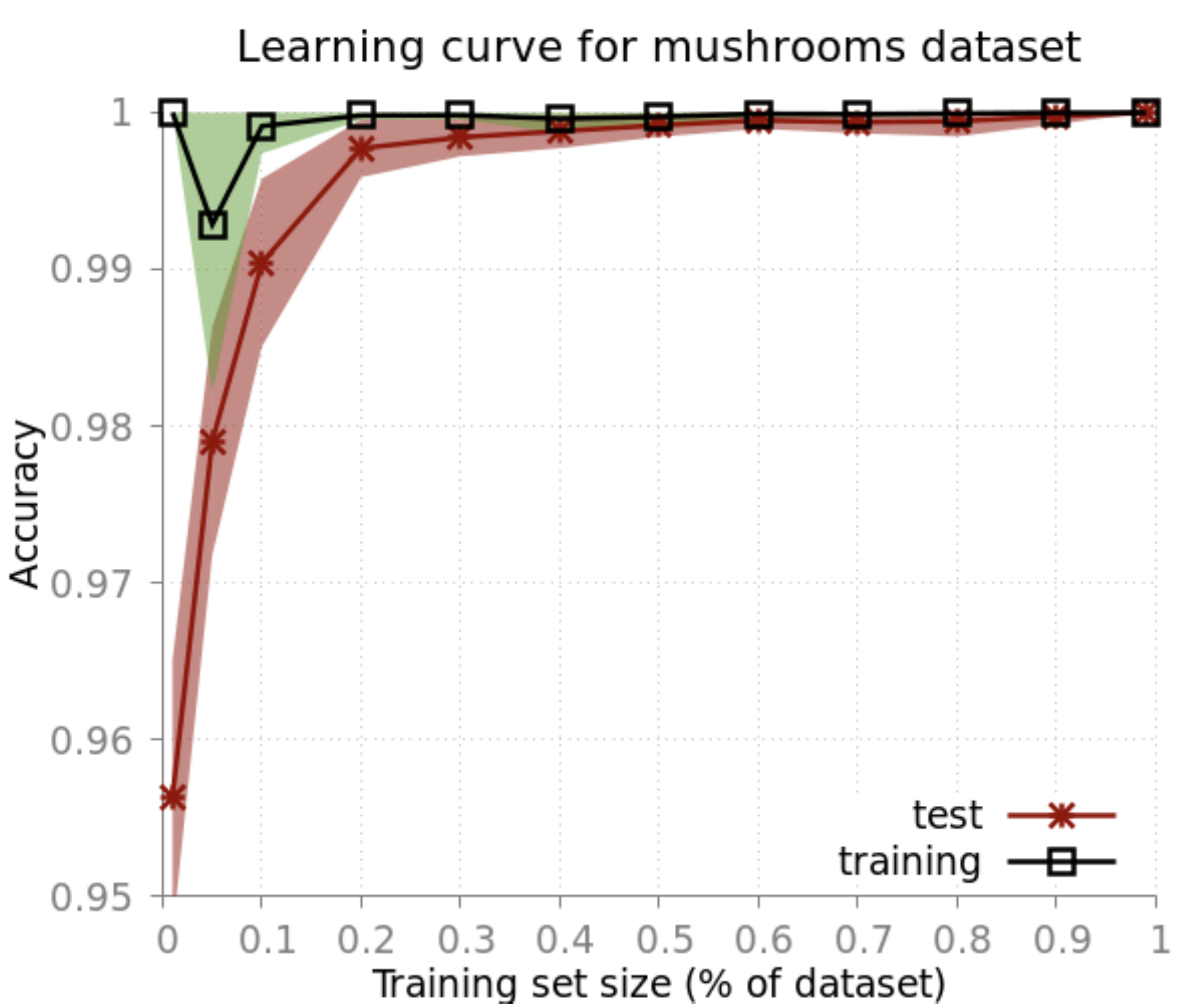}
\hfill
\includegraphics[scale=0.3]{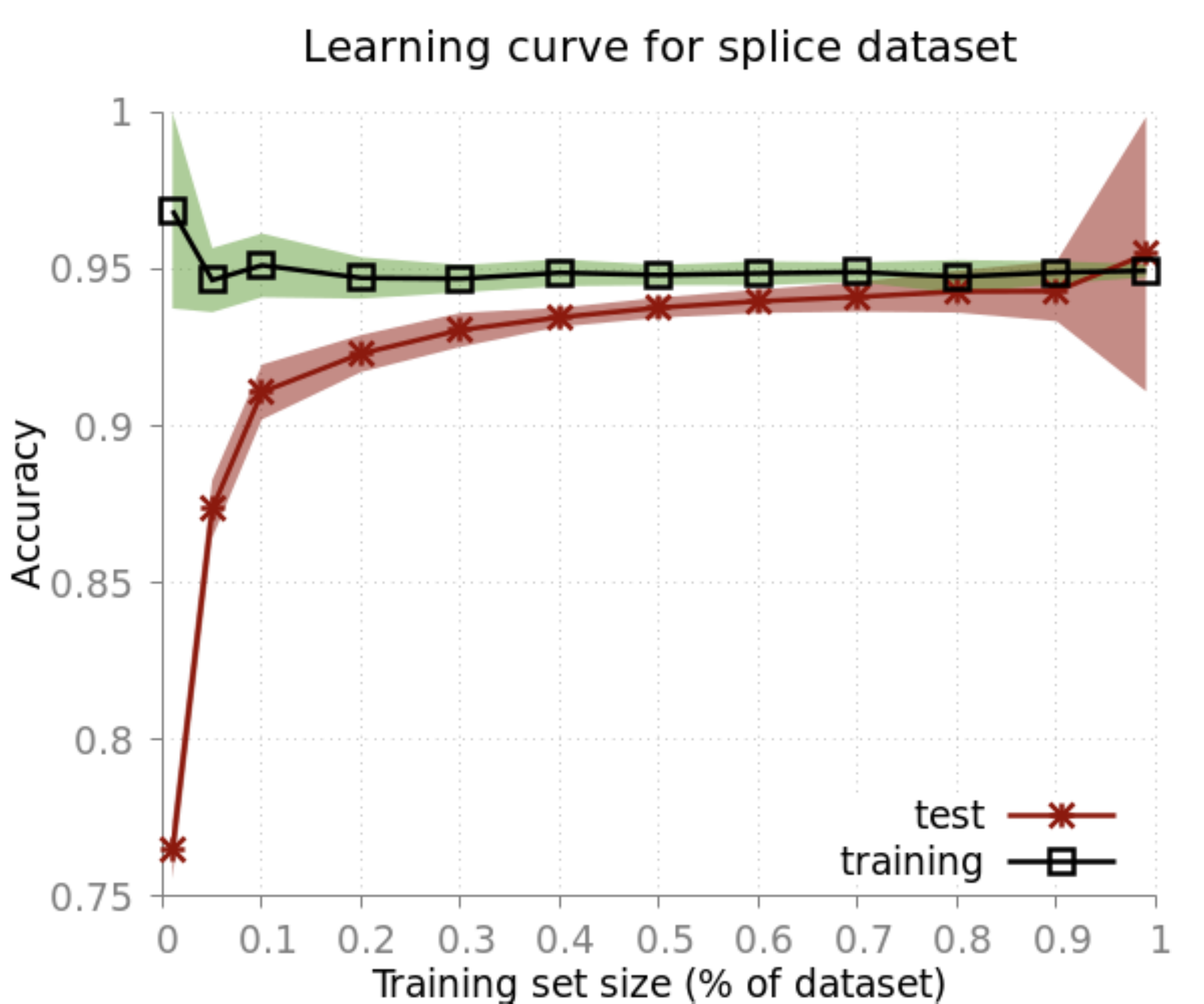}
\caption{Learning curves for \texttt{heart}, \texttt{breast}, \texttt{mushrooms} and \texttt{splice} datasets. Despite a bias and/or irreducible error observed on  \texttt{heart}, \texttt{breast} and \texttt{splice}. The model variance appears very low on all datasets. Conversely, the bias and/or irreducible error ranges from very low on \texttt{mushrooms}, relatively low on \texttt{breast} and \texttt{splice}, to high for \texttt{heart}.}
\label{fig:learning_curve_2}
\end{figure}

\begin{figure}[!h]\centering
\includegraphics[scale=0.3]{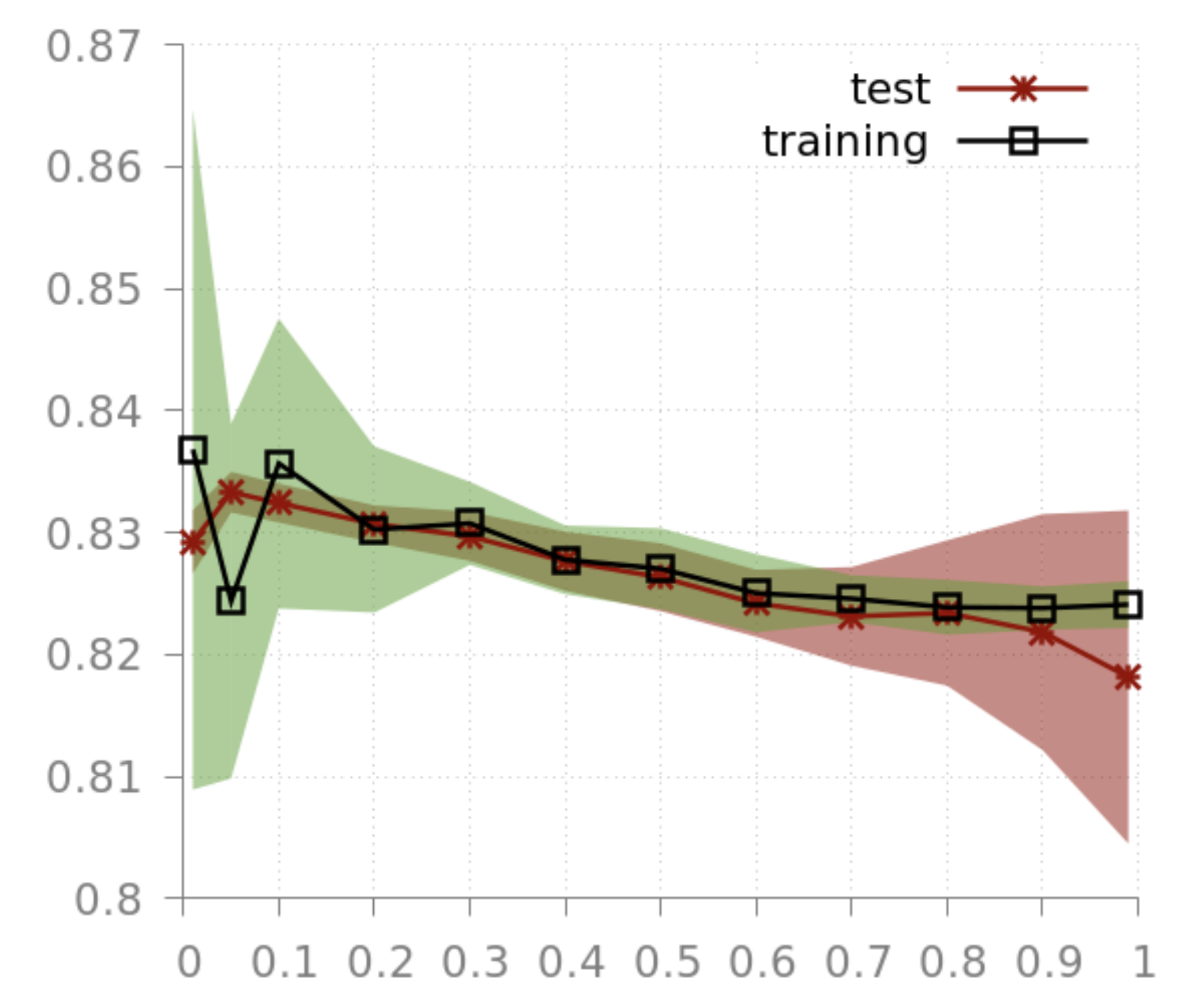}\hfill
\includegraphics[scale=0.3]{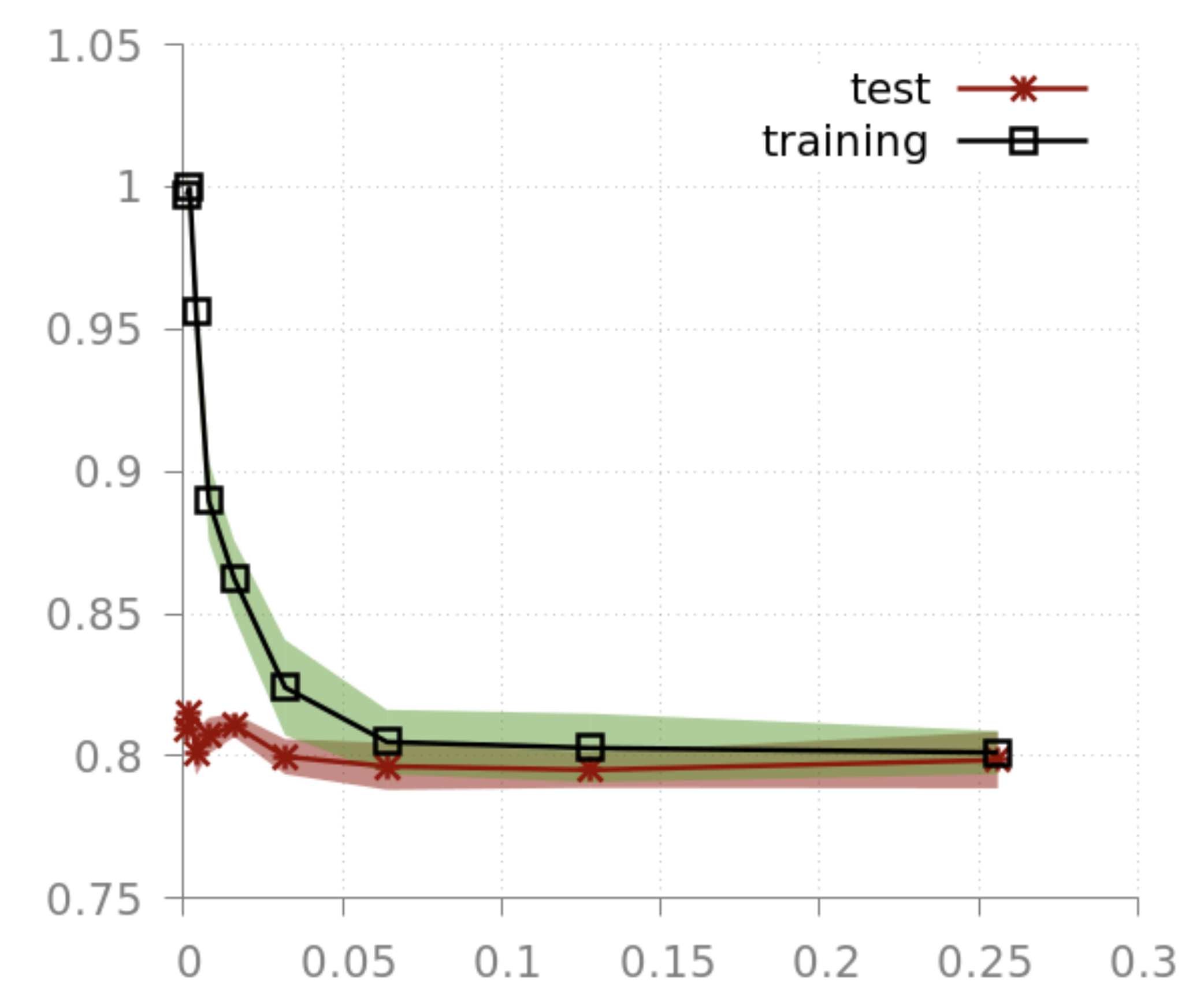}
\caption{Learning curve for \texttt{adult}. On the right, the training set size is restricted from 0 to 30\% of the dataset.}
\label{fig:learning_curve_1}
\end{figure}

The learning curves are calculated as the average over 10 runs with random splits and are shown in Figures \ref{fig:learning_curve_2} and \ref{fig:learning_curve_1}. 
The first remark concerns the variance. On all datasets, the variance is very low and the accuracy on the test set extremely close to accuracy of the training set. Despite a relatively low accuracy, \HCBR~seems to reach a good bias-variance compromise on \texttt{heart} suggesting a more complex model space might help. In general, the remark applies on the four datasets displayed in Figure \ref{fig:learning_curve_2} as the observed error rate results from bias and not only from irreducible error. Indeed, the literature comparison provided by Table \ref{table:prev_results} proves that the accuracy could be improved.

For \texttt{phishing} and \texttt{skin}, we arrive at the same conclusion. However, for those datasets, a heuristic is used during the prediction phase. We discuss its implication on the learning curves in the Additional Material, Section 6.
Finally, the learning curve of \texttt{adult} in Figure \ref{fig:learning_curve_1} shows once again a small variance but high bias.

The analysis of the learning curves indicates that the main limitation of \HCBR~lies in a large bias. This may come either from the model space complexity or the model selection method that does not properly fit the parameters. The number of parameters in the model is equal to the cardinality of $\mu$, itself proportional to the number of atoms in the training set. However, a closer look at how a decision is taken reveals that each case has its own {\it small model} as a convex combination of its features. In other words, the real number of parameters to model the decision for a given case never exceed its number of features. This might not be enough to represent fairly complex functions. To discard the second hypothesis, we conducted additional investigations on the model selection in the next section.

\subsubsection{Assessing model space limitations} We are interested in understanding whether it is possible for \HCBR~to overfit the training set or at least improve significantly the accuracy or MCC on the training set. We consider the matricial formulation of the support,
\begin{equation}
  \mathbf s = W\mu
\end{equation}
and we are interested in knowing if there exists a $\mu$ s.t. $\mathbf s$ leads to classify correctly all the examples. 
As we already verified that the initial $\mu$ provides better results than baseline models, we would like to perturbate as little as possible $\mu$. For this, we calculate $\mathbf k$ s.t. $\mathbf s + \mathbf k$ would correctly classify all the examples. We are looking for $\delta$ s.t.,

\begin{align}
 & \mathbf s + \mathbf k  = W(\mu + \delta) \\
\Leftrightarrow & \mathbf s + \mathbf k  = W\mu + W\delta \\
\Leftrightarrow & \mathbf k  = W\delta \\
\implies & \delta  = W^+ \mathbf k + [I - W^+W]\mathbf w, ~ \forall \mathbf w \in \mathbb{R}^N
\end{align}
with $W^+$ the Moore-Penrose pseudo-inverse of $W$. When $W$ is not of full rank, the solutions of the undetermined system are given for any vector $\mathbf w \in \mathbb{R}^N$, however, it can be shown that $\delta  = W^+\mathbf k$ is a least-square minimizer, i.e. $\forall \mathbf x \in \mathbb{R}^M ~ ||W\mathbf x -\mathbf k||_2 \geq ||W\delta - \mathbf k||_2$. In particular, if $||W\delta - \mathbf k||_2 = 0$, then all the elements would be correctly classified.

Of course, it might not be possible to obtain  $\delta$ s.t. $||W\delta - \mathbf k||_2 = 0$, but it does imply that there exist no couple $(\mathbf k, \delta)$ s.t. the accuracy is 1. Solving directly for all such couples seem to be a difficult problem without additional assumptions and thus, we adopted a slightly different approach: instead of fixing $\mathbf k$ a priori, we formulated the problem as optimizing the Matthew Correlation Coefficient:

\begin{equation}
\label{eq:opti_MCC}
\delta^* = \underset{\delta \in \mathbb{R}^M}{\max} ~ \text{MCC}(\delta, \mu, W) - c ||\delta||^2_2
\end{equation} where MCC is the Matthew Correlation Coefficient associated to $\mathbf s = W \delta$ and $c$ a regularization factor. The regularization factor translates the idea that a smaller perturbation to obtain the same MCC is better than a larger one. We arbitrarily set $c$ to $0.1$. Despite further work could be needed to determine a more tailored value, the conclusions drawn from this section would remain valid.

To solve \eqref{eq:opti_MCC}, we used a $(\mu+\lambda)$ genetic algorithm with an evolution strategy. Each individual is made of two vectors: $\delta \in \mathbb{R}^M$ and a $\nu \in \mathbb{R}^M$ representing the mutation probability of each corresponding component of $\delta$. The mutation operator is a centered gaussian perturbation and the crossover a 2-points crossover. The implementation is provided by DEAP \cite{DEAP_JMLR2012}.

We performed 10 runs of a 10-fold cross-validation with random split and compared to the result obtained without the optimization process. The dataset \texttt{mushrooms} has been discarded as \HCBR~ reached an accuracy of 1. We set the population to 100, and for each dataset, we adjusted the number of generations and the standard deviation of the gaussian mutation manually (see the Additional Material, Section 7). To set the standard deviation of the mutation, we used $\sigma = \frac{\mu^-}{\alpha}$ where $\alpha$ is an factor determined empirically, and $\mu^-$ defined by $\underset{i,j}{\min}{~ \mu_i - \mu_j}$ i.e. the minimal difference between two components of $\mu$. The rationale behind is that, once again, we would like to slightly perturbate $\mu$ and it is reasonable to think that a perturbation should be small enough not to directly switch the estimation of two elements of $\mu$. To conclude, we used a Wilcoxon signed-rank test at 5\% and 1\% on the test MCC obtained with and without the optimization process. There are three possible scenarios: 1) if the MCC can be improved on the training set and on the test set, it means the problem comes from the model selection method (estimation of $\mu$ and training) and results might be improved without changing the model space, 2) if the MCC can be improved on the training set but remains the same or deteriorate on the test set, the model space can represent the training set but overfits, and thus, should be extended 3) if the MCC cannot be improved even on the training set, then the model space is definitely not capable of representing properly the underlying decision mapping and should be extended.

\begin{figure}[!h]\centering
\includegraphics[scale=0.2]{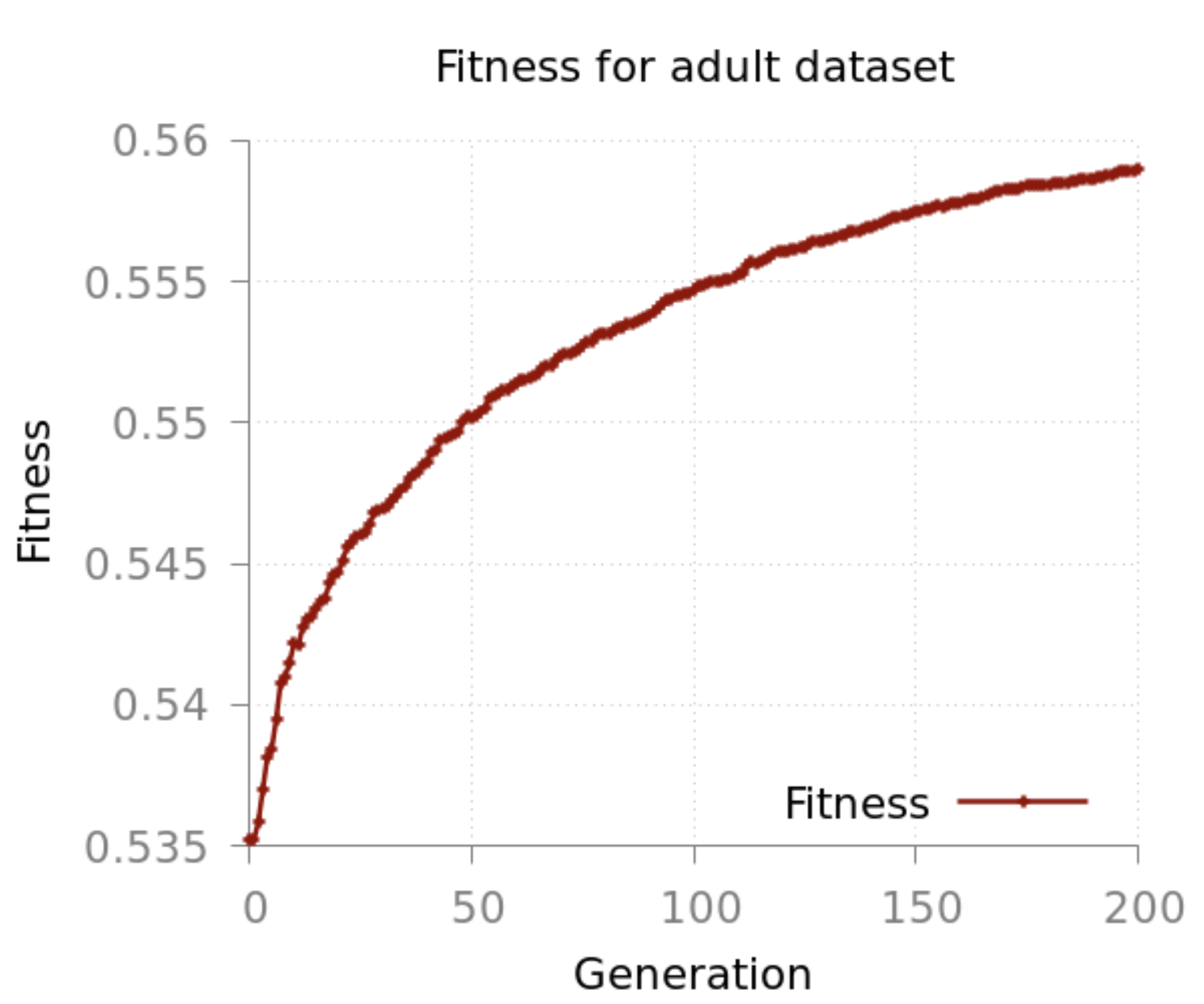}
\hfill
\includegraphics[scale=0.2]{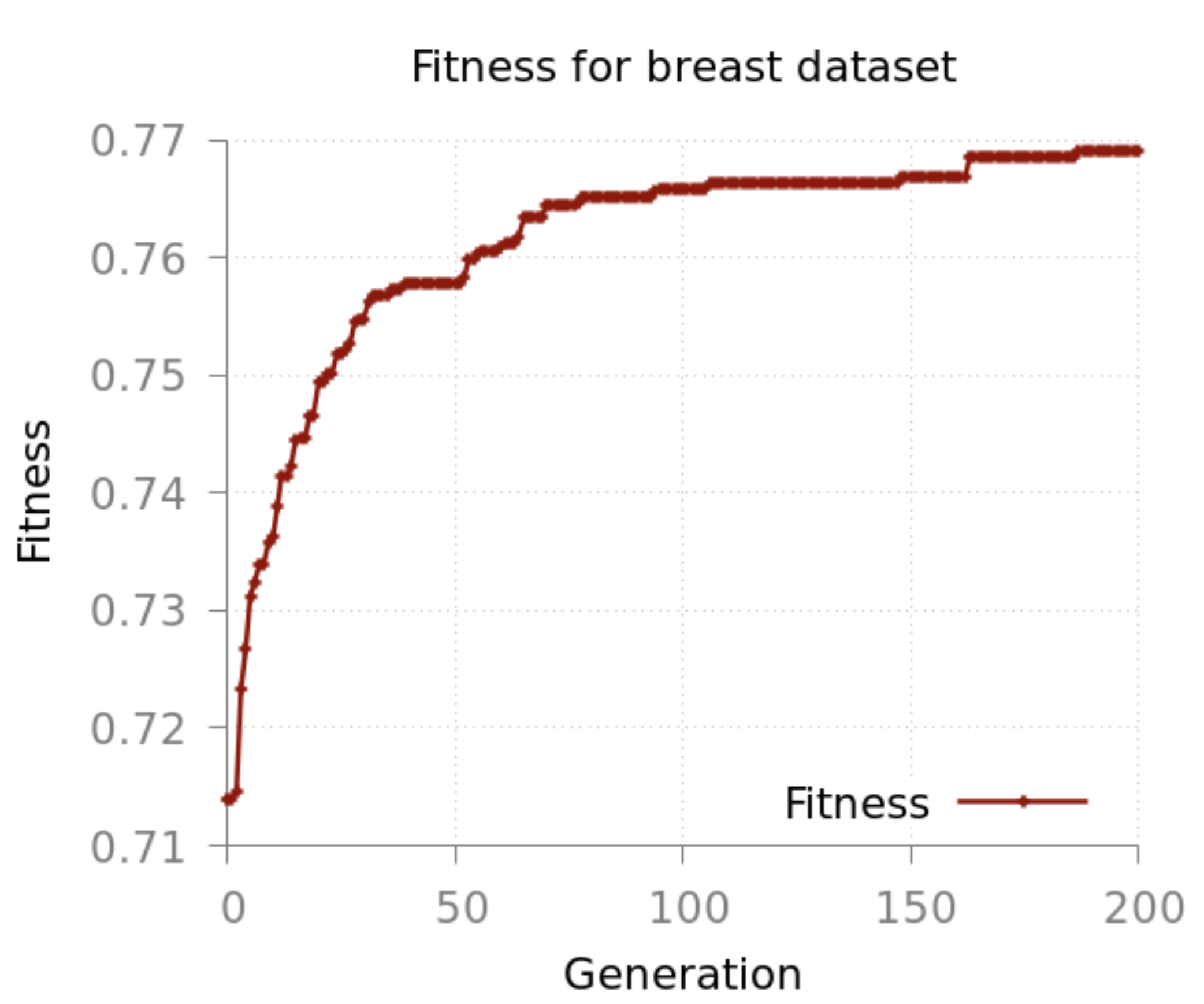}
\hfill
\includegraphics[scale=0.2]{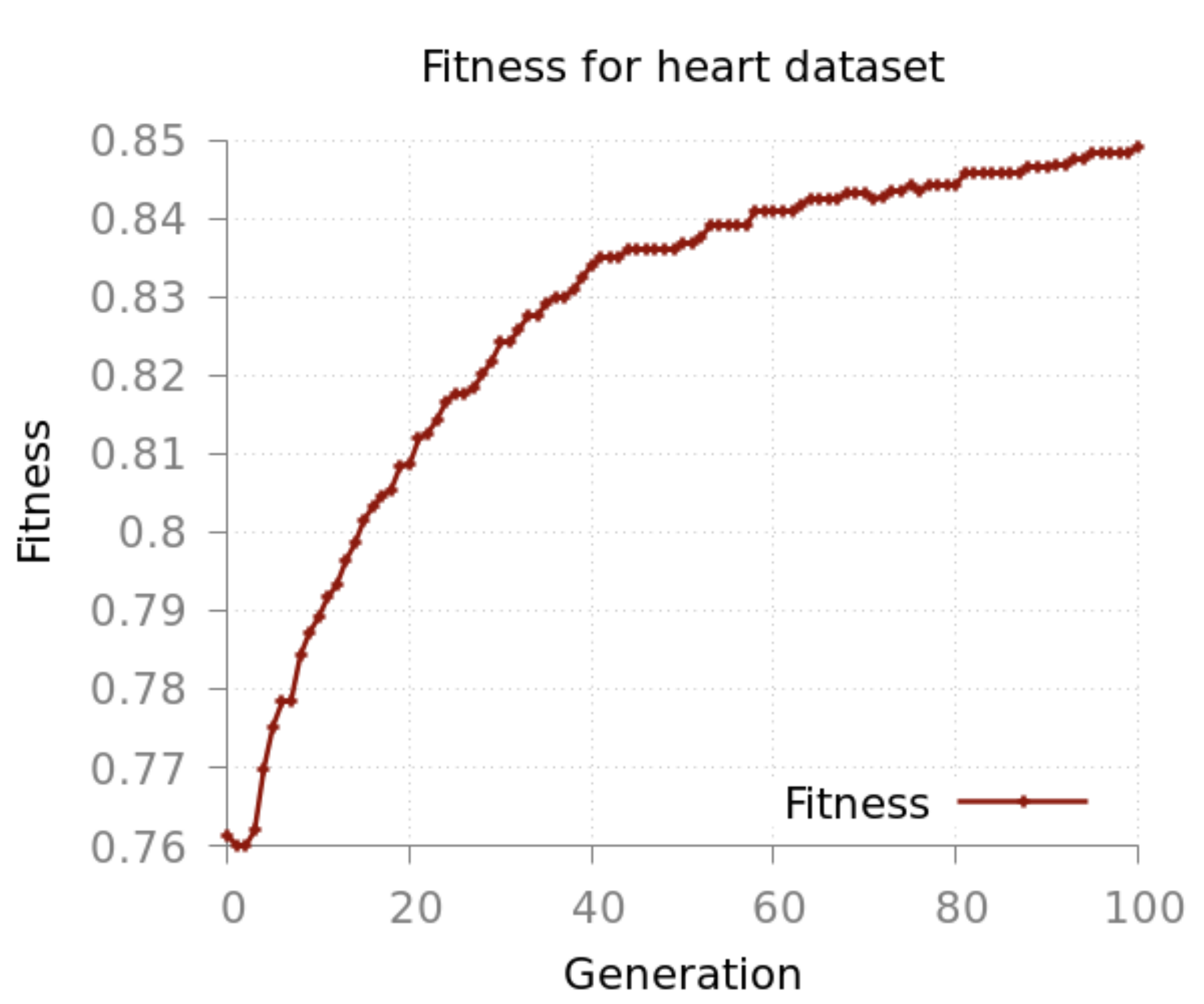}
\hfill
\includegraphics[scale=0.2]{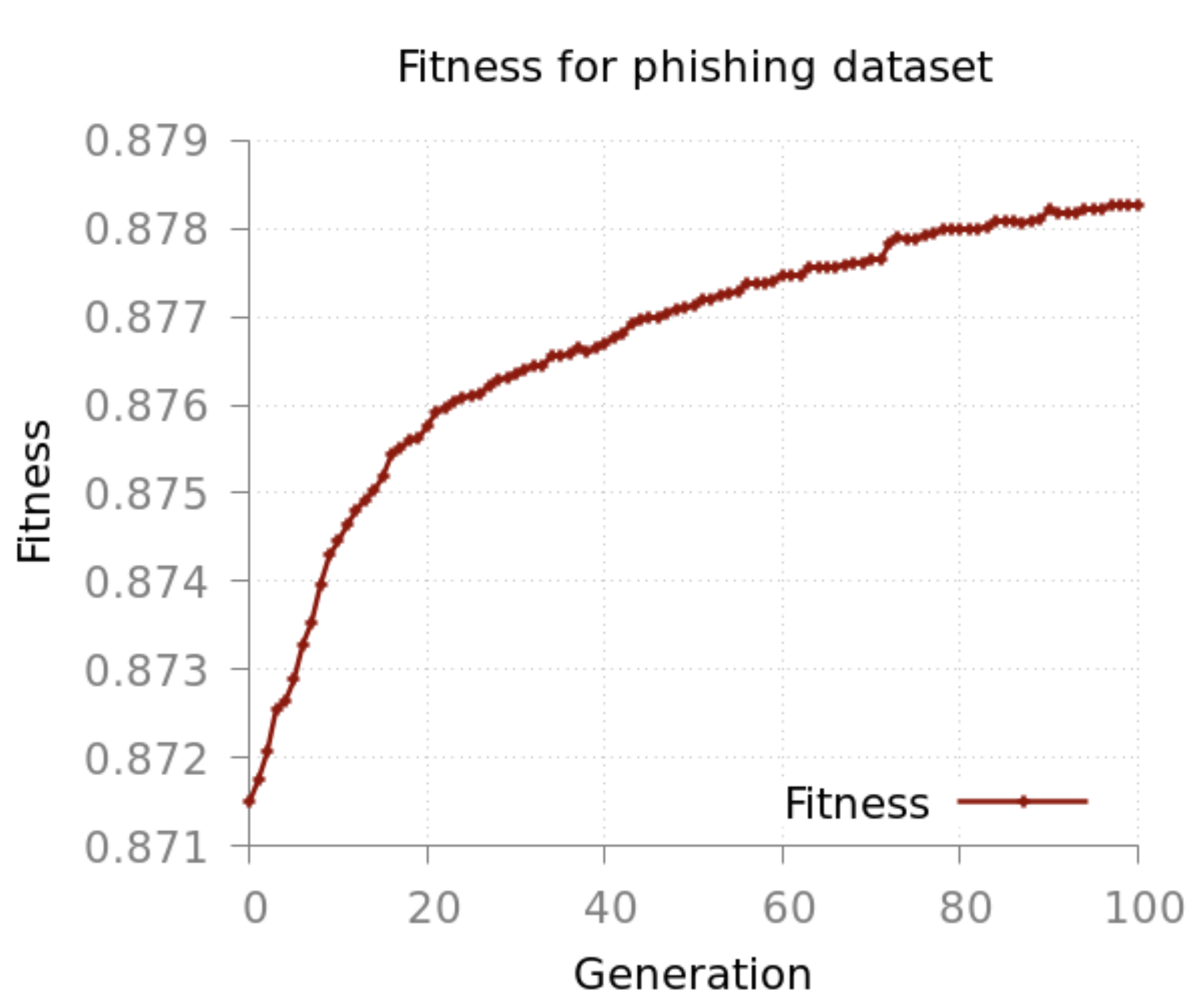}
\hfill
\includegraphics[scale=0.2]{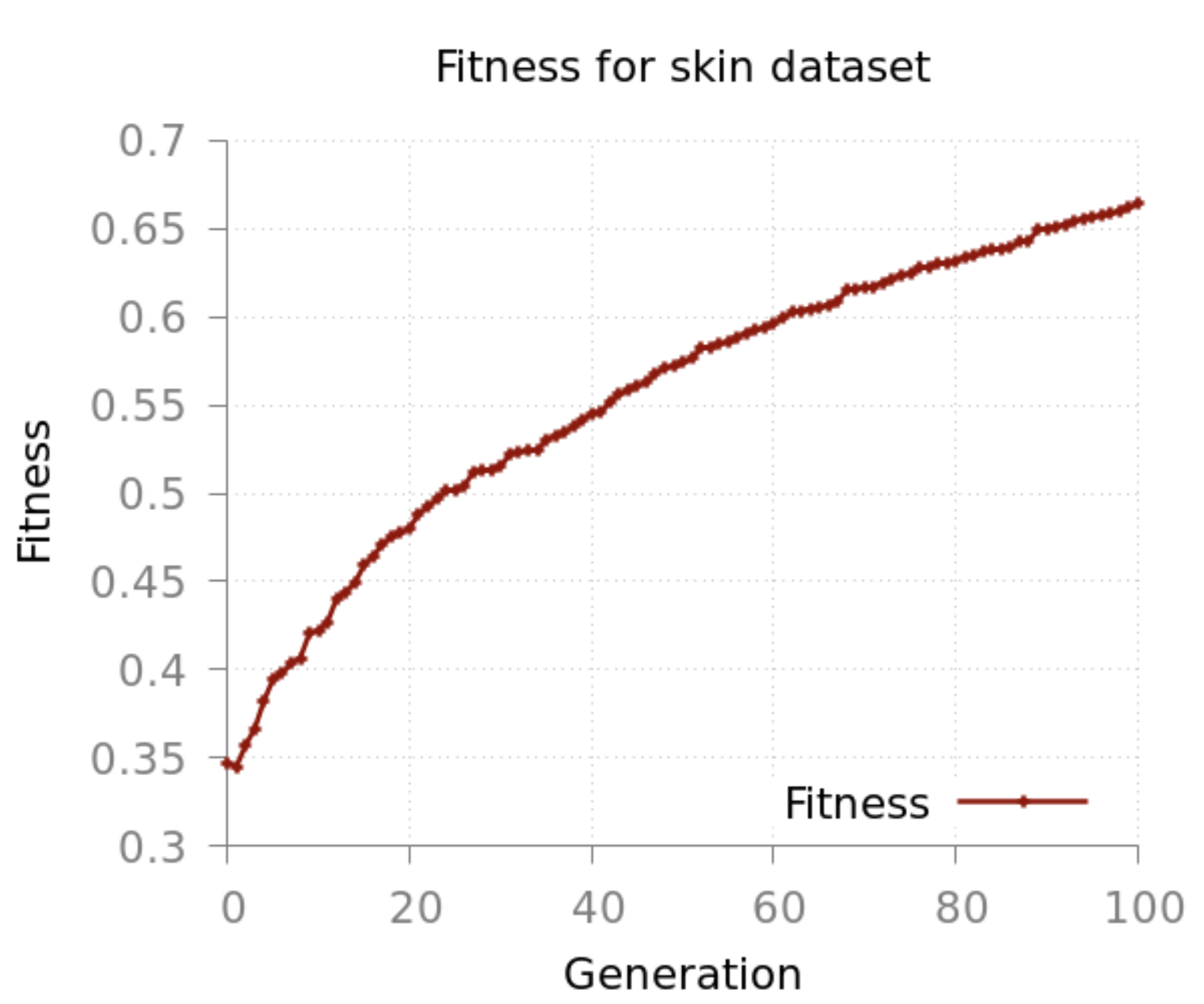}
\hfill
\includegraphics[scale=0.2]{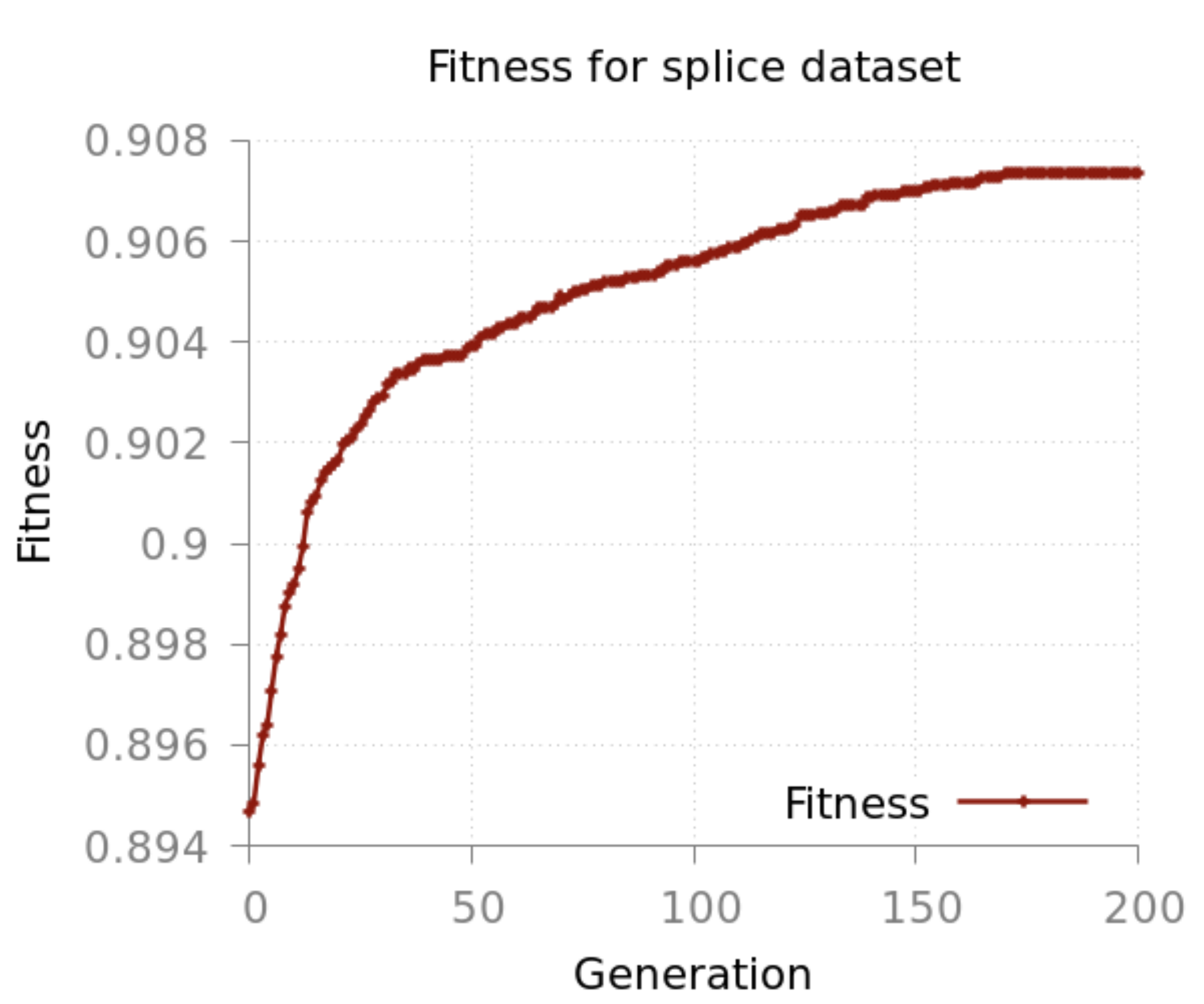}
\caption{Maximal fitness value in the population in function of generations for \texttt{adult}, \texttt{breast},  \texttt{heart},  \texttt{phishing},  \texttt{skin} and \texttt{splice}. The fitness is lower than the MCC by definition of \eqref{eq:opti_MCC}.}
\label{fig:fitness}
\end{figure}

A summary of the results are provided in Table \ref{table:opti_results} and the evolution of the fitness by Figure \ref{fig:fitness}. A look at Figure \ref{fig:fitness} confirms that the genetic algorithm converged or was closed to converged and was able to optimize the cost function. In all cases except \texttt{skin}, the optimization procedure succeeded to find a vector $\delta$ that yields a better MCC on the test sets. However, the improvement is quantitatively different from a dataset to another. For instance, on \texttt{heart} the absolute difference in MCC is 16\% (relative difference: 23.64\%) while on \texttt{phishing} the difference is barely 1\%. In general, the higher is the initial MCC and the less the improvement is visible. 

The variations of MCC on the test set are mitigated. The procedure returns significant changes in only two cases (\texttt{adult}) and (\texttt{skin}) for which one is improved (\texttt{adult}) and one deteriorated (\texttt{skin}). It indicates that the model selection and training phase described in Section \ref{sec:model_selection} and \ref{sec:decision_training} return one of the best MCC achievable within the model space. More precisely, the vector $\mu$ represents one of the best support approximation in its neighborhood. Notice that for \texttt{skin}, the optimization process deteriorated the MCC on the training set. A smaller mutation factor might help, however we believe this would barely change the results and thus entail the conclusion on the model space limitation.

\begin{table}[h!]
\begin{center}
  \caption{Results obtained on solving \eqref{eq:opti_MCC}. The columns {\it initial MCC}, {\it $\Delta$ MCC training} and {\it$\Delta$ MCC test} represent respectively the initial MCC on the training set, the difference of MCC obtain on the training with and without the genetic algorithm, the difference of MCC obtain on the test sets with and without the genetic algorithm. {\it WSR r\%} indicates the result of the Wilcoxon signed-rank test at $r$\% risk (yes for significant difference in the sample median, no otherwise).}
  \begin{small}
\begin{tabular}{|l|c|c|c|c|c|}
\hline
  Dataset & initial MCC & $\Delta$ MCC training & $\Delta$ MCC test & WSR 5\% & WSR 1\%\\ \hline
 \texttt{adult} &  0.5190 & 0.0400 & 0.0084 & yes & yes \\
  \texttt{breasts} & 0.9360 & 0.0312 & -0.0025 & no & no \\
  \texttt{heart} & 0.6912 & 0.1634 & -0.0023 & no & no  \\
    \texttt{phishing} & 0.8690 &0.0093 & 0.0002  & no & no  \\
    \texttt{skin} & 0.8432 & -0.0242 & -0.0118 &  yes & yes \\
    \texttt{splice} & 0.8866 & 0.0208 & 0.0006 & no & no  \\\hline
\end{tabular}
\end{small}
  \label{table:opti_results}
\end{center}
\end{table}

\subsection{Hyperparameters $\eta$ to control the accuracy} \label{sec:hyperparam}

We showed in Section \ref{sec:result_literature_comp} that the hyperparameters can increase the overall performances while the impact is limited by the model space. In this experiment, we show how $\eta$ can be used to control the accuracy by specifying a threshold on the risk associated to a prediction.

\begin{figure}[!h]
\centering
\includegraphics[scale=0.3]{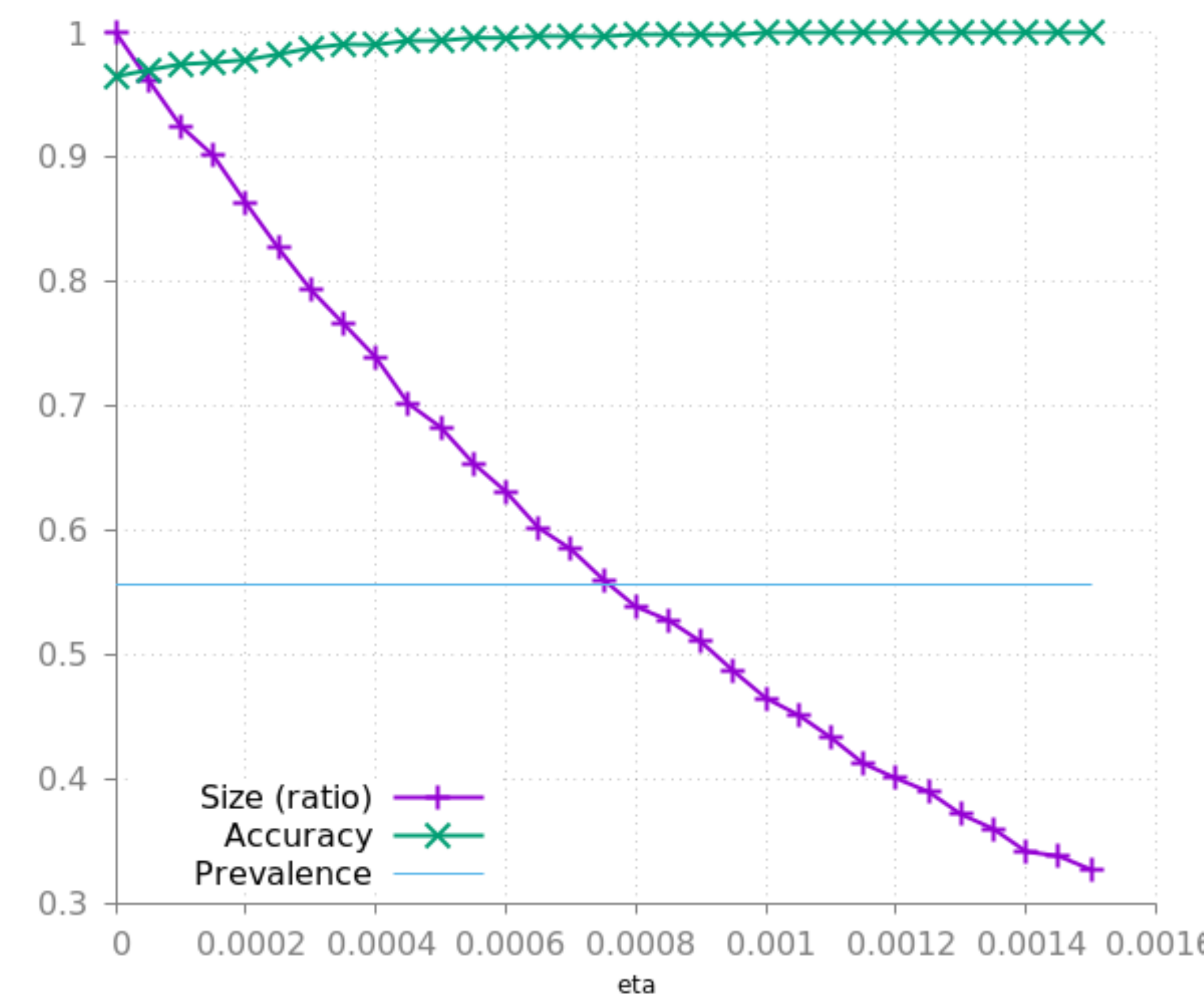}
\hfill
\includegraphics[scale=0.3]{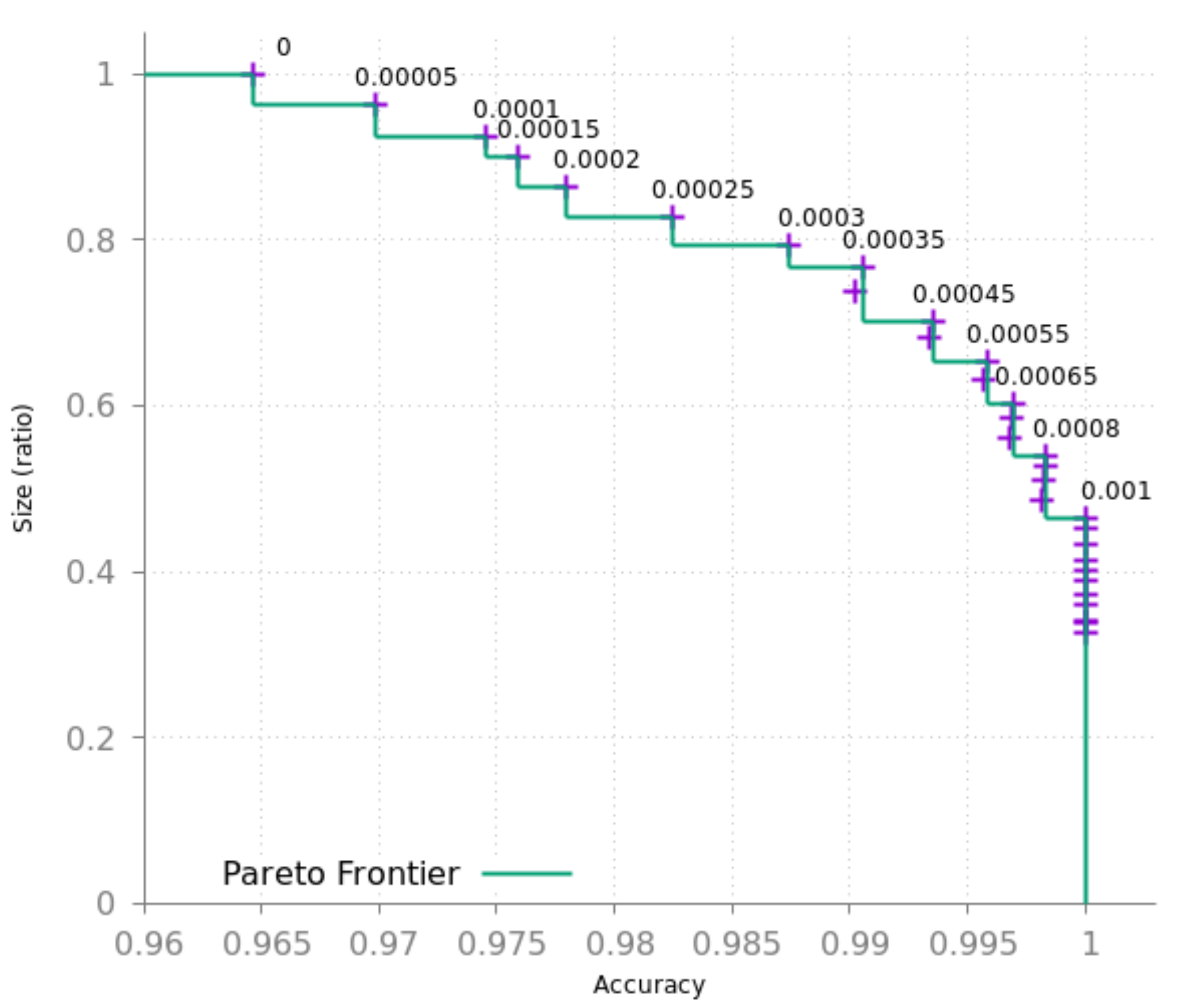}
\caption{Influence of $\eta$ on \texttt{phishing} dataset.}
\label{fig:meta_phishing}
\end{figure}

\begin{figure}[!h]\centering
\includegraphics[scale=0.3]{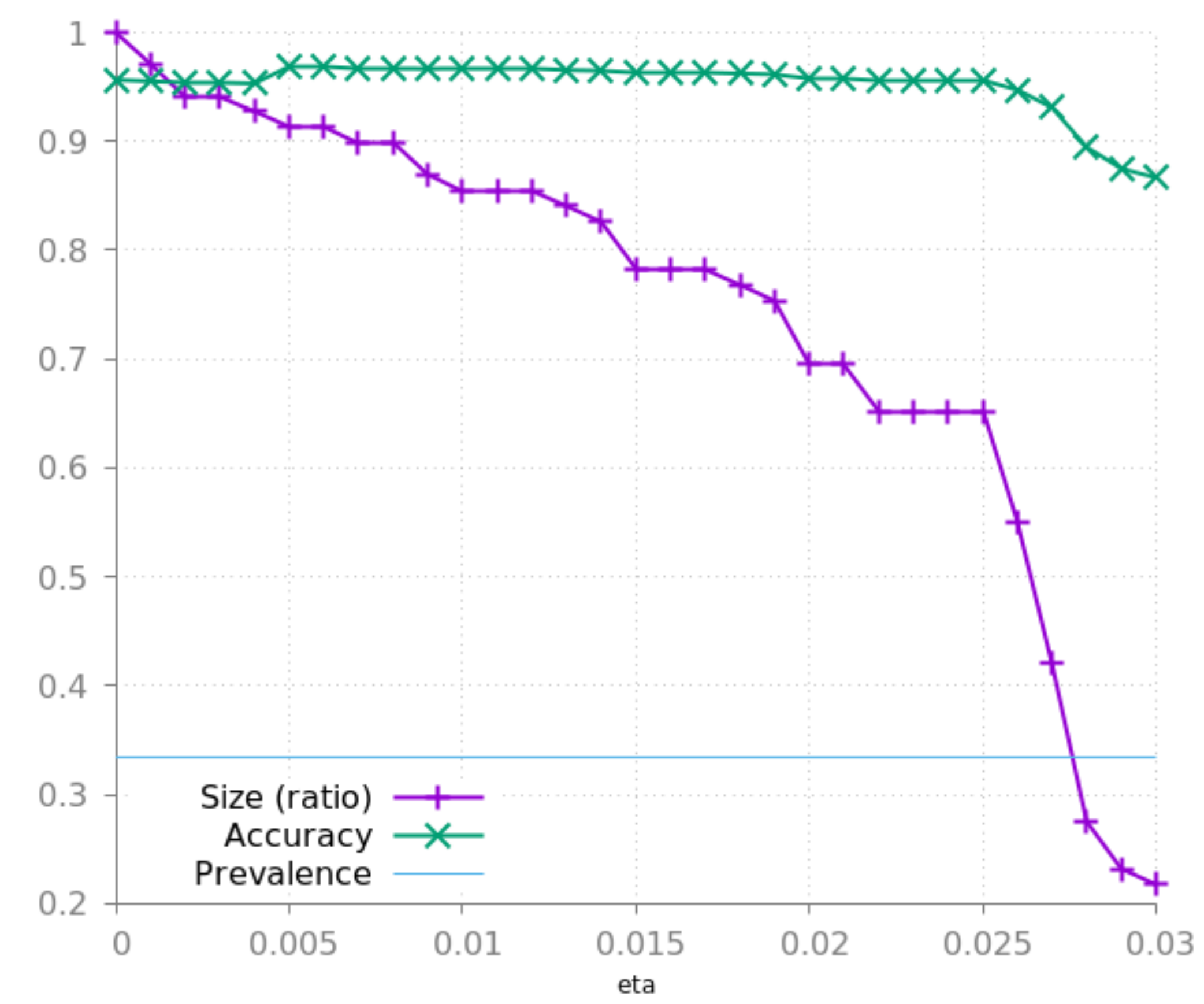}
\hfill
\includegraphics[scale=0.3]{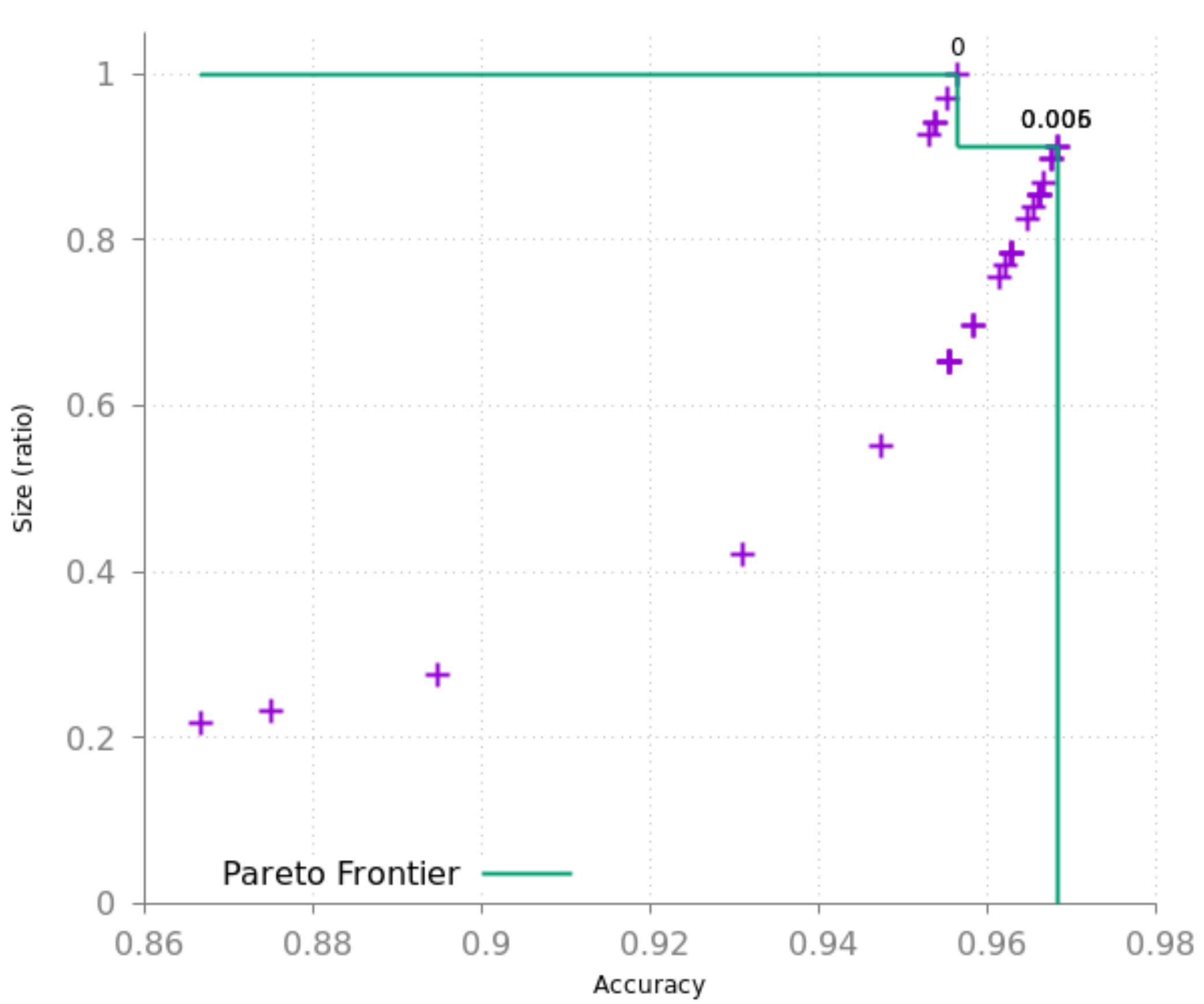}
\caption{Influence of $\eta$ on \texttt{breast} dataset.}
\label{fig:meta_breast}
\end{figure}

\begin{figure}[!h]
\centering
\includegraphics[scale=0.3]{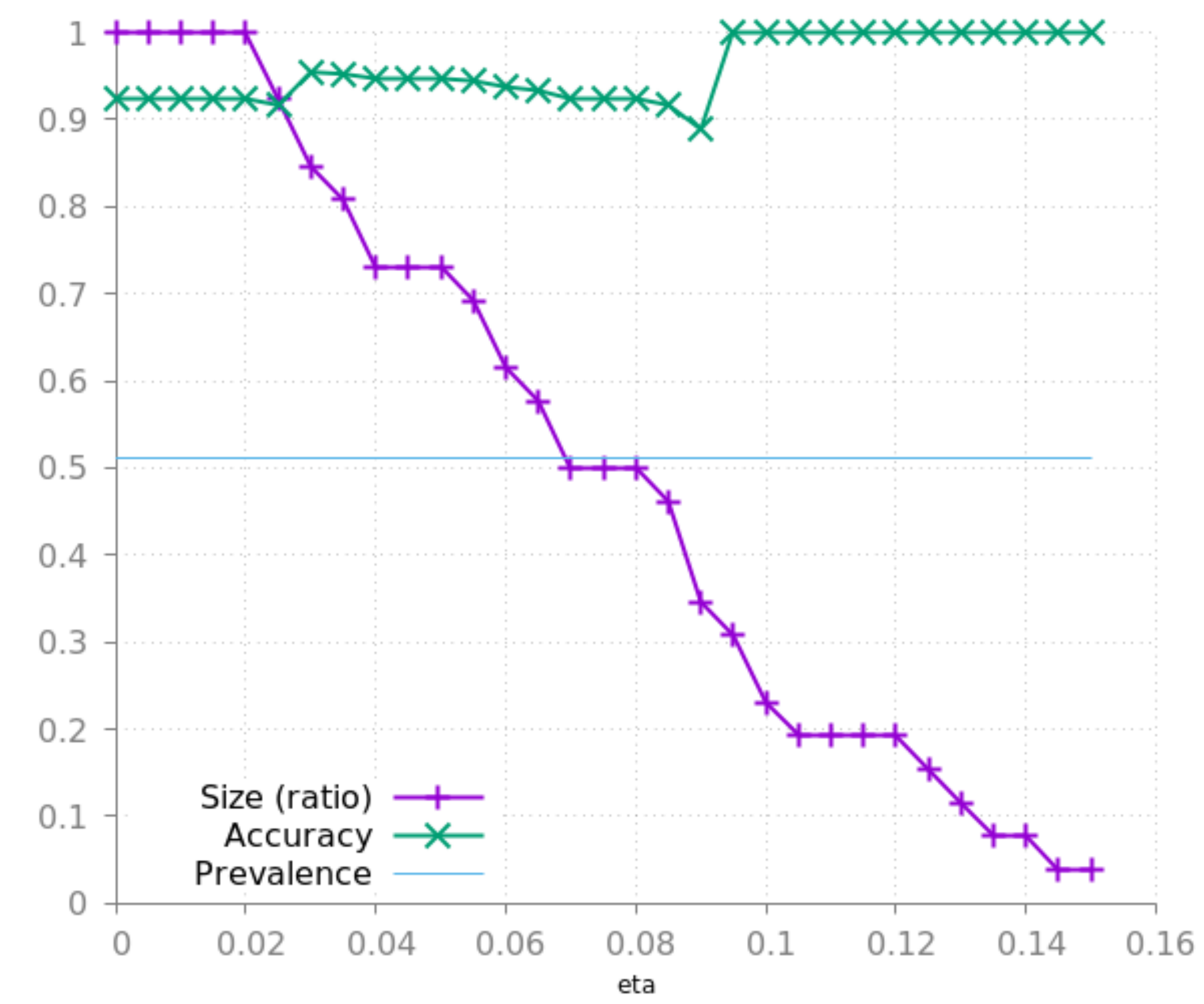}
\hfill
\includegraphics[scale=0.3]{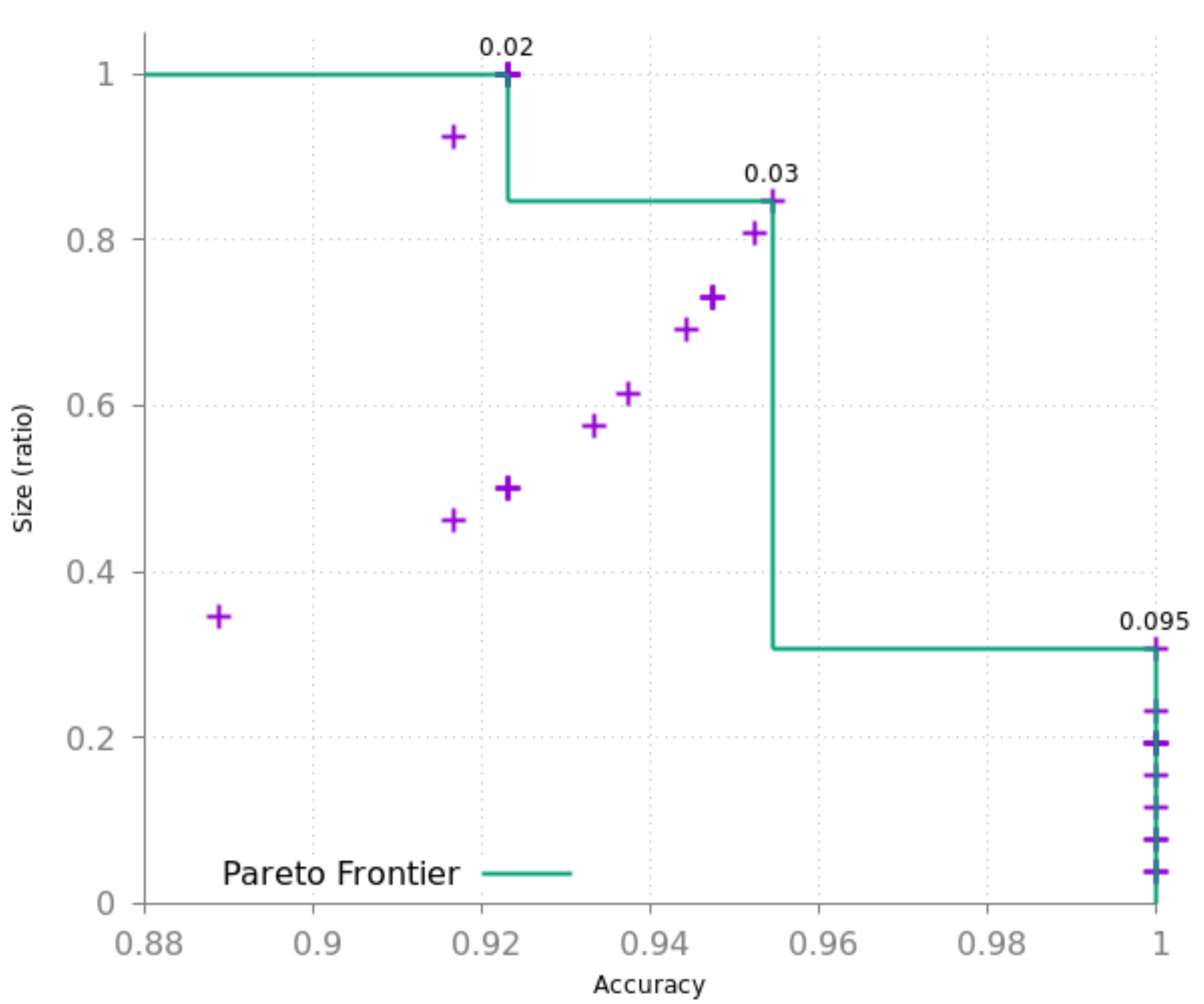}
\caption{Influence of $\eta$ on \texttt{heart} dataset.}
\label{fig:meta_heart}
\end{figure}

\begin{figure}[!h]\centering
\includegraphics[scale=0.3]{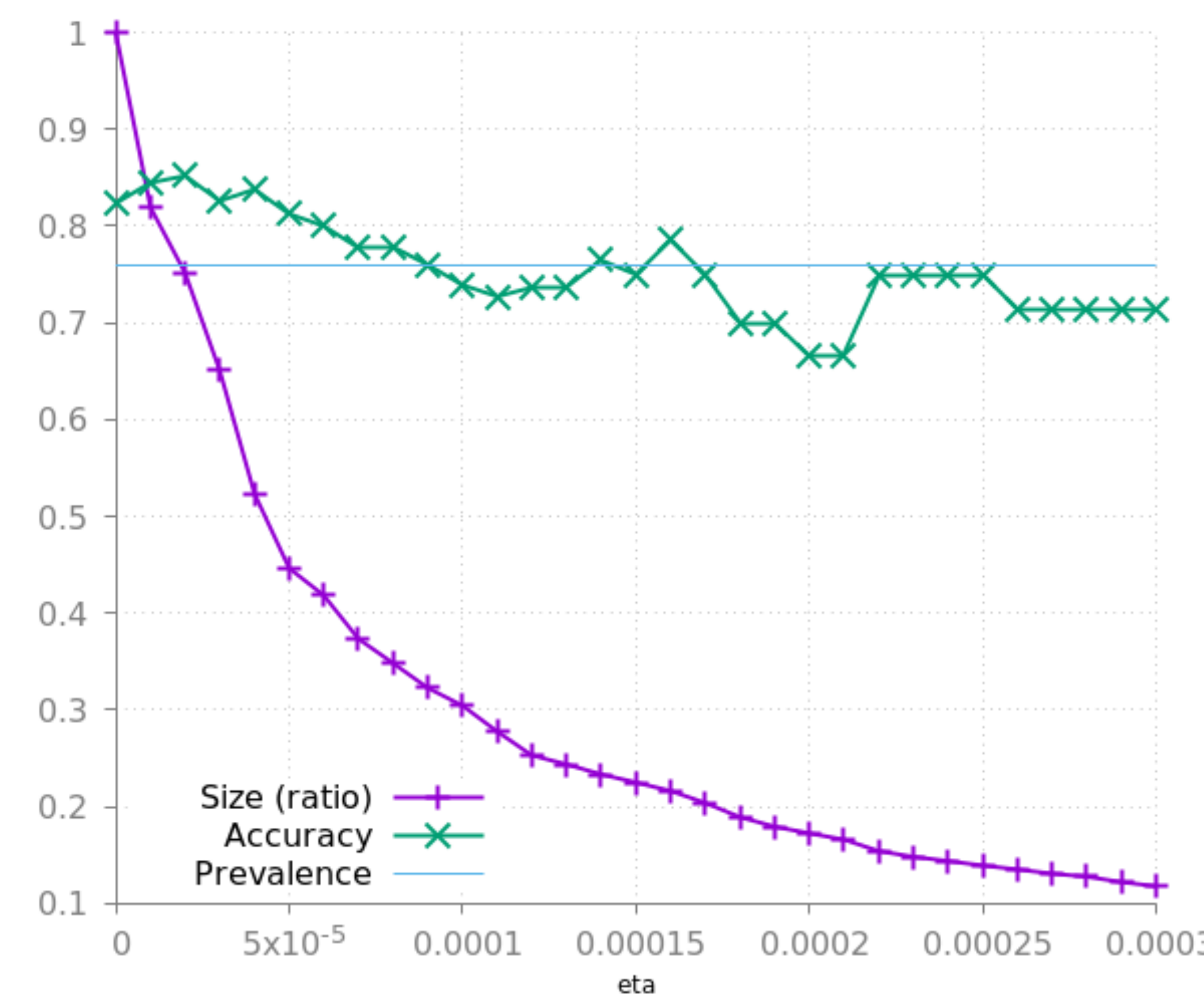}
\hfill
\includegraphics[scale=0.3]{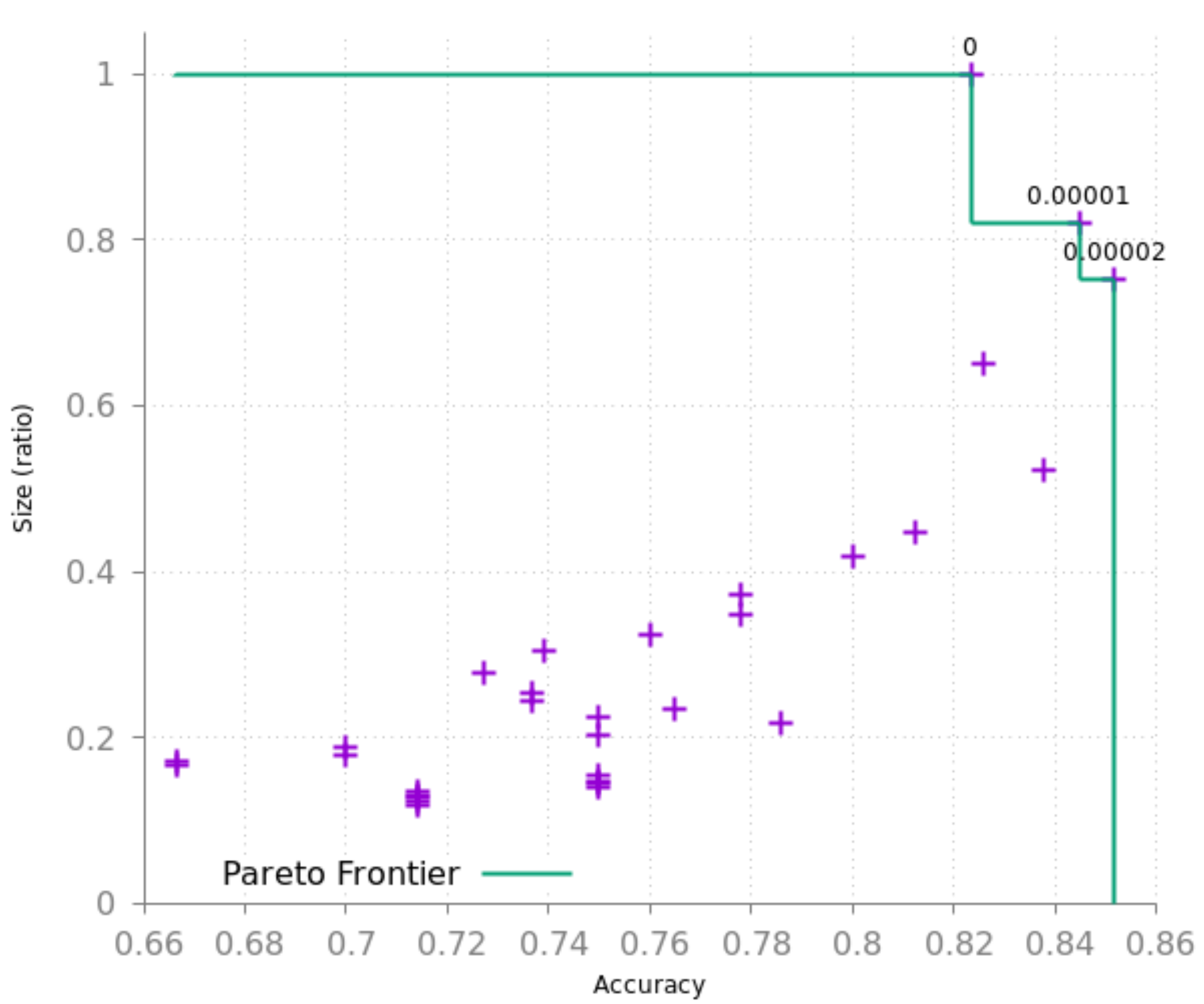}
\caption{Influence of $\eta$ on \texttt{adult} dataset.}
\label{fig:meta_adult}
\end{figure}

\begin{figure}[!h]\centering
\includegraphics[scale=0.3]{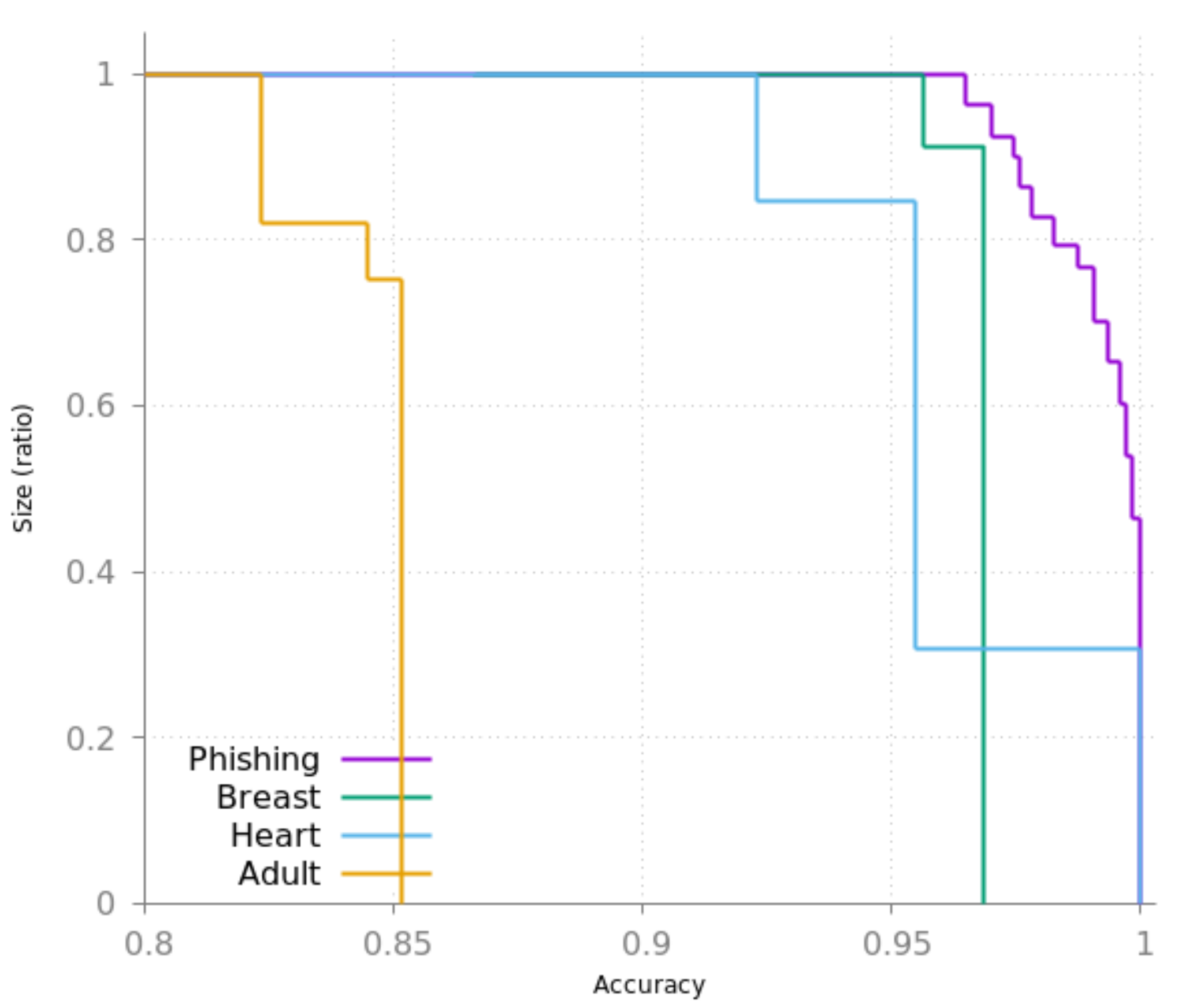}
\caption{Pareto Frontiers comparison.}
\label{fig:meta_all}
\end{figure}

We used a 90-10 split and set $\eta_{-1} = \eta_1$ to ease the visualization. Instead of using the decision function defined by \eqref{eqn:updated_cr}, we did not produce a prediction if the constraints $C_1$ or $C_0$ were not respected. It can be viewed as creating a third class {\it unknown} for which we consider \HCBR~cannot produce a decision. We measured the accuracy and the test set size ratio for which a prediction has been produced for different values of $\eta := \eta_{-1} = \eta_1$. If the model correctly approximates the underlying mapping function $J$, increasing $\eta$ should increase the accuracy while the test set ratio should remain high. Additionally, we plot the test set ratio in function of the accuracy and calculate the Pareto frontier\footnote{Points s.t. improving one component would deteriorate  the other one.} which represents the best compromises accuracy/ratio. The closer the points are to $(1,1)$ the better it is. A Pareto frontier consisting of $(1,1)$ represents the perfect model (e.g. reached on \texttt{mushroom} dataset). Figures \ref{fig:meta_phishing}, \ref{fig:meta_breast}, \ref{fig:meta_heart} and \ref{fig:meta_adult} provides the result for the best and worst two datasets. Figure \ref{fig:meta_all} shows all of the four Pareto frontiers. 
As expected, the results are better on \texttt{phishing} and \texttt{breast}. On \texttt{phishing}, \texttt{breast} and \texttt{heart}, the accuracy globally increases with $\eta$ while on \texttt{heart} the accuracy slightly decreases indicating poor influence of the hyperparameters and model.

Notice that for certain values of $\eta$ it is possible to reach 100\% accuracy with \texttt{heart} (sacrificing over 70\% of the dataset) while it is not with \texttt{breast}. Also, for high values of $\eta$, we observe a fall in accuracy for \texttt{breast}. We suspect those two phenomena to appear because we used the same value for $\eta_0$ and $\eta_1$. 


\section{Discussion}
\label{sec:discussion}

\subsection{On the hypergraph representation}

{The reader may have observed that the model space has a purely set-theoretic interpretation, and no hypergraph-specific property is used so far. Another possibility would be to use a bipartite graph with the first class being the cases of $\mathbf X$ and the second the elements of $\mathcal E$. We justify viewing the method from a hypergraph perspective by three axes currently being investigated:
\begin{itemize}
 \item {\bf Model space extension.} The current model space is not complex enough (see Section \ref{sec:msl}). An extension using hyperpaths is proposed in Section \ref{sec:mse}. Sets or graphs are not suitable to manipulate such space.
 \item {\bf Justification.} Most justification techniques provide a justification about the model itself (e.g. a decision tree\footnote{\url{https://github.com/andosa/treeinterpreter}}) or provide hints about each decisions under the form of weights for each variable \cite{ribeiro2016should}. In particular, \cite{lundberg2017unified} unifies the literature and formulates the justification as learning a simple model. 
 With \HCBR~we are exploring the possibility to generate explanations tailored for an element depending on its neighbors, in a case-based reasoning fashion. For this, we need a notion of neighborhood given by the hyperpaths. This is not possible with a set  approach, feasible with graphs but less natural than with hypergraphs.
 \item {\bf Performances.} Hypergraphs have computational advantages over graphs. Indeed, hypergraphs can be represented as graphs using clique expansion technique. However, increasing hyperdeges cardinality leads to a larger increase in the graph counterpart \cite{pu2012hypergraph}. 
\end{itemize}}

\subsection{On the model space extension}
\label{sec:mse}

{It is now clear that the model space is too limited. The number of parameters to describe a case is lower than $m$, the cardinal of the partition $\mathcal E$, and at most $|x|$. The number of parameters to describe the whole datasets is $m$.}

{To increase the model space complexity, there are two paths to explore in future work:
\begin{enumerate}
\item {\bf Increasing the number of parameters per case.} Currently, the support of $\mathbf x$ is modeled by a linear combination over the partition of its features s.t. the interactions between the elements of the partition are not taken into account. 
A reasonable way to go is to use some combinations of those elements. 
 For instance, if $\mathbf{x}$ is partitioned into $e_1$, $e_2$ instead of modeling the support as $s(\mathbf x) = w(e_1, \mathbf x)\mu(e_1) + w(e_2, \mathbf x)\mu(e_2)$, we would have $s(\mathbf x) = w(e_1, \mathbf x)\mu(e_1) + w(e_2, \mathbf x)\mu(e_2) + w(e_{1\cup2}, \mathbf x)\mu(e_{1 \cup 2})$. This would allow to be more discriminative and solve the problem of cases having the same support.
\item {\bf Increasing the number of parameters of the whole model.} A hyperpath of length $k$ is a sequence of hyperedges $(\mathbf x_1, ..., \mathbf x_k)$ such that $\forall i \in \{1,...k -1\}, ~ \mathbf x_i \cap \mathbf x_{i+1} \neq \emptyset$. Instead of modeling the support as a combination of the elements $\mathbf e_i$ belonging to a specific case $\mathbf x_i$, it could be extended to include the elements $\mathbf e \in \mathcal E$ belonging to the neighbor cases where a neighbor is a case that can be reached by a hyperpath of length $k$. Thus, the model space proposed in this paper would be the particular case with $k=0$.
\end{enumerate}
Naturally, it raises many questions, notably how to extend $w$ and $\mu$. The second point seems to be the most interesting because it really requires the hypergraph representation and cannot be interpreted using set theory.}

\subsection{On the model locality}
\label{sec:model_locality}

{The generalization capacity of \HCBR~depends on the number of intersections between the examples. Conversely, if a new case does not intersect with the examples, it is impossible to generate a prediction. Therefore, the locality property depends on how much the examples cover the feature space.}

{Consider that it is possible to choose $n$ examples of $k$ features in a space with $K$ features and $K \gg nk$. There is a tradeoff between covering as much space as possible and having enough intersections to construct a meaningful model. The first extreme configuration consists in maximizing the space cover: there is no intersection between cases, the accuracy during the training phase is 1 but the generalization capacity is null. The probability that a new case will intersect with some examples is maximized.
The second extreme configuration maximizes the intersections between the examples: all cases intersect with each other, therefore minimizing the space cover and thus the probability that a new case will intersect with the training set. The accuracy on the training set might not be 1 but the generalization capacity is high compared to the previous case.}

{For configurations close to the first extreme, it is possible to generate more intersections by using clustering or discretization in the original feature space. For instance, ``variable\_1=v\_1'' and ``variable\_1=v\_2'' could be encoded the same if both values belong to the same cluster. This requires slightly more feature engineering that previously stated.
For the second case, the problem lies in the training set itself. It is not specific to \HCBR, but to the fact the training set is not representative of the underlying distribution that generates the cases to classify.
In both cases, acquiring more data can help.}

{In the Additional Material, Section 8, we discuss a way of redefining the decision function to take into account this locality property.}

\section{Conclusion}
\label{sec:conclusion}

This paper presented \HCBR, a method for binary classification in unstructured space. 
The method can be seen as learning a metric that optimizes the classification score on the training set. Contrarily to most classification methods or metric learning methods, \HCBR~does not require to work in vector space and is agnostic to data representation. Therefore, it allows combining information from multiple sources by simply {\it stacking} the information. It does not require transforming the data to obtain satisfactory results.

The general framework introduced in Section \ref{sec:model} is instantiated in Section \ref{sec:model_selection} and \ref{sec:decision_training} where the support is determined using all the interactions between the hyperedges. Beyond this generic implementation, one can imagine different model selection methods using some assumptions or prior on the data.

\HCBR~has been tested on seven well-known structured datasets and demonstrated similar accuracy when compared to the best results from the literature, {with and without hyperparameter tuning.} We showed that the model is properly calibrated. Additionally, we performed a comparison with nine alternative methods to find out \HCBR, along with Neural Network, outperforms in average with a constant good result. {Those experiments showed that \HCBR~ can easily be used and deployed in practice, as it lowers the requirement for feature engineering, data preprocessing and hyperparameter tuning, i.e. the most consuming operations in practical machine learning nowadays.}

{In Section \ref{sec:unstructured_datasets}, we tested \HCBR~on unstructured datasets and showed it improves the accuracy in most cases compared to reference study.}

We empirically validated the worst-case complexity. Finally, we studied the properties and limitations of the model space.
We showed that the model selection procedure provides one of the best possible performance within the model space. Hence, further work will focus on extending the model space as proposed in Section \ref{sec:mse}.

This proof of concept raises many questions and offer many improvement axes. First, it seems relatively easy to extend the method to several classes, with a linear increase of the computation time. As calculating the class support represents most of the computational effort, working on an approximation of the main measure should be investigated. The solution may come from exploring the feature selection capacity of \HCBR. 
It may be possible to remove from the partition some elements that are not discriminative enough, reducing the computation time.

Additionally, we plan to investigate explanation generation about each prediction, using the link between cases in a similar way a lawyer may use past cases to make analogies or counter-examples. We also work on an online version of \HCBR~where the hypergraph is constructed case after case, including forgetting some old cases (which would allow handling non-stationary environment). It seems possible not only to add new examples dynamically, but also some vertices (i.e. adding some pertinent information to some cases) without generating the whole model from scratch. 

Last but not least, we would like to answer some questions: {can we provide some quality bounds depending on the initial hypergraph configuration w.r.t. the number of intersections and space cover?} How to handle continuous values without discretization?

\section*{Acknowledgment}

The author warmly thanks Dr. Jean-Fran\c{c}ois Puget, IBM Analytics, for his useful suggestions and advice in order to improve this paper.


\end{document}


\begin{frontmatter}

\title{Binary Classification in Unstructured Space With Hypergraph Case-Based Reasoning\\ (Additional Material)}

\author{Alexandre Quemy}
\address{IBM Krakow Software Lab, Cracow, Poland}
\ead{aquemy@pl.ibm.com}
\address{Faculty of Computing, Pozna\'n University of Technology, Pozna\'n, Poland}


\end{frontmatter}
\section*{Introduction}
This document contains additional material for the paper 'Binary Classication in Unstructured Space With Hypergraph Case-Based Reasoning', accepted for publication in the Information Systems special issue from DOLAP 2018.

\noindent
This material includes:
\begin{itemize}
\item Section \ref{1}: A formal proof of the training algorithm convergence,
\item Section \ref{2}: The average confusion matrix obtained in the experiments on structured datasets,
\item Sections \ref{3} and \ref{4}: The detailed comparison of \HCBR~with other methods (w.r.t. accuracy and MCC),
\item Section \ref{5}: The evolution of MCC depending on the training set size,
\item Section \ref{6}: A discussion on the learning curves,
\item Section \ref{7}: The hyperparameters for the genetic algorithms used to assess the model space limitation,
\item Section \ref{8}: A discussion on the model locality.
\end{itemize}

\section{On the convergence of the training algorithm}
\label{1}
For a training iteration $k$, let assume a case $\mathbf x_i$ with a strength $S_k(\mathbf x_i)$ that is wrongly classified. The update rule on $\mu$ implies that $S_{k+1}(\mathbf x_i) \in [-|S_k(\mathbf x_i)|, S_k(\mathbf x_i)]$. Any case $\mathbf x_j$ such that $\mathbf x_i \cap \mathbf x_j \neq \emptyset$ is modified, and in the same direction (toward the same class). Let us assume that $S_k(\mathbf x_i) < 0$ such that after the update rule $S_{k+1}(\mathbf x_i) > S_k(\mathbf x_i)$. Then, $S_{k+1}(\mathbf x_j) > S_{k}(\mathbf x_{j})$.

The only problematic case is when $y_j = 0$, $S_{k}(\mathbf x_j) < 0$ but $S_{k+1}(\mathbf x_j) > 0$ (that is to say that $\mathbf x_j$ become wrongly classified due to the modification of $\mu$ involved by $\mathbf x_i$).

Let us consider any $\mu_l$ for a $e_l$ that is included in $\mathbf x_i$ and $\mathbf x_j$. When $\mu_l$ is modified, then $|S(x_j)|$ is smaller because by definition $|w(e_l, x_j) \mu_l| \leq |S(x_j)|$.

As a result, both the case that {\it triggers} the modification of $\mu$ and the cases that are consequently modified have a strength that is closer to $0$ than before the modification.

Therefore, there are only two possible cases: \begin{itemize} \item All cases become correctly classified and the process stops. \item Some cases cannot be properly classified within the model space and switch iterations after iterations between classes. Their strength converges toward 0. It does not imply that the process converges toward the best possible accuracy. \end{itemize}

\newpage
\section{Average confusion matrix obtained with a 10-fold cross-validation.}
\label{2}
\begin{table}[h!]
\begin{center}
  \caption{Average confusion matrix obtained with a 10-fold cross-validation.}
  \begin{tabular}{|l|l|r|r|r|r|r|r|r|r|r|}
    \hline
     & & TP & FN & FP & TN \\
    \hline
    \multirow{2}{*}{\texttt{adult}} & without tuning  & 2182.4 & 295.3 & 288.5 & 488.8 \\
      & with tuning & 2226.6 & 245.3 & 311.4 & 472.7 \\\hline
    \multirow{2}{*}{\texttt{breast}} & without tuning & 23.0 & 1.4 & 0.7 & 43.9 \\
      & with tuning & 23.5 & 1.1 & 0.4 & 44.0 \\\hline
    \multirow{2}{*}{\texttt{heart}} &without tuning & 12.4 & 1.8 & 1.9 & 9.9 \\
      & with tuning & 13.5& 1.0 & 1.4 & 10.1 \\\hline
    \texttt{mushrooms} & without tuning & 390.6 & 0.0 & 0.0 & 420.4 \\\hline
    \multirow{2}{*}{\texttt{phishing}} & without tuning & 595.4 & 23.8 & 19.8 & 465.0\\
      & with tuning & 599.1 & 19.2 & 15.9 & 468.9 \\\hline
    \multirow{2}{*}{\texttt{skin}} & without tuning & 4886.3 & 132.4 & 199.4 & 19286.9 \\
      & with tuning & 4888.3 & 130.4 & 194.1 & 19292.2 \\\hline
    \multirow{2}{*}{\texttt{splice}} &without tuning & 155.7 & 9.1 & 8.5 & 142.7 \\
      & with tuning & 156.2 & 8.6 & 6.9 & 144.3 \\
    \hline
  \end{tabular}
  \label{table:confusion_matrix}
\end{center}
\end{table}

\newpage  
\section{Comparison of HCBR with several methods (Scikit-Learn implementation) w.r.t. MCC.}
\label{3}

\begin{table}[h!]
  \caption{Comparison of \HCBR~with several methods (Scikit-Learn implementation) w.r.t. MCC. }
  \resizebox{12cm}{!}{
  \begin{minipage}{.5\linewidth}
\fontsize{10pt}{12pt}\selectfont 
\begin{tabular}{|c|l|l|c|}
\hline
 Dataset & Method & MCC & \#  \\ \hline

\multirow{10}{*}{\texttt{adult}} & \bfHCBR & 0.5146 & 3 \\
& AdaBoost & 0.5455  & 1 \\
& k-NN & 0.4785 & 7 \\
& Linear SVM & 0.4918 & 5\\
& RBF SVM & 0.5065 & 4\\
& Decision Tree & 0.4821  & 6\\
& Rand. Forest & 0.3776 & 8 \\
& Neural Net & 0.5349  & 2 \\
& Naive Bayes & 0.2493 & 9\\
& QDA & 0.4785  & 7 \\ \hline

\multirow{10}{*}{\texttt{breast}} & \bfHCBR & 0.9222 & 3 \\
& AdaBoost & 0.9023 & 6\\
& k-NN & 0.9163 & 4\\
& Linear SVM & 0.9126 & 5\\
& RBF SVM & 0.8829 & 8\\
& Decision Tree & 0.8760 & 9\\
& Rand. Forest & 0.9296 & 1\\
& Neural Net & 0.9280 & 2\\
& Naive Bayes & 0.8991 & 7\\
& QDA & 0.8616 & 10\\ \hline

\multirow{10}{*}{\texttt{heart}} & \bfHCBR & 0.7082 & 1 \\
& AdaBoost & 0.5972  & 6\\
& k-NN & 0.5879 & 7\\
& Linear SVM & 0.6849 & 4\\
& RBF SVM & 0.6287 & 5 \\
& Decision Tree & 0.5763  & 8\\
& Rand. Forest &0.5703 & 9\\
& Neural Net & 0.6995 & 2\\
& Naive Bayes & 0.6932 & 3\\
& QDA & 0.4500 & 10 \\ \hline

\multirow{10}{*}{\texttt{mushrooms}} & \bfHCBR & 0.9995 & 2 \\
& AdaBoost & 1.0000 & 1\\
& k-NN & 0.9993 & 3\\
& Linear SVM & 1.0000 & 1\\
& RBF SVM & 0.9990 & 5\\
& Decision Tree &  0.9991& 4\\
& Rand. Forest & 0.8840  & 7\\
& Neural Net &1.0000 & 1\\
& Naive Bayes & 0.9767 & 6\\
& QDA & 1.0000 & 1\\ \hline

\end{tabular}
\end{minipage}%
    \begin{minipage}{.5\linewidth}
\fontsize{10pt}{12pt}\selectfont 
\begin{tabular}{|c|l|l|c|}
\hline
 Dataset & Method & MCC & \#  \\ \hline

\multirow{10}{*}{\texttt{phishing}} & \bfHCBR & 0.9191 &  1\\
& AdaBoost & 0.8637 & 6\\
& k-NN & 0.9138 & 4\\
& Linear SVM & 0.8740 & 5\\
& RBF SVM & 0.9286 & 2\\
& Decision Tree & 0.8585 & 7\\
& Rand. Forest & 0.7582 & 8\\
& Neural Net & 0.9448 & 1\\
& Naive Bayes & 0.5292 & 10\\
& QDA & 0.5872 & 9\\ \hline

\multirow{10}{*}{\texttt{skin}} & \bfHCBR & 0.9551 & 4 \\
& AdaBoost & 0.8552 & 8\\
& k-NN & 0.9982 & 1 \\
& Linear SVM & 0.8090 & 9\\
& RBF SVM & 0.9950 & 3 \\
& Decision Tree & 0.9544 & 5\\
& Rand. Forest & 0.9539 & 6\\
& Neural Net & 0.9967 & 2\\
& Naive Bayes & 0.7600 & 10\\
& QDA & 0.9483 & 7\\ \hline

\multirow{10}{*}{\texttt{splice}} & \bfHCBR & 0.8857 & 2 \\
& AdaBoost & 0.8801 & 3\\
& k-NN & 0.6072 & 9\\
& Linear SVM & 0.7282 & 8\\
& RBF SVM & 0.8461 & 4 \\
& Decision Tree & 0.8998 & 1\\
& Rand. Forest & 0.5925 & 10\\
& Neural Net & 0.8390 & 5\\
& Naive Bayes & 0.7595 & 7\\
& QDA & 0.8251 & 6\\ \hline
\end{tabular}
\end{minipage}
  }
  \label{table:scikit_all_mcc}
\end{table}

\section{Comparison of HCBR with several methods (Scikit-Learn implementation) w.r.t. accuracy.}
\label{4}
\begin{table}[h!]
  \caption{Comparison of \HCBR~ with w.r.t. accuracy. }
\resizebox{12cm}{!}{
  \begin{minipage}{.5\linewidth}
\fontsize{10pt}{12pt}\selectfont 
\begin{tabular}{|c|l|l|c|}
\hline
 Dataset & Method & Acc. & \#  \\ \hline

\multirow{10}{*}{\texttt{adult}} & \bfHCBR & 0.8232 & 5 \\
& AdaBoost & 0.8444  & 1 \\
& k-NN & 0.8156 & 7 \\
& Linear SVM & 0.8274 & 4\\
& RBF SVM & 0.8327 & 3 \\
& Decision Tree & 0.7918  & 8 \\
& Rand. Forest & 0.8223 & 6 \\
& Neural Net & 0.8378  & 2 \\
& Naive Bayes & 0.4675 & 10\\
& QDA & 0.7528  & 9 \\ \hline

\multirow{10}{*}{\texttt{breast}} & \bfHCBR & 0.9833 & 1 \\
& AdaBoost & 0.9563 & 4\\
& k-NN & 0.9614 & 3 \\
& Linear SVM & 0.9614 & 3\\
& RBF SVM & 0.9457 & 7 \\
& Decision Tree & 0.9429 & 9 \\
& Rand. Forest & 0.9543 & 5\\
& Neural Net & 0.9671 & 2\\
& Naive Bayes & 0.9533 & 6\\
& QDA & 0.9430 & 8 \\ \hline

\multirow{10}{*}{\texttt{heart}} & \bfHCBR & 0.8538 & 1 \\
& AdaBoost & 0.8037  &  6\\
& k-NN & 0.7926 & 7 \\
& Linear SVM & 0.8444 & 3\\
& RBF SVM & 0.8148 & 5 \\
& Decision Tree & 0.7556  & 9 \\
& Rand. Forest &0.7741 & 8 \\
& Neural Net & 0.8519 & 2\\
& Naive Bayes & 0.8444 & 3\\
& QDA & 0.8185 & 4\\ \hline

\multirow{10}{*}{\texttt{mushrooms}} & \bfHCBR & 0.9998 & 2 \\
& AdaBoost & 1.0000 & 1\\
& k-NN & 0.9996 & 3 \\
& Linear SVM & 1.0000 & 1\\
& RBF SVM & 0.9995 & 4 \\
& Decision Tree & 0.9996 & 3 \\
& Rand. Forest & 0.9582  & 6 \\
& Neural Net &1.0000 & 1\\
& Naive Bayes & 0.9882 & 5 \\
& QDA & 1.0000 & 1\\ \hline

\end{tabular}
\end{minipage}%
    \begin{minipage}{.5\linewidth}
\fontsize{10pt}{12pt}\selectfont 
\begin{tabular}{|c|l|l|c|}
\hline
 Dataset & Method & Acc. & \#  \\ \hline
\multirow{10}{*}{\texttt{phishing}} & \bfHCBR & 0.9645 & 3 \\
& AdaBoost & 0.9477 & 8\\
& k-NN & 0.9505 & 7\\
& Linear SVM & 0.9532 & 6\\
& RBF SVM & 0.9550 & 5\\
& Decision Tree & 0.9625 & 4\\
& Rand. Forest & 0.9738 & 1\\
& Neural Net & 0.9726 & 2\\
& Naive Bayes & 0.7062 & 10\\
& QDA & 0.7656 & 9\\ \hline

\multirow{10}{*}{\texttt{skin}} & \bfHCBR & 0.9847 & 4 \\
& AdaBoost & 0.9399 & 7 \\
& k-NN & 0.9994 & 1 \\
& Linear SVM & 0.9297 & 8 \\
& RBF SVM & 0.9984 & 3 \\
& Decision Tree & 0.9456 & 5 \\
& Rand. Forest & 0.9415 & 6 \\
& Neural Net & 0.9989 & 2 \\
& Naive Bayes & 0.8802 & 9 \\
& QDA & 0.8978 & 10 \\ \hline

\multirow{10}{*}{\texttt{splice}} & \bfHCBR & 0.9430 & 3 \\
& AdaBoost & 0.9528 & 2 \\
& k-NN & 0.7843 & 10 \\
& Linear SVM & 0.8645 & 9 \\
& RBF SVM & 0.9230 & 7 \\
& Decision Tree & 0.9415 & 4 \\
& Rand. Forest & 0.9399 & 5 \\
& Neural Net & 0.9195 & 8 \\
& Naive Bayes & 0.9245 &6 \\
& QDA & 0.9838 & 1 \\ \hline
\end{tabular}
\end{minipage}
}
  \label{table:scikit_results_acc}
\end{table}

\newpage
\section{Matthew Correlation Coefficient depending on the training set size.}
\label{5}
\begin{figure}[!h]\centering
\includegraphics[scale=0.3]{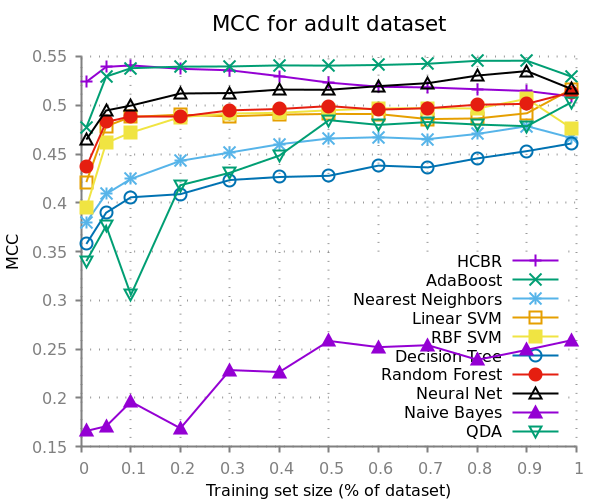}
\includegraphics[scale=0.3]{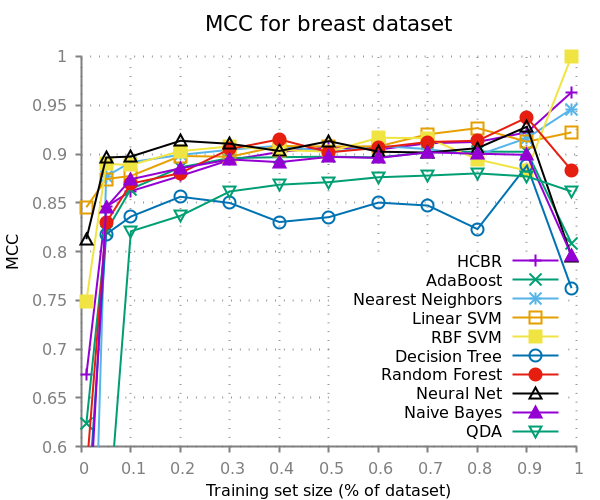}
\includegraphics[scale=0.3]{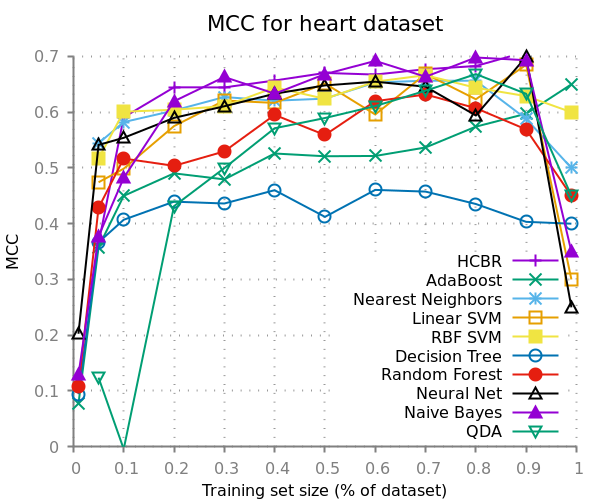}
\includegraphics[scale=0.3]{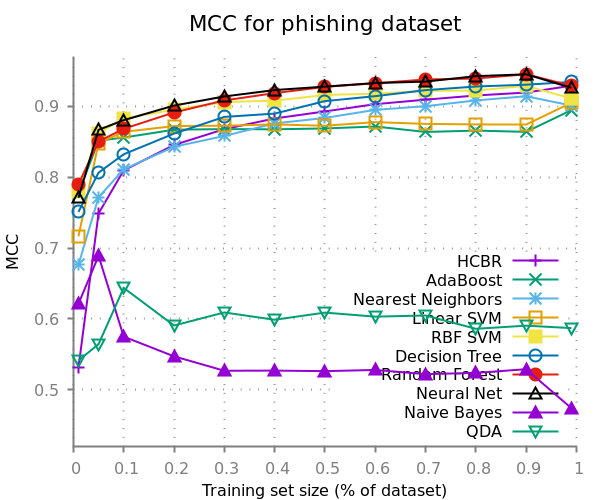}
\includegraphics[scale=0.3]{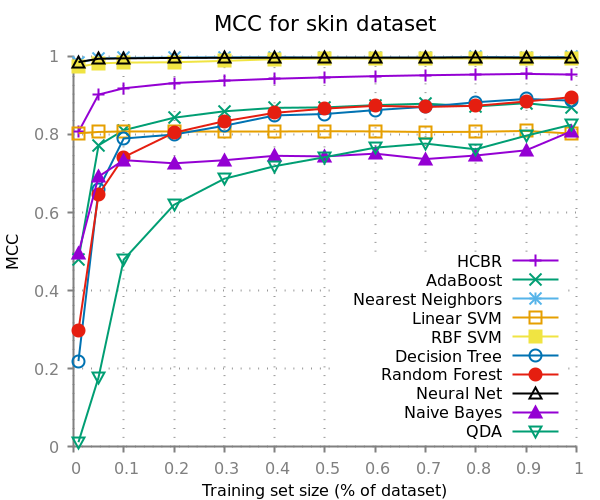}
\includegraphics[scale=0.3]{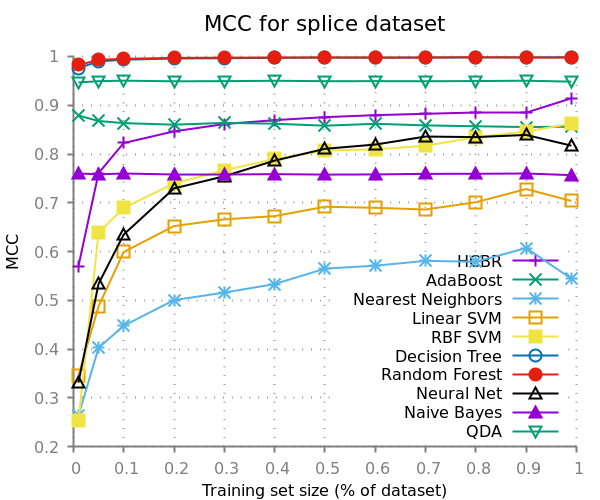}
\caption{Matthew Correlation Coefficient depending on the training set size.}
\label{fig:mcc_by_examples}
\end{figure}

\newpage
\section{Discussion on the learning curves}
\label{6}
\subsection{Heuristic}

In almost all datasets, in particular phishing and skin, many input vectors are the same, but with different output (e.g. for skin it's 583 cases i.e. 5.27\% of the dataset). By definition, there is no way to distinguish between those vectors, so the best we can do is to assign the class with the highest prevalence.

In most cases the model choses itself the class with the highest prevalence among those redundant input vectors or if it does not, it has very little impact (because the prevalence is close to 0.5 and the size of the redundant vector set is small in comparison to the dataset size).

The heuristic consists in estimating the prevalence of the set of redundant vectors in the training set, and then in bypassing the model prediction with the class associated to this prevalence. For instance, the overall prevalence of phishing is 0.5562 but the prevalence of the redundant vectors is 0.9674. About 5.27\% of the test set should consists of cases already in the casebase, among which 0.9674 are of class 1. It represents the insurance of a 5.27 x 0.9674 = 5.20\% of the training set correctly labeled.

The reason why the heuristic works in this case is because the prevalence of the redundant vectors set is higher than the accuracy obtained by the model (about 92.5\% versus 96\%). The gain corresponds exactly to this 5.20\% (because without the heuristic, with the grain used by the experiment, the model uses the wrong label).

This heuristic is independant of HCBR as it could be applied to any margin-based discrimative methods (e.g. SVM also cannot discriminate between redundant points with different labels).

\subsection{Learning curve}

The learning curves for phishing and skin are surprising for two reasons. First, the test accuracy is significantly higher than the training accuracy as depicted by Figure A6.1 and A6.2. The prediction phase on those two datasets uses the heuristic described above. For both datasets, the difference in accuracy between the two curves is much higher than the possible gain due to the heuristic. The learning curves for the same experiment without the heuristic are displayed on Figures A6.3 and A6.4 and it seems to explain perfectly the phenomena. Notice that this is specific to those two datasets that contains several redundant points with the same output. For instance, the heuristic is activated also on splice but as the number of redundant elements is very low w.r.t. the dataset size, the impact on the accuracy is not significant.

\begin{figure}[!h]\centering
\includegraphics[scale=0.3]{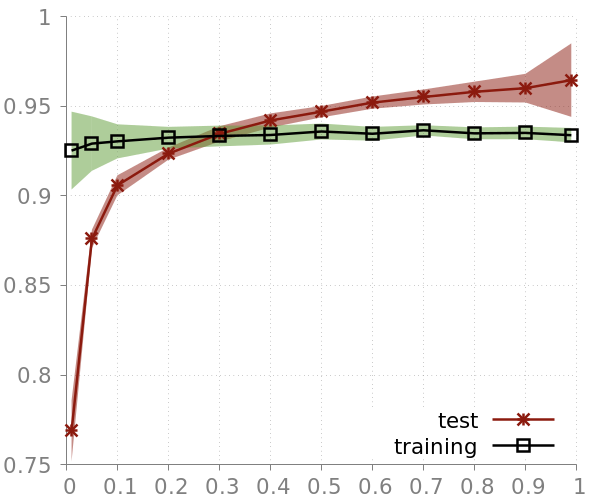}
\hfill
\includegraphics[scale=0.3]{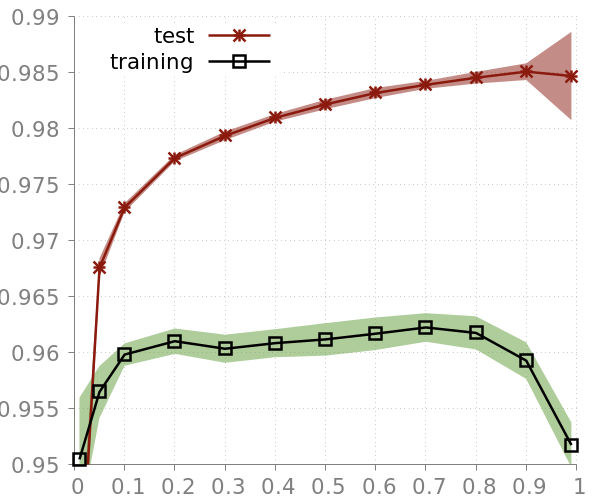}
\caption{Learning curves on \texttt{phishing} and \texttt{skin} datasets. The accuracy on the test set is higher than on the training set. On the training set, the accuracy starts by increasing and remains globally stable (with a drop for \texttt{skin}).}
\label{fig:learning_curve_3}
\end{figure}

\begin{figure}[!h]\centering
\includegraphics[scale=0.3]{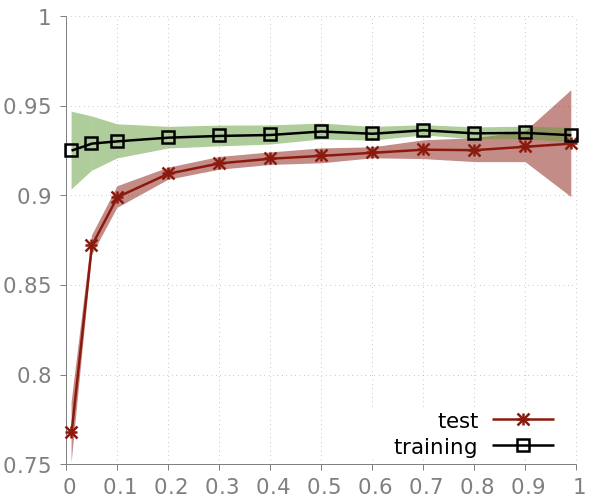}
\hfill
\includegraphics[scale=0.3]{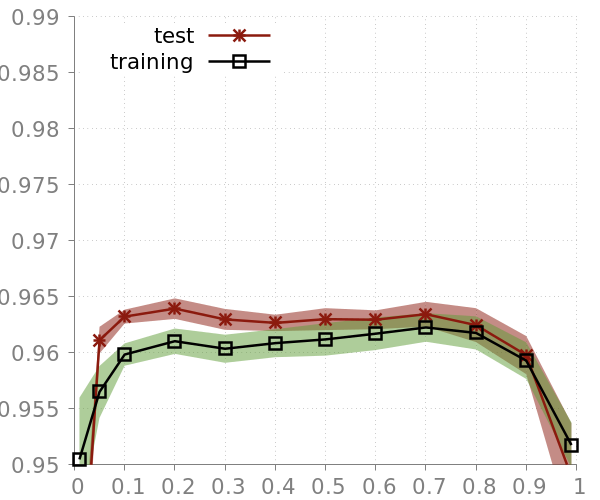}
\caption{Learning curves on \texttt{phishing} and \texttt{skin} datasets without the heuristic. Compared to Figure \ref{fig:learning_curve_3}, the learning curve behave as expected in theory.}
\label{fig:learning_curve_4}
\end{figure}

\begin{figure}[!h]\centering
\includegraphics[scale=0.3]{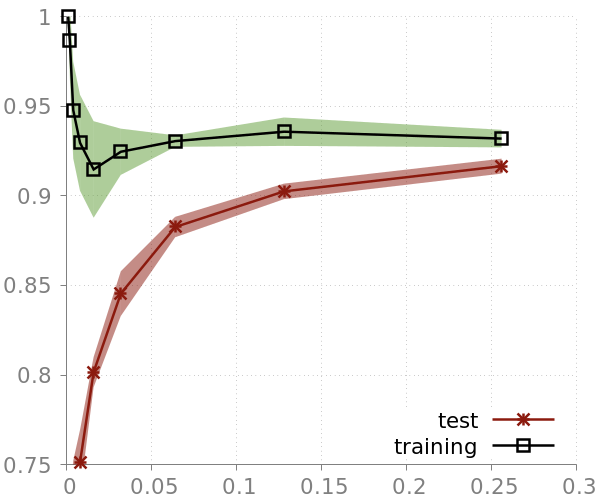}
\caption{Learning curves on \texttt{phishing} for very small training set sizes.}
\label{fig:learning_curve_5}
\end{figure}

\newpage
\section{Hyperparameters for the genetic algorithm.}
\label{7}
\begin{table}[h!]
\begin{center}
  \caption{Hyperparameters for the genetic algorithm. {\it Mutation $\sigma$ factor} is a factor used to set the standard deviation of the gaussian mutation. We use $\sigma = \frac{\mu^-}{\alpha}$ where $\alpha$ is the factor and $\mu^-$ defined by $\underset{i,j}{\min}{~ \mu_i - \mu_j}$.}
\begin{tabular}{|l|c|c|}
\hline
  Dataset & Generations & Mutation $\sigma$ factor \\ \hline
 \texttt{adult} & 200 & 10000 \\
  \texttt{breasts} & 200 & 10 \\
  \texttt{heart} & 200 & 10 \\
    \texttt{phishing} & 100 & 1000 \\
    \texttt{skin} & 100 & 1000 \\
    \texttt{splice} & 200 & 100 \\\hline
\end{tabular}
  \label{table:hyperparameters_ga}
\end{center}
\end{table}

\newpage
\section{Model Locality}
\label{8}

In case $\mathbf x$ has too many discretionary features, the classification rule is likely to be irrelevant. Indeed, the intersection between $\mathbf x$ and $\mathbb{F}_{\mathbf X}$ is to small to hold enough information and make strong analogies with $\mathbf x$. To overcome this drawback, $2^{\mathbb F}$ is split into two subsets:
\begin{itemize}
  \item $\mathcal{F}_1 = \{ \mathbf  x \in 2^{\mathbb F} ~ | ~ |\mathbf x \cap \mathbb{F}_{\mathbf X}| \geq \delta\}, ~ \forall \delta \in \mathbb{N}$
  \item $\mathcal{F}_2 = 2^{\mathbb F} \setminus \mathcal{F}_1$
\end{itemize}
$\mathcal{F}_1$ corresponds to the elements s.t. they share some features with the examples. An alternative may be considered by using $\mathcal{F}_1 = \{\mathbf  x \in 2^{\mathbb F} ~ | ~ \frac{D_\mathbf x}{|\mathbf x|} \leq \delta\}, ~ \forall \delta \in [0,1]$. In this case, $\mathcal{F}_1$ contains the elements for which we have enough information provided by the examples. From our preliminary tests, the choice depends on the dataset structure.

Finally, the decision rule for new cases is built as follows:
\begin{align}
\tag{R2} \label{eqn:updated_cr}
 \bar J(\mathbf x) = \left\{\begin{matrix}
 \tilde J(\mathbf x) & \text{if} ~ \mathbf x \in \mathcal{F}_1 \\
 o_{\mathbf x} & \text{if} ~ \mathbf x \in \mathcal{F}_2
\end{matrix}\right.
\end{align} where $o_{\mathbf x}$ is one draw from a random variable that has Bernoulli law with parameter $p=\frac{|\{\mathbf x \in \mathbf X | J(\mathbf x) = 1 \}|}{|\mathbf X|}$, i.e. the prevalence of class $1$ in $\mathbf X$. It assumes that the prevalence of $\mathbf X$ is close to the prevalence over $2^{\mathbb F}$ (or that the prevalence does not change in time for sequential problems in which the new cases are generated by an unknown random measure). The rationale behind is that if for a case $\mathbf x$, it is not possible to exploit the model built on the hypergraph, then we can still model $J$ as a Bernoulli random variable and use a maximum likelihood estimation for $p$. In a sense, it is extending the {\it local} model to the entire input space $2^\mathbb F$.